\documentclass[11pt, a4paper, parskip=half, fleqn]{scrartcl}

\usepackage[T1]{fontenc}
\usepackage[utf8]{inputenc}
\usepackage{lmodern}
\usepackage{microtype}
\usepackage{geometry}
\usepackage{amsmath, amssymb, amsthm}
\usepackage{caption}
\usepackage{subcaption}
\usepackage{graphicx}

\usepackage{array, booktabs, multirow}

\usepackage{algorithm, algpseudocodex}
\floatname{algorithm}{algorithm} %make caption algorithm nb rather than Algorithm nb

\usepackage{xcolor}
\definecolor{brick}{RGB}{178,34,34}

\usepackage[colorlinks=true,linkcolor=brick,citecolor=brick,urlcolor=brick]{hyperref}
\usepackage{cleveref}

\usepackage[backend=biber,style=alphabetic, maxnames=30, backref=true]{biblatex}
\usepackage[scaled]{helvet} % --> will use Helvetica font for sans-serif text

\setkomafont{section}{\large}        % default is Huge → reduce it
\setkomafont{subsection}{\normalsize}
\setkomafont{subsubsection}{\normalsize}

\RedeclareSectionCommand[
  beforeskip=0.7\baselineskip,
  afterskip=0.3\baselineskip
]{section}

\RedeclareSectionCommand[
  beforeskip=0.5\baselineskip,
  afterskip=0.2\baselineskip
]{subsection}

\RedeclareSectionCommand[
  beforeskip=0.3\baselineskip,
  afterskip=0.1\baselineskip
]{subsubsection}

\newcommand{\pbreak}{%
    \par\medskip
    \centerline{{\sffamily\textbf{Intermediate summary}}}
}

\newcommand{\rtheta}{\rho_\theta}
\newcommand{\rzero}{\rho_0}
\newcommand{\xk}{x^{(k)}}

\newcommand{\R}{\mathbb{R}}
\newcommand{\E}{\mathbb{E}}
\newcommand{\X}{\mathcal{X}}
\newcommand{\N}{\mathcal{N}}
\newcommand{\Z}{\mathcal{Z}}

\newcommand{\dd}{\mathrm{d}}
\usepackage{pdfpages}

\begin{document}

\begin{flushleft}
  {\Large\sffamily\bfseries Leveraging generative models to assist Monte Carlo sampling\par}
  {\large\sffamily\itshape (An example of scientific computing assisted by modern machine learning techniques)\par}
  {\normalsize Marylou Gabrié\par}
  {\emph{Laboratoire de Physique Statistique, École normale supérieure,
   PSL Research University,\\
    24 rue Lhomond, 75005 Paris, France}\\
    \href{mailto:marylou.gabrie@ens.fr}{marylou.gabrie@ens.fr}}
\end{flushleft}

\begin{abstract}
  Sampling high-dimensional probability distributions is a central task in scientific computing, with applications ranging from Bayesian inference to statistical physics and molecular simulation. Despite decades of methodological developments, two major challenges remain: scaling to high dimensions and efficiently exploring multimodal distributions characterized by metastable states. Classical approaches such as Markov chain Monte Carlo, tempering methods, or enhanced sampling based on collective variables have achieved major successes, but they also face intrinsic limitations. This tutorial review explores a new paradigm that has recently emerged at the interface of machine learning and computational statistical physics: the use of generative models as tools for sampling. In this context, models such as normalizing flows and diffusion models are not used in their traditional data-driven setting, but rather as flexible probabilistic models that can assist the sampling of distributions known only up to a normalization constant. This manuscript reviews the early development of this rapidly evolving field and discusses several methodological directions, including exact samplers based on generative models and strategies to train such models in the absence of data. While an exhaustive survey of the literature is not attempted, we present a selection of key ideas and methods, along with a discussion of their strengths and limitations. The review is intended to be an accessible tutorial for both physics and machine learning audiences, and it aims to provide a starting point for researchers interested in exploring this exciting area of research.
\end{abstract}

% \pagebreak

\tableofcontents
\setcounter{tocdepth}{1}

% \pagebreak
\section*{List of Notations and Acronyms}

\begin{tabular}{p{3cm}p{12cm}}
\textbf{Notation} & \textbf{Definition} \\
\hline
$I_d$ & Identity matrix of dimension $d$ \\
$0_d$ & Zero vector of dimension $d$ \\
$\N(\mu, \Sigma)$ & Gaussian distribution with mean $\mu$ and covariance $\Sigma$ \\
$\N(\,\cdot\,; \mu, \Sigma)$ & Gaussian probability density function with mean $\mu$ and covariance $\Sigma$ \\
$\mathrm{StopGrad}(\cdot)$ & Stop gradient: dependency on parameters is not backpropagated \\
\end{tabular}

\begin{tabular}{p{3cm}p{12cm}}
\textbf{Acronym} & \textbf{Definition} \\
\hline
AIS & annealed importance sampling \\
BAR & Bennett acceptance ratio \\
CNF & continuous normalizing flows \\
CV & collective variable \\
DM & diffusion models \\
ESS & effective sample size \\
HMC & Hamiltonian Monte Carlo \\
IMH & independent Metropolis-Hastings \\
IS & importance sampling \\
iSIR & iterated sampling importance resampling \\
MALA & Metropolis-adjusted Langevin algorithm \\
MCMC & Markov chain Monte Carlo \\
MH & Metropolis-Hastings \\
NF & normalizing flows \\
ODE & ordinary differential equation \\
OU & Ornstein-Uhlenbeck \\
PINN & physics-informed neural network \\
PT & parallel tempering \\
SDE & stochastic differential equation \\
SI & stochastic interpolants \\
SMC & sequential Monte Carlo \\
TFEP & targeted free energy perturbation \\
ULA & unadjusted Langevin algorithm \\
VI & variational inference \\
\end{tabular}

\listofalgorithms

\pagebreak

\section{Introduction}
\label{sec:01-intro}
\subsection{Why sampling? Why is it hard?}

\paragraph{Sampling and its challenges}
Sampling is a fundamental task in scientific computing, as it enables the efficient approximation of high-dimensional integrals. In observational and experimental sciences, for instance, sampling is essential for uncertainty quantification and model calibration through Bayesian inference. In theoretical physics, sampling is an indispensable computational tool for studying many-body systems across a broad spectrum of applications, ranging from lattice gauge theory for the strong nuclear force to quantum Monte Carlo methods for solving the Schrödinger equation or investigating the thermodynamic properties of materials. 

The task of sampling consists in generating realizations of a random variable according to a given target distribution. In this tutorial, we focus on the common setting where the state space is continuous and the probability density function—denoted by $\pi$ throughout—is known up to a normalization constant. Sampling from $\pi$ shares certain features with optimization: both require identifying the most probable regions of the state space, typically located near local maxima of the target density. Crucially, however, sampling must also correctly capture fluctuations around these maxima and reproduce the relative weights of different regions. It is therefore a fundamentally different problem from optimization, with its own intrinsic challenges.

The first modern sampling algorithms emerged in the early 1950s, enabled by the advent of digital computers. The Metropolis algorithm \cite{metropolisMonteCarloMethod1949,vonNeumannVariousTechniques1951,metropolisEquationStateCalculations1953} was the first Markov Chain Monte Carlo (MCMC) method and laid the foundation for a vast family of techniques that followed. Since then, an extensive literature has developed, proposing a wide variety of algorithms tailored to the diverse challenges arising across application domains and target distributions (see, e.g., \cite{liuMonteCarloStrategies2004} for a comprehensive introduction). 

To this day, two major sources of difficulty remain central: \emph{scalability}, the ability to handle high-dimensional target measures, and \emph{metastability}—also referred to as multimodality—the presence of multiple well-separated local maxima in the measure. MCMC methods can often cope with moderately high-dimensional spaces by exploring the state space through random walks. However, their essentially local exploration mechanism typically makes transitions between distant high-probability regions highly unlikely. As a result, the chain may remain trapped for extremely long times in so-called metastable states. A striking illustration is provided by protein folding: although the folded configuration corresponds to the global minimum of the Boltzmann free energy, locating this free-energy minimum by sampling the associated Boltzmann distribution with traditional, local molecular dynamics or MCMC methods is essentially impossible in practice, as the dynamics remains trapped in long-lived metastable intermediates.

\paragraph{Important families of sampling algorithms to tackle metastability in high dimension}  To clarify why metastability is such a difficult problem, it is useful to briefly review the main families of sampling algorithms that have been developed to address it.

\emph{Non-local jump samplers.}
As previously mentioned, Markov Chain Monte Carlo (MCMC) methods typically rely on local proposals. While this ensures reasonable scalability with dimension, it severely limits mixing in the presence of metastability, making transitions between modes exponentially rare with respect to the energy barriers separating them. To address this issue, various approaches have been proposed to incorporate non-local moves. 

In certain specific settings, remarkably effective solutions have been developed. 
In lattice spin systems with discrete variables, cluster algorithms such as the Swendsen-Wang \cite{swendsenNonuniversalCriticalDynamics1987} and Wolff \cite{wolffCollectiveMonteCarlo1989} updates enable collective moves of large groups of spins and can dramatically accelerate sampling close to the critical region despite of a diverging correlation length.
In the context of particle systems, several non-local strategies have also proven highly effective. Swap Monte Carlo \cite{grigeraFastMonteCarlo2001,berthierEfficientSwapAlgorithms2019}—which proposes exchanges of particle identities in addition to local displacements—can accelerate equilibration of glass-forming systems by many orders of magnitude. Event-chain Monte Carlo \cite{bernardEventchainMonteCarlo2009}, a rejection-free irreversible algorithm that propagates displacements through chains of particles, has likewise led to major improvements in the simulation of dense particle systems. 

Meanwhile, the statistics literature has attempted to construct general-purpose mode-hopping proposals \cite{tjelmelandModeJumpingProposals2001b,pompeFrameworkAdaptiveMCMC2020}. However, designing effective non-local proposals in general is challenging as it requires identifying high-density regions that typically occupy an increasingly small fraction of the state space as the dimension grows. Simple parameterizations of such proposals, such as Gaussians or mixtures of Gaussians, usually fail in high dimension, since these approximations remain too crude and lie too far from the true target distribution.

\emph{Annealing-based samplers.}
A more robust family of methods for sampling complex distributions is based on tempering, also known as annealing. The central idea is to define a sequence of intermediate distributions that bridge a simple base distribution and the complex target distribution. This bridge is commonly constructed by gradually increasing an effective temperature or weakening interactions between variables until reaching an independent (product) measure. The resulting path is then exploited to progressively transport samples toward the target, which can substantially improve exploration in multimodal settings.
Among the most widely used tempering-based methods are annealed importance sampling (AIS) \cite{nealAnnealedImportanceSampling2001}, parallel tempering (PT, also known as replica exchange) \cite{swendsenReplicaMonteCarlo1986,geyerMarkovChainMonte1991,hukushimaExchangeMonteCarlo1996}, and sequential Monte Carlo (SMC, also known as population annealing) \cite{hukushimaPopulationAnnealingIts2003,delmoralSequentialMonteCarlo2006a}. 

Despite their effectiveness, annealing-based methods present practical challenges. Notably, the choice of a suitable annealing path and of the number of intermediate distributions have a decisive impact on performance. Self-adaptive SMC \cite{chopinIntroductionSequentialMonte2020} provides a principled mechanism to adjust the number of steps dynamically and is among the most effective algorithms currently available for multimodal targets. However, annealing-based methods may still struggle when the reference measure differs too drastically from the target distribution. In protein conformational sampling, for example, high-temperature annealing may lead to complete denaturation, so that configurations obtained at high temperature carry little useful information about the folded state. Similarly, in statistical mechanics, the presence of a discontinuous (first-order) phase transition along the annealing path can break the connection between high- and low-temperature regimes, rendering the bridge ineffective.
Thus, while annealing-based samplers represent the state of the art for addressing multimodality, alternative strategies remain necessary — in particular in situations where constructing a suitable annealing path is difficult, or where one may hope to bypass annealing altogether through more direct and potentially more efficient approaches.

\emph{Dimensionality reduction to enhance sampling.}
Another classical strategy to tackle high-dimensional multimodal sampling problems consists in identifying a low-dimensional projection of the full configuration space — often referred to as a set of collective variables (CVs), or equivalently a coarse-graining of the system — that resolves the different modes of the target distribution. Sampling across these modes is then driven by forcing the exploration in the CV space which is typically of much lower dimension than the full space. These approaches, particularly popular in molecular simulation, come in many variants; a recent review can be found in \cite{heninEnhancedSamplingMethods2022}.
However, these techniques face two major challenges. 
First, they are mostly restricted to very low-dimensional CV spaces, since the non-parametric estimation of the marginal distribution they rely on becomes rapidly intractable as the dimension of the reduced space increases.
Second, identifying an appropriate set of collective variables — that is, a suitable low-dimensional projection capable of separating the relevant metastable states — is itself a difficult task, especially under stringent dimensionality constraints.

Going beyond these different families of samplers for multimodal distributions, and building on recent advances in machine learning, a new paradigm for sampling has begun to emerge, leveraging the expressive power of generative models to learn complex probability distributions and produce representative samples.
In the following, we aim to present the main ideas and results in this rapidly developing field. In what sense do these methods overcome the limitations discussed above? How can they be combined with classical approaches? What challenges and open questions remain?

\subsection{Generative models for sampling}

\paragraph{Generative modelling vs sampling}
Let us first clarify the difference between generative modelling and sampling, as these two terms are often used interchangeably in the literature. \emph{Generative modelling} consists in learning a probabilistic model -- namely the generative model -- given an extensive set of samples from an unknown distribution. It is a fundamentally data-driven task of density estimation, where the goal is to learn a model that can generate new samples that are similar to the training data, this similarity being not necessarily well-defined\footnote{For one thing, the distribution most faithful to a training dataset is the sum of Dirac deltas at the training points.}. In contrast, \emph{sampling} is the task of generating samples from a target distribution whose probability density function is known (possibly up to a normalizing constant). The task is different in two fundamental ways: (i) instead of a large dataset, information must be extracted from the knowledge of the mathematical expression of the target measure and little if any data is available; (ii) the objective is to produce samples that are exactly distributed according to the target measure, which has a well-defined mathematical meaning.

\paragraph{Generative models for scientific computing}
Generative models per se are highly flexible probabilistic models, usually parametrized via deep neural networks, which were primarily developed for generative modelling tasks on images and text, with the incredible successes we know. They have also  already proven successful in data-driven computational tasks in physics and chemistry, such as protein folding with AlphaFold \cite{jumperHighlyAccurateProtein2021} and predicting protein-ligand binding structures
\cite{corsoDiffDockDiffusionSteps2023}. 

Here, we will discuss how generative models are also proving to be powerful tools for sampling.
Pioneering works in this direction \cite{liNeuralNetworkRenormalization2018,albergoFlowbasedGenerativeModels2019,noeBoltzmannGeneratorsSampling2019} have demonstrated that subclasses of generative models, namely normalizing flows and autoregressive networks, can rather easily be repurposed for exact sampling by enabling reweighing of the generated realizations with respect to the target measure. Proofs of concept have been shown for various areas of physics and chemistry, including lattice quantum field-theories \cite{abbottApplicationsFlowModels2024}, biomolecules \cite{noeBoltzmannGeneratorsSampling2019} or, in sampling nano-clusters of heavy atoms at quantum accuracy \cite{molina-tabordaActiveLearningBoltzmann2024a}.

These early successes have also uncovered important limitations of such direct reweighing strategies, and have spurred the development of more sophisticated approaches.
As a result, a broad landscape of methods has emerged, ranging from refined flow-based constructions to the most recent diffusion models adapted to sampling tasks. This diversity has created a rapidly evolving “zoo” of techniques in which conceptual connections are not always transparent and methodological differences can be subtle. In this tutorial review, we survey a representative subset of these approaches with the aim of clarifying their underlying principles, highlighting their common structure, and emphasizing the trade-offs they entail. While striving to remain concise, we focus on the core ideas and main results, and refer the reader to the original works for technical details and further developments.

\paragraph{Previous partial reviews of sampling with generative models}
As the field continues to gain momentum, any attempt at exhaustiveness would be futile. Nevertheless, assembling the first chapters of this rapidly developing story can be valuable, particularly for newcomers seeking an entry point into the subject.
%  As of March 2026, the field is evolving at a fast pace, and new ideas have already appeared that are not covered in the present manuscript.
% We would also like to point to s
Also, several works have already attempted to summarize specific aspects of the field. Chapter 7.2 of the textbook \cite{dawidMachineLearningQuantum2025}, coordinated by Anna Dawid following a doctoral school on machine learning for quantum physics held in September 2021, reviews early developments in the use of generative models for many-body physics. The lecture notes by Eric Vanden-Eijnden for the Les Houches Summer School on Statistical Physics \& Machine Learning (August 2022) introduce foundational principles of sampling with generative models \cite{albergoLearningSampleBetter2024}. The commentary \cite{corettiBoltzmannGeneratorsNew2024} discusses the pioneering work \cite{noeBoltzmannGeneratorsSampling2019} on Boltzmann generators and its immediate extensions.
More focused reviews include \cite{cranmerAdvancesMachinelearningbasedSampling2023}, which surveys advances with specific applications to lattice quantum chromodynamics, \cite{heNoTrickNo2025}, which places different diffusion-based sampling approaches into perspective and \cite{rotskoffSamplingThermodynamicEnsembles2024,olssonGenerativeMolecularDynamics2026}, which reviews the use of generative models for molecular dynamics simulations.

\subsection{Outline of the review}
The ``pipeline'' of sampling with generative models can be decomposed into three main stages, which will guide the structure of what follows. The first stage (i) consists in choosing a suitable family of generative models, typically a transport-based model, such as a normalizing flow or a diffusion model, or an autoregressive model.
\begin{itemize}
    \item[]  \Cref{sec:02-tuto_gm} briefly reviews these generative models, both for completeness and to introduce the notations used throughout the manuscript. It also recalls how these models are traditionally trained from data in the standard “generative modelling” setting.
\end{itemize}
The next two stages in sampling with generative models are (ii) training the generative model in the specific context of sampling, where no data is available, and (iii) designing sampling algorithms that leverage the trained model to generate samples from the target distribution. These two stages are closely intertwined. In particular, as we will see, the training procedure can itself exploit the sampling capabilities of the generative model. For this reason, the exposition does not strictly follow the chronological order of the pipeline, but instead adopts an organization that we believe makes the underlying ideas clearer.
\begin{itemize}
    \item[] \Cref{sec:03-exact_samplers} focuses on the design of exact sampling algorithms, assuming that a generative model has already been trained. This section concentrates in particular on algorithms based on exact-likelihood generative models, “discrete-time” normalizing flows and autoregressive models, which were the first generative models to be used for sampling and provide a transparent setting to understand the core principles. We move from early, straightforward reweighing strategies to more sophisticated approaches aimed at extending the scope of these methods, notably in the presence of metastability.
    \item[] \Cref{sec:04-training} then takes a step back to examine the training of generative models for sampling. Starting with procedures originally developed for “discrete-time” normalizing flows, we discuss how these ideas extend to more recent classes of models, such as continuous normalizing flows and diffusion models.
    \item[] \Cref{sec:05-cnf_fm_sampling} returns to the design of sampling algorithms, this time focusing on recent developments involving “continuous-time” normalizing flows and diffusion models.
    \item[] \Cref{sec:07-perspectives} concludes with perspectives on the field and points to directions of interest that were not covered in this review.
\end{itemize}

\pagebreak
\section{A brief introduction to generative models used for sampling}
\label{sec:02-tuto_gm}   
We start with a brief review of the different types of generative models used to assist Monte Carlo sampling, introducing the notations employed throughout the rest of the review. These models are of two kinds: autoregressive models, which generate samples sequentially, and transport-based models, which transform samples from a simple base distribution into complex outputs. These generative approaches are particularly well-suited for sampling tasks as they give more or less direct access to the probability density of the generated samples, which is a key ingredient to build exact samplers from generative models as we will discuss in \Cref{sec:03-exact_samplers}. 

\emph{In this section, we focus on the traditional setting of generative modelling, where a dataset of training samples is available while their underlying target distribution is unknown.} In \Cref{sec:04-training}, we will discuss how to train generative models in the absence of data, which is the relevant setting for sampling tasks where instead the target distribution is known up to a normalization constant.

\subsection{Autoregressive models}
Autoregressive models (ARMs) are generative models that factorize the joint distribution of a random vector $X=(X_1,\ldots,X_d) \in \X$ into a product of univariate conditional distributions. The probability density function, or the probability distribution if $\X$ is discrete, is parameterized as
\begin{align}
\label{eq:02-ARM-factorization}
    \rtheta(x) = \prod_{i=1}^d \rho_{\theta,i}(x_i \mid x_{<i}),
\end{align}   
where $x_{<i} = (x_1,\ldots,x_{i-1})$ denotes the vector of the first $i-1$ components of $x$, and $\rho_{\theta,i}$ is a learnable conditional distribution with parameter vector $\theta \in \Theta$. Any joint distribution, for any ordering of the coordinates, can be factorized in this autoregressive fashion. However, the finite expressiveness of the parametric families of $\rho_{\theta,i}$, built in practice with neural networks, renders the model sensitive to the choice of ordering which can be treated as an hyperparameter. 
Examples of choices of parametrizations for the conditional distributions $\rho_{\theta,i}$ are given by \cite{germainMADEMaskedAutoencoder2015,uriaNeuralAutoregressiveDistribution2016c}.

Within this autoregressive framework, sampling is performed sequentially, by first sampling $x_1 \sim \rho_{\theta,1}$, then $x_2 \sim \rho_{\theta,2}(\cdot \mid x_1)$, and so on until $x_d \sim \rho_{\theta,d}(\cdot \mid x_{<d})$. Moreover, 
owing to their tractable probability (density), autoregressivee models can be trained for density estimation via maximum likelihood. Given a dataset of training samples $\{\xk\}_{k=1}^N$ assumed to be i.i.d. from an unknown target probability distribution $\pi$, one learns parameters $\theta$ of the NF by minimizing the negative log-likelihood of the training set,
\begin{align}
\theta \in \arg\min - \sum_{k=1}^{N} \log \rtheta(\xk).
\end{align}

This approach has been successfully applied in various domains, including image generation and natural language processing. However, autoregressive models are mainly used for discrete state spaces, as for continuous state spaces they tend to be less flexible than transport-based models, which we discuss in the next sections.

\subsection{Normalizing flows}

\subsubsection{Original formulation}
Normalizing flows (NFs) were introduced by Tabak \& Vanden Eijnden \cite{tabakDensityEstimationDual2010} as a method for density estimation on $\R^d$, based on learning the transport map from an empirical distribution toward a simple, typically Gaussian, base distribution, hence the name \emph{normalizing} flow. More generally, denote $\rzero$ the density of the base distribution on the state space $\X \subset \R^d$, and $T_\theta$ a diffeomorphism on $\X$ with parameter vector $\theta \in \Theta$. A NF is a learnable probabilistic model with realizations $x$ sampled through,
\begin{align}
    z \sim \rzero, \quad x = T_\theta(z),
\end{align}
and associated probability density function given by the change-of-variables formula,
\begin{align}
\label{eq:02-nf-change-of-variable}
\rtheta(x) = \rzero\left(T_\theta^{-1}(x)\right)\left|\det \nabla\left(T_\theta^{-1}(x)\right)\right|.
\end{align}

According to Brenier's theorem \cite{brenierPolarFactorizationMonotone1991}, provided that $\rzero$ is absolutely continuous with respect to the Lebesgue measure, there exists, for any absolutely continuous target distribution $\pi$, a transport map $T^*$ that pushes forward $\rzero$ to $\pi$. This theoretical result justifies the use of transport maps as generative models, as they are, in principle, capable of representing arbitrary target distributions. In practice, however, the expressivness of the model depends on the chosen parametrization for $T_\theta$.

The central challenge is therefore to design parametrized maps $T_\theta$ that are sufficiently expressive while allowing efficient evaluation of the inverse map and the determinant of its Jacobian. To this end, normalizing flows are typically constructed as compositions of simpler transformations
\begin{align}
    T_\theta = T_{\theta_L} \circ \ldots \circ T_{\theta_1},
\end{align}
each $T_{\theta_i}$ interpreted as a layer involving learnable neural networks and $L$ being the number of layers, or depth, of the NF. The so-called coupling layer trick is behind many popular architectures for these simple invertible transformations, with prominent examples including real non-volume preserving flows (RealNVP) \cite{dinhDensityEstimationUsing2017a} and neural spline flows (NSF) \cite{durkanNeuralSplineFlows2019a}. Comprehensive reviews of design choices for NFs can be found in \cite{papamakariosNormalizingFlowsProbabilistic2021a,kobyzevNormalizingFlowsIntroduction2021}.

For training in the traditional generative modeling setting, the negative log-likelihood of a training dataset $\{\xk\}_{k=1}^N$ can be minimized:
\begin{align}
    \theta \in \arg\min - \sum_{k=1}^{N} \log \rtheta(\xk),
\end{align}
which is available in closed form thanks to the change-of-variables formula (\cref{eq:02-nf-change-of-variable}) and architectures designed to allow efficient evaluation of the inverse map and the determinant of its Jacobian.
While extremely convenient for density estimation, these architectures  
can nevertheless prove too restrictive when the target distribution is highly complex and multimodal. To overcome this issue, the more recent \emph{continuous normalizing flows} rely on continuous-time dynamics to define the transport map. By opposition, the original NF architecture are sometimes referred to as \emph{discrete-time} normalizing flows.

\subsubsection{Continuous normalizing flows}
\label{subsec:02-cnfs}

As an alternative to finite compositions of invertible layers,
Chen and co-authors introduced neural ordinary differential equations (NeuralODEs) \cite{chenNeuralOrdinaryDifferential2018}.
In this framework, samples from the base distribution are transported by the integration of a learnable velocity field $v_\theta : \X \times [0,T] \to \X$, from time $t=0$ up to time $t=T$,
\begin{align}
% \begin{cases}
    \displaystyle \frac{\dd X_t}{\dd t} = v_\theta(X_t, t)\, , \quad
    X_0 \sim \rzero.
% \end{cases}.
\end{align}
The samples produced by the continuous normalizing flow (CNF) correspond to the final states $X_T$.
Denoting by $\rho_{\theta,t}$ the time-marginal distribution of the process $(X_t)_{t\in[0,T]}$ defined above, the associated continuity equation reads
\begin{align}
\label{eq:02-continuity-equation}
\frac{\partial \rho_{\theta,t}(x)}{\partial t} + \nabla \cdot \big(\rho_{\theta,t}(x)\, v_\theta(x, t)\big) = 0.
\end{align}
Thus, the probability density function of the generated samples can be computed by integrating, along the flow trajectory, the instantaneous change-of-variables formula derived from the continuity equation:
\begin{align}
\label{eq:02-instantaneous-change-of-variables-a}
& \frac{\dd \log \rho_{\theta,t}(X_t)}{\dd t}
= \frac{\partial \log \rho_{\theta,t}(X_t)}{\partial t}
+ v_\theta(X_t, t) \cdot \nabla \log \rho_{\theta,t}(X_t)
= - \nabla \cdot v_\theta(X_t, t), \\
\label{eq:02-instantaneous-change-of-variables-b}
\Rightarrow \quad &
\log \rho_{\theta,T}(X_T)
= \log \rho_{\theta,0}(X_0)
- \int_0^T \nabla \cdot v_\theta(X_t, t), \dd t.
\end{align}

In practice, this formula is evaluated either along a forward trajectory during sampling (increasing time $t$ from $X_0 \sim \rzero$), or along a backward trajectory when estimating the likelihood of samples from a training set (decreasing time $t$ from $X_T = X^{(k)}$).

With CNFs, both sampling and likelihood evaluation require the numerical integration of an ordinary differential equation (ODE), which is computationally costly and introduces discretization errors. In particular, likelihood computations involve evaluating the divergence of the velocity field, which can be obtained via automatic differentiation but further increases the computational cost.\footnote{The divergence $\nabla \cdot v(x)$ is the trace of the Jacobian $\mathrm{Tr}(\nabla v(x))$. To avoid forming the full Jacobian requiring $d$ backpropagation passes, Hutchinson's estimator approximates this trace by using the identity $\mathrm{Tr}(A)=\mathbb{E}_\xi[\xi^\top A \xi]$, where $\xi$ is a random vector with zero mean and identity covariance (e.g. standard Gaussian or Rademacher) \cite{hutchinsonStochasticEstimatorTrace1989,grathwohlFFJORDFreeFormContinuous2018}. In practice, this reduces the computation to Jacobian-vector products, which can be evaluated efficiently by one backpropagation pass. However, this stochastic approximation of the divergence only yields a stochastic approximation of the likelihood.}
Moreover, training CNFs by maximum likelihood requires the use of an adjoint method to differentiate through the trajectory followed along the likelihood integration (\cref{eq:02-instantaneous-change-of-variables-b}), which itself depends on the velocity field and its parameters \cite{chenNeuralOrdinaryDifferential2018}. Despite these disadvantages, CNFs were shown to be remarkably expressive compared to discrete-time NFs and represent a conceptual step toward the highly successful models discussed next—diffusion models and flow matchings—which also benefit from the same underlying dynamical construction.

In the context of generative modeling for physical systems, continuous-time generative models
% —such as CNFs, diffusion models, and flow matching—
offer an additional, non-negligible advantage: they enable the design of probabilistic models whose densities respect the invariances of the problem at hand, such as rotational or translational invariances. To describe this concisely, consider a symmetry group $\mathcal{G}$ leaving the target distribution $\pi$ invariant, i.e. $\forall g \in \mathcal{G}, \pi(g(x)) = \pi(x)$. One can show that using an invariant base measure $\rzero$ together with an equivariant velocity field, i.e. satisfying
\begin{align}
    \label{eq:02-equivariance-condition}
     \forall x \in \X ,  \; \rzero(g(x)) = \rzero(x), \quad \text{and} \quad
     \forall x \in \X ,  \; v_\theta(g(x),t) = g\left(v_\theta(x,t)\right),
\end{align}
yields an invariant marginal probability density $\rho_{\theta,t}$ at all times \cite{kohlerEquivariantFlowsExact2020,garciasatorrasEquivariantNormalizingFlows2021}. Such equivariant architectures can be constructed either explicitly, for instance as gradients of invariant functions \cite{kohlerEquivariantFlowsExact2020}, or by relying on dedicated libraries \cite{satorrasEquivariantGraphNeural2021,geigerE3nnEuclideanNeural2022}.

By contrast, obtaining invariant densities with discrete-time normalizing flows is considerably more difficult. In principle, one could combine an invariant base distribution with an equivariant transport map, but this would require layers that are simultaneously invertible, equipped with a tractable Jacobian determinant, and equivariant. The coupling-layer constructions typically used to satisfy the first two requirements make it non-trivial to enforce the third.\footnote{One possible workaround is a lifting strategy in which the state space is augmented, allowing the construction of equivariant architectures through finite compositions of coupling layers \cite{midgleySEEquivariantAugmented2023a,schebekScalableBoltzmannGenerators2025}. This approach, however, comes at the cost of losing access to the exact likelihood on the original space.}

Diffusion models and stochastic interpolants, which we review next, inherit the flexibility and symmetry compatibility of CNFs. Thanks to training objectives that differ from direct likelihood maximization, they additionally sidestep the computational burden associated with integrating the instantaneous change-of-variables formula.

\subsection{Diffusion models}
\label{subsec:02-diffusion-models}

CNFs trained by maximum likelihood encourage the velocity field to match the model final time marginal $\rho_{\theta,T}$ to the data target distribution, without imposing constraints on the intermediate $\rho_{\theta,t}$ distributions. In contrast, diffusion models (DMs)
prescribe a noising-denoising path for the intermediate distributions to follow. For simplicity, we consider here the case $\X=\R^d$ and an Ornstein-Uhlenbeck (OU) noising process defined by the stochastic differential equation (SDE)
\begin{align}
% \begin{cases}
\label{eq:02-noising-sde}
\dd \tilde X_\tau = - \tilde X_\tau \dd \tau + \sqrt{2}\, \dd \tilde W_\tau,\quad
\tilde X_0 \sim \pi,
% \end{cases}
\end{align}
where $\tilde W_\tau$ is a standard Brownian motion and $\pi$ denotes the unknown target distribution from which the training set is sampled. In what follows, the marginal distribution of the noising process at time $\tau$ is denoted $\pi_\tau$. The forward OU process — equivalently, an overdamped Langevin equation with quadratic confining potential — admits the Gaussian $\N(0_d,I_d)$ as unique invariant measure and converges to it at an exponential rate. Consequently, one can assume that $\pi_T \approx \N(0_d,I_d)$ for $T \gg 1$.

Anderson \cite{anderson82} established that the process defined in backward time $t = T - \tau$ by
\begin{align}
\label{eq:02-denoising-sde}
% \begin{cases}
\dd X_t = \big(X_t + 2 \nabla \log \pi_{T-t}(X_t)\big)\, \dd t + \sqrt{2}\, \dd W_t,\quad 
X_0 \sim \pi_T \approx \N(0_d,I_d),
% \end{cases}
\end{align}
admits $\pi_{T-t}$ as its marginal density at time $t$, which can be readily verified using the Fokker-Planck equation. In other words, the process defined in \cref{eq:02-denoising-sde} has the same marginal distributions as the noising process but evolves backward in time, and is therefore referred to as the denoising process. 

As originally proposed by \cite{sohl-dicksteinDeepUnsupervisedLearning2015a}, a generative model can be constructed 
% from this noising-denoising framework 
by running the backward denoising process starting from samples drawn from the limiting distribution of the noising process — here the standard Gaussian $\rzero=\N(0_d,I_d)$ — playing the role of the base distribution by analogy with normalizing flows.
However, the denoising process defined in \cref{eq:02-denoising-sde} involves the gradient of the logarithm of the time-marginal density, known as the \emph{score}. Let's examine the score closely.

The conditional distribution of the noising process at time $\tau$ given the initial state $x_0$, denoted $p_{\tau \mid 0}(\cdot \mid x_0)$, is Gaussian, yielding the time marginal
\begin{align}
    \pi_\tau(x) & = \int \dd x_0  \; p_{\tau \vert 0}(x|x_0) \, \pi(x_0) 
     = \int \dd x_0 \; \N(x; \, x_0e^{-\tau}, \, (1-e^{-2\tau})I_{d}) \, \pi(x_0),
\end{align}
where $\N(\, \cdot \,; m, \Sigma)$ denotes the density of a Gaussian distribution with mean $m$ and covariance $\Sigma$.
From here, the score can be written as a conditional expectation over the initial state, using Bayes formula to recognize the conditional measure:
\begin{align}
    \label{eq:02-score-conditional-expectation}
    \nabla \log \pi_\tau(x) & = - \frac{1}{\pi_\tau(x)}\int \dd x_0 \; \frac{x - x_0 e^{-\tau}}{1 - e^{-2\tau}} \, \N(x; \, x_0e^{-\tau}; \, (1-e^{-2\tau})I_{d}) \,\pi(x_0) \notag \\
    & = - \int \dd x_0 \; \frac{x - x_0 e^{-\tau}}{1 - e^{-2\tau}} \, \frac{ p_{\tau \mid 0}(x \mid x_0) \pi(x_0) }{\pi_\tau(x)} \notag \\
    & = - \E_{x_0}\left[ \left.\frac{x - x_0 e^{-\tau}}{1 - e^{-2\tau}} \right\vert \tilde X_\tau = x  \right]. 
\end{align}
As a result, the score is doubly intractable: in generative modeling tasks, the target distribution $\pi$ is unknown and only accessible through an empirical training dataset. Additionally, even in sampling applications when $\pi$ is known, the conditional expectation remains a high-dimensional integral without a closed-form expression.

In generative modeling applications, however, the score can be regressed directly from training data, typically using a neural network parametrization $s_\theta : \X \times [0,T] \to \X$. Its expression as a conditional expectation \eqref{eq:02-score-conditional-expectation} motivates the denoising score matching objective introduced in \cite{songScoreBasedGenerativeModeling2020,hoDenoisingDiffusionProbabilistic2020}, which takes the form of a mean-squared error at a fixed time $\tau$,
\begin{align}
    \label{eq:02-denoising-score-matching-loss}
\min_\theta \;
\E_{x_0} \left[ \rule{0pt}{2.2em}
\E_{x \mid x_0} \left[ \rule{0pt}{2em}
 \left\Vert \rule{0pt}{1.5em} s_\theta(x, \tau) + \frac{x - x_0 e^{-\tau}}{1 - e^{-2\tau}} \right\Vert^2
\; 
% \right| \; \tilde X_\tau = x
\right]
\right] \, ,
\end{align}
where the reversed order of the expectations compared to \cref{eq:02-score-conditional-expectation} stems from the law of total expectation and reflects the practical implementation of the loss.
An empirical approximation of this loss is constructed using samples $x_0$ from the training set and samples of the form $\tilde X_\tau = x_0 e^{-\tau} + \sqrt{1-e^{-2\tau}} \, \xi$ with $\xi \sim \N(0_d,I_d)$ 
drawn from the conditional distribution of the noising process $p_{\tau \mid 0}(\cdot \mid x_0)$. A weighted integral of this loss over $\tau \in [0,T]$ is commonly employed in practice, enabling the simultaneous training of the score across all time marginals.

Presented here with the canonical illustration of the OU process, this noising-denoising construction admits many natural generalizations. Different forward diffusions can be considered, for instance the variance-exploding SDE, which relies on a purely diffusive process without drift, in contrast to the variance-preserving OU dynamics. We refer in particular to \cite{karrasElucidatingDesignSpace2022}, which clarifies several important design choices underlying the practical success of diffusion models, including noise scheduling, discretization schemes and neural architectures used to parametrize the score.

Beyond modifications of the forward diffusion, alternative noising processes can also be employed. In particular, Schrödinger bridges — which achieve convergence in finite time and relax the assumption of a Gaussian base distribution — can be leveraged within the same noising-denoising paradigm for generative modelling \cite{debortoliDiffusionSchrodingerBridge2021}.
Motivated by this idea of finite-time constructions and increased flexibility, another class of models — stochastic interpolants or flow matchings — has recently attracted attention in the generative modelling community. They offer a flexible framework that interpolates between prescribed endpoint distributions and will be the focus of the next section.

\subsection{Stochastic interpolants a.k.a. flow matchings}
\label{subsec:02-stochastic-interpolants}

The idea of constraining the path of intermediate distributions, has inspired the development of another class of transport-based continuous-time generative models, proposed independently in \cite{albergoBuildingNormalizingFlows2023,liuFlowStraightFast2023,lipmanFlowMatchingGenerative2023}. Here, we adopt the most general formulation of stochastic interpolants (SIs) introduced in \cite{albergoStochasticInterpolantsUnifying2023}.

While the base distribution of diffusion models is necessarily given by the limiting distribution of a noising process, the base distribution of stochastic interpolants (SIs), denoted $\rzero$ in the following, can be any distribution that is easy to sample from. The bridge of distributions is no longer defined through an SDE, but, instead, constructed using an interpolation function $I_t : [0,1] \times \X \times \X \to \X$ together with a positive scalar function $\gamma : [0,1] \to \R^+$,
\begin{align}
\label{eq:02-interpolation-process}
X_t = I_t(X_0, X_1) + \gamma(t) Z, \quad X_0 \sim \rzero, \quad X_1 \sim \pi, \quad Z \sim \N(0_d,I_d).
\end{align}
The interpolation function satisfies $I_0(x_0, x_1)=x_0$ and $I_1(x_0, x_1)=x_1$, the noise amplitude vanishes at the extremities, $\gamma(0)=\gamma(1)=0$, and suitable regularity conditions are required on both functions (see \cite{albergoStochasticInterpolantsUnifying2023}, Definition 2.1).

With this construction, the marginal distribution of $X_t$ at time $t \in [0,1]$ is given by
\begin{align}
\label{eq:02-interpolation-bridge}
\rho_t(x) = \int \dd x_0 \, \dd x_1 \, \dd z \,
\rzero(x_0) \, \pi(x_1) \, \N(z; 0, I_d) \,\delta \left(x - I_t(x_0, x_1) - \gamma(t) z\right),
\end{align}
where $\pi$ is the target data distribution and $\delta$ denotes the Dirac distribution. Although this marginal density is generally intractable, sampling from the interpolation process is straightforward, as it only requires sampling from the base distribution, the target data distribution (of which the training data are samples), and the Gaussian noise.

To construct a generative model, one observes that the velocity field defined as the conditional expectation
\begin{align}
v^*(x,t) = \E_{x_0,x_1,z} \left[ \left. \partial_t I_t(x_0, x_1) + \dot \gamma z \, \right\vert \, X_t = x \right],
\end{align}
where $\dot \gamma$ denotes the time derivative of $\gamma$, satisfies the continuity equation associated with the interpolating path of distributions \cref{eq:02-interpolation-bridge}:
\begin{align}
\frac{\partial \rho_t(x)}{\partial t} + \nabla \cdot \Big(\rho_t(x) v^*(x, t)\Big) = 0 \, .
\end{align}
As a consequence, the deterministic process $(X_t^*)_{t\in[0,1]}$ defined by the ODE
\begin{align}
\frac{\dd X_t^*}{\dd t } = v^*(X^*_t, t), \quad X^*_0 \sim \rzero,
\end{align}
admits the same time marginal distributions $\rho_t$ as the interpolation process at all times. This observation enables the construction of a generative model by learning a neural network approximation $v_\theta : \X \times [0,1] \to \X$ of $v^*$.

As for the score of diffusion models, the conditional expectation form of the target velocity field leads to a mean-squared regression loss at a fixed time $t$,
\begin{align}
    \label{eq:02-SI-loss}
\min_\theta \E_{x_0,x_1,z} \left[ \left. \left\Vert v_\theta(X_t, t) - \big(\partial_t I_t(x_0, x_1) + \dot \gamma z\big) \right\Vert^2 \right\vert X_t = I_t(x_0, x_1) +  \gamma(t) z \right].
\end{align}
In practice, an empirical approximation of this loss is obtained by drawing samples $x_1$ from the training set, samples $x_0$ from the base distribution, Gaussian noise samples $z$, and averaging the mean squared error over (uniform or weighted) samples of time $t$ in $[0,1]$. 

Once the velocity field has been learned, new samples are generated by integrating the ODE driven by $v_\theta$ from samples of the base distribution $\rzero$ up to time $t=1$.
This sampling procedure boils down to the sampling procedure of CNFs, with the key difference that the velocity field is explicitly trained to match a prescribed interpolation path of distributions. Avoiding the maximum likelihood training of CNFs, this approach is significantly more efficient computationally.
The generation process can also be generalized to a stochastic differential equation, at the cost of additionally regressing the score of the time marginals (see \cite{albergoBuildingNormalizingFlows2023}, Corollary 2.9). A commonly used choice for the interpolation process is the noiseless linear interpolation $I_t(x_0, x_1) = (1-t)x_0 + t x_1$, as proposed in \cite{liuFlowStraightFast2023,lipmanFlowMatchingGenerative2023} who introduced the denomination of flow matching for this particular case.

\pbreak

With this, we reviewed the main generative models that will be used in the sampling methods described in the rest of the text. Autoregressive models and normalizing flows both offer direct access to the likelihood of generated samples and can be trained by maximum likelihood. Yet, they are limited in expressivity and difficult to make compatible with symmetries. Continuous-time generative models, on the other hand, are more expressive and can be made compatible with symmetries, but computing their likelihood is challenging. 
Relying on the notations and mathematical tools introduced in this section, we now turn to the application of these different models to sampling problems.

\pagebreak
\section{Exact samplers with exact likelihood generative models}
\label{sec:03-exact_samplers}
Normalizing flows (NF) and autoregressive models (ARM) were the first generative models to be repurposed from generative modelling to the task of sampling. As we will see below, their tractable likelihood allows to use them as helpers in a variety of Monte Carlo scheme in the place of traditional simple probabilistic models such as Gaussians or factorized distributions. 
In all described approaches, the first step is to train a generative model to approximate the target distribution of interest $\pi$ on the state space $\X$. While this training is not trivial given the absence of direct samples from the target a priori, we postpone the discussion on training strategies to the next section. \emph{In this section, we assume access to an ARM or a NF defining a distribution or probability density $\rtheta$ (respectively given by \cref{eq:02-ARM-factorization} and \cref{eq:02-nf-change-of-variable}) which, loosely speaking, is close to $\pi$. It is cheap to evaluate the likelihood $\rtheta$ of these models and to sample from them. We also assume that $\pi(x)$ is easy to compute up to a normalization constant, for any $x \in \X$.}

Note that even if the agreement between $\rho_\theta$ and $\pi$ is imperfect — as is inevitably the case in practice due to approximation and optimization errors — the samplers described below remain exact. In other words, they generate samples from $\pi$ in the limit of large sample sizes or many iterations. This makes the present application of deep learning to scientific computing rather distinctive: the unavoidable imperfections introduced during the learning stage do not necessarily translate into biased results, provided that the subsequent sampling procedure is properly designed.

Below, we first describe a set of simple approaches originally proposed to leverage ARMs and NFs for sampling (\cref{subsec:03-initial-samplers}), and then discuss the advantages and limitations of these approaches in a systematic benchmark (\cref{subsec:03-comparison}). We then review more recent approaches looking to push further the abilities of samplers powered with exact likelihood models (\cref{subsec:03-tempering,subsec:03-coarse-graining}). Part of our discussion will only apply to NFs only as the samplers exploit their transport construction.

\subsection{Initial samplers powered by exact likelihood generative models}
\label{subsec:03-initial-samplers}
\subsubsection{Importance sampling}
\label{subsubsec:03-is}

\paragraph{Elementary algorithm}
A simple way to leverage ARMs and NFs for sampling is to use them as proposal distributions in importance sampling (IS). For a target distribution $\pi$, known up to a normalization constant, the IS estimation of the expectation of an observable $f:\X \to \R$ follows the steps:
% \vspace{0.3em}

\newlength{\savedparskip}
\setlength{\savedparskip}{\parskip}

\begin{minipage}{\textwidth}
\captionof{algorithm}{\textbf{Neural importance sampler (Neural-IS)}} 
\label{alg:03-neural-IS}
\vspace{-0.4em}
\begin{algorithmic}
    \For {$i = 1 \ldots N$}
        \State Sample the proposal distribution $x_i \sim \rtheta(\cdot)$,
        \State Compute weight $w_i = \frac{\pi(x_i)}{\rtheta(x_i)}\,$ (up to a multiplicative constant),
    \EndFor 
    \For {$i = 1 \ldots N$}
        \State Compute normalized weight $\hat w_i = w_i / \sum_{j=1}^N w_j$,
    \EndFor
    \Return self-normalized IS estimator $\hat{f}_{\text{SNIS}} = \sum_{i=1}^N \hat w_i f(x_i)$    
\end{algorithmic} 
\setlength{\parskip}{\savedparskip}
\vspace{0.5em}
\end{minipage}

A practical IS proposal needs (i) to be straightforward to sample from to obtain an extensive set of $N$ i.i.d. samples easily and (ii) to have a tractable probability density function to compute weights, both conditions being satisfied by NFs. 

Note that if the normalization constant of $\pi$ is unknown, the IS weights can only be computed up to a multiplicative constant which however cancels in the self-normalized weights and hence has no impact on the self-normalized IS estimator $\hat{f}_{\text{SNIS}}$ returned by the algorithm. Conversely, were the normalization constant of $\pi$ known, one could use the plain IS estimator $\hat{f}_{\text{IS}} = \frac{1}{N} \sum_{i=1}^N w_i f(x_i)$ which is unbiased, unlike the self-normalized version.

Under mild conditions on the proposal distribution, 
$\hat{f}_{\text{SNIS}}$ is a consistent estimator of $\E_{\pi}[f]$ in the sense that it converges (at least in probability) to $\E_{\pi}[f]$ as $N \to \infty$.
However, at finite $N$, the agreement of the proposal distribution $\rtheta$ with the target $\pi$ directly impacts the quality of the IS estimator. In particular, the variance of the IS weights is minimized when $\rtheta$ is equal to $\pi$, yielding normalized weights uniformly equal to $1/N$. 
This observation justifies the relevance of leveraging recent progress in deep generative models, in the form of ARMs and NFs, to build accurate proposals in place of usual simple tractable probabilistic models (e.g. Gaussian or factorized). In turn, the quality of the IS estimator is directly related to the quality of the generative model's approximation of the target distribution.
Pioneering works in this direction include
\cite{mullerNeuralImportanceSampling2019a,noeBoltzmannGeneratorsSampling2019,nicoliAsymptoticallyUnbiasedEstimation2020,nicoliEstimationThermodynamicObservables2021}.

\paragraph{Free energy computations} Building on similar ideas of IS, one can also use exact likelihood models to estimate normalization constants, or free energies in the language of statistical mechanics. The idea is to use the generative model as a reference state of known normalization constant and estimate its ratio with normalization constant of the target $\pi$ using either Target Free Energy Perturbation (TFEP) \cite{jarzynskiTargetedFreeEnergy2002} or Bennett Acceptance Ratio (BAR) \cite{BENNETT1976245,HahnUsingBijectiveMaps2009}. This idea was simultaneously proposed with Bayesian inference in mind \cite{jiaNormalizingConstantEstimation2020}, and for applications to atomic solids \cite{wirnsbergerTargetedFreeEnergy2020,wirnsbergerNormalizingFlowsAtomic2022,wirnsbergerEstimatingGibbsFree2023,schebekScalableBoltzmannGenerators2025}. Computations of protein-lingand binding free energies with NFs were also explored in \cite{dinhDensityEstimationUsing2017a}, as well as computations of conformal free energy differences in \cite{molina-tabordaActiveLearningBoltzmann2024a,olehnovicsAssessingAccuracyEfficiency2024}.

By convention, the normalization constant of a NF or an ARM is 1 due to writing $\rtheta$ directly in its normalized form. For $\pi(x) = e^{-U(x)}/\Z_\pi$, the TFEP estimator of its free energy $F_\pi = -\log \Z_\pi$ is build upon the equality
\begin{align}
\Z_\pi = \E_{\rtheta}\left[e^{-U(x)} / \rtheta(x)\right] \quad \Rightarrow F_\pi = -\log \E_{\rtheta}\left[e^{-U(x)} / \rtheta(x)\right] 
\end{align}
and reads
\begin{align}
\label{eq:03-tfep-estimator}
\hat F_\pi^{\text{TFEP}} = -\log \hat \Z_\pi^{\text{TFEP}}=  -\log \left(\frac{1}{N} \sum_{i=1}^N \frac{e^{-U(x_i)}}{\rtheta(x_i)}\right), \quad \text{with } x_i \sim \rtheta(\cdot) \text{ for } i=1,\ldots,N \; , 
\end{align}
where one recognizes the IS weights $w_i$ in the summation. This estimator is easy to implement, requiring mere samples from the flow and yields an unbiased estimator $\hat \Z_\pi^{\text{TFEP}}$ of $\Z_\pi$ and a consistent estimator $\hat F_\pi^{\text{TFEP}}$ of $F_\pi$ as $N \to \infty$. However, as for IS, it can suffer from large variance at finite $N$ if $\rtheta$ is not close enough to $\pi$. 

The BAR method improves over TFEP in cases where samples from both $\rtheta$ and $\pi$ are available \cite{BENNETT1976245}. Given two sets of samples $\{x_i\}_{i=1}^N$ from $\rtheta$ and $\{x_j\}_{j=1}^M$ from $\pi$, the BAR estimator can be interpreted as the maximum likelihood estimate of the free energy difference obtained by viewing the problem as a binary classification task. The main ideas of the derivation (described in detail in \cref{app:bar-derivation}) are: given a sample $x$, one asks whether it was drawn from $\rtheta$ or from $\pi$. Using Bayes’ rule, the posterior probability that $x$ originates from $\rtheta$ can be written as a logistic function of the energy difference between the two models, involving the unknown free energy difference $F_\pi$. Maximizing the corresponding likelihood over the two sets of samples leads to a self-consistent equation whose solution defines the BAR estimator \cite{shirtsEquilibriumFreeEnergies2003}. 
Namely, the estimator $\hat F_\pi^{\text{BAR}}$ is defined as the solution of the implicit equation
\begin{align}
\label{eq:03-bar-implicit equation}
\sum_{i=1}^N \frac{1}{1 + \frac{N}{M}e^{U(x_i) + \log \rtheta(x_i) - \hat F_\pi^{\text{BAR}}}} = \sum_{j=1}^M \frac{1}{1 + \frac{M}{N}e^{-U(x_j) - \log \rtheta(x_j) + \hat F_\pi^{\text{BAR}}}} \; .
\end{align}
This estimator is known to be asymptotically optimal, meaning that, among estimators using samples from both distributions, it has minimal variance in the limit of infinite samples. A derivation specifically in the case of bijective mapping (relevant for NFs) is given in \cite{HahnUsingBijectiveMaps2009}. If BAR improves on TFEP, it strictly requires samples from the target distribution $\pi$, which are not available a priori in the context discussed here. If some references use approximate samples obtained from molecular dynamics \cite{dingDeepBARFastExact2021}, we note that provided an exact likelihood model approximating $\pi$ is available, the NF can also be used to converge an MCMC to generate samples from $\pi$ to be used in BAR \cite{molina-tabordaActiveLearningBoltzmann2024a}. The description of NF assisted MCMC schemes is precisely the topic of the next section.

\subsubsection{Independent proposal Markov chains}
\label{subsubsec:03-indep-proposal}

MCMCs can also greatly benefit from an exact likelihood model trained to approximate the target distribution $\pi$. A simple way to do so is to use the generative model as an independent proposal distribution in a Metropolis-Hastings (MH) algorithm \cite{albergoFlowbasedGenerativeModels2019,
noeBoltzmannGeneratorsSampling2019, mcnaughtonBoostingMonteCarlo2020,nicoliAsymptoticallyUnbiasedEstimation2020,Gabrie2021a}. The resulting algorithm reads:
% \vspace{0.3em}

\begin{minipage}{\textwidth}
\captionof{algorithm}{\textbf{flow/ARM independent Metropolis Hastings (flow-IMH/ARM-IMH)}}
\label{alg:03-flow-IMH}
\vspace{-0.4em}
\begin{algorithmic}
    \State Initialize $x^{(0)}$,
    \For {$k = 1 \ldots K$}
        \State Sample proposal $x' \sim \rtheta(\cdot)$,
        \State Compute acceptance probability 
        $\displaystyle \alpha = \min\left(1 \,, \; \frac{\pi(x')}{\pi(x^{(k-1)}) } \times \frac{\rtheta(x^{(k-1)})}{\rtheta(x')}\right) \, ,$
        \State Set $x^{(k)} = x'$ with probability $\alpha$, else $x^{(k)} = x^{(k-1)}$,
    \EndFor
    \Return samples $\{x^{(k)}\}_{k=1}^K$
\end{algorithmic}
\setlength{\parskip}{\savedparskip}
\vspace{0.5em}
\end{minipage}

As in IS, observe that the proposal distribution of a MH algorithm needs to be (i) easy to sample from and (ii) its density needs to be tractable to evaluate the acceptance probability, both conditions being satisfied by NFs and ARMs.

The resulting independent proposal MCMC is asymptotically unbiased regardless of the quality of the generative model approximation, meaning that the distribution of $x^{(K)}$ converges to $\pi$ as $K \to \infty$. However, at finite $K$, the acceptance rate of the algorithm directly depends on the agreement of $\rtheta$ with $\pi$, being equal to $1$ when $\rtheta = \pi$ and dropping to zero when $\rtheta$ is a poor approximation of $\pi$. A high acceptance rate implies an extremely efficient sampler here as the proposal are \emph{global moves}, uncorrelated with the current state of the chain. Attempting global moves across the state space is made possible by the flexibility of ARMs and NFs to approximate complex distributions, and represents a key advantage of using deep generative models as proposal distributions in place of local proposals such as random-walk or Langevin dynamics. In particular, global moves can help escape from metastable states in multimodal distributions, a situation where local MCMC proposals often fail.

\paragraph{Local-global Markov Chains} Variants around the independent Metropolis-Hastings presented above can also be of interest. For instance, one can consider a hybrid scheme alternating local MCMC moves (e.g. Langevin dynamics or MH with Gaussian jump proposals, see \cref{app:local-samplers} for reminders on the local samplers used throughout this review) and global NF-based proposals \cite{Gabrie2021a,samsonov2022localglobal}. This is particularly useful when the generative model misrepresents some part of the target distribution, as the local moves can help improve exploration in such regions. The advantage of alternating local and global moves was theoretically proven for simple statistical physics models \cite{delbonoPerformanceMachinelearningassistedMonte2025}. 

\paragraph{Multi-proposal version} Another interesting variant considers more than one proposal per iteration \cite{samsonov2022localglobal}, following the iterated sampling importance resampling (iSIR) scheme \cite{andrieuParticleMarkovChain2010a}. Given the cheap sampling cost of NFs, one can consider $M>1$ proposals $\{x'_i\}_{i=1}^M$ sampled from $\rtheta$ at each iteration of the MCMC yielding the following algorithm:

\begin{minipage}{\textwidth}
\captionof{algorithm}{\textbf{flow iterated sampling importance resampling (flow-iSIR)}}
\label{alg:03-flow-iSIR}
\vspace{-0.4em}
\begin{algorithmic}
    \State Initialize $x^{(0)}$,
    \For {$k = 1 \ldots K$}
        \State Set $x'_0 = x^{(k-1)}$,
        \For {$i = 1 \ldots M$}
            \State Sample proposal $x'_i \sim \rtheta(\cdot)$,
        \EndFor
        \For {$i = 0 \ldots M$}
            \State Compute weight $w_i = \frac{\pi(x'_i)}{\rtheta(x'_i)}$,
        \EndFor
        \For {$i = 0 \ldots M$}
            \State Compute normalized weight $\hat w_i = w_i / \sum_{j=0}^M w_j$,
        \EndFor
        \State Sample index $i$ from the categorical distribution with probabilities $\{\hat w_j\}_{j=0}^M$,
        \State Set $x^{(k)} = x'_i$,
    \EndFor
    \Return samples $\{x^{(k)}\}_{k=1}^K$
\end{algorithmic}
\setlength{\parskip}{\savedparskip}
\vspace{0.5em}
\end{minipage}

This approach was proposed using NFs but can readily be adapted for ARMs.
By considering multiple proposals, this scheme increases the chance to propose a state in high probability regions of the target distribution, improving the mixing of the chain per iteration. However, the computational cost per iteration also increases with $M$. As a result, a large $M$ is typically advantageous if parallel computing resources are available, even if the overall computational efficiency of the sampler does not necessarily improve with $M$.

\subsubsection{Reparametrization through transport maps}
\label{subsubsec:03-neutra-MCMC}
Yet another way to leverage NFs for sampling is to use them as reparametrization maps to simplify the geometry of the target distribution \cite{parnoTransportMapAccelerated2018,hoffmanNeuTralizingBadGeometry2019,noeBoltzmannGeneratorsSampling2019,cabezasTransportEllipticalSlice2023,schonleOptimizingMarkovChain2023}. Here, we reference this set of methods as neutra-MCMCs following \cite{hoffmanNeuTralizingBadGeometry2019} play of words on neural transport maps "NeuTra-lizing" the target's bad geometry. As this approach relies on the transport construction of NFs, it cannot be directly applied to ARMs.

Recall that a NF is built from a base distribution $\rzero$ and a transport map $T_\theta$. Once $T_\theta$ has been trained such that $\rtheta$ (given by \cref{eq:02-nf-change-of-variable}) approximates the target distribution $\pi$ is available, one can consider the pullback distribution of $\pi$ through $T_\theta$ defined as
\begin{align}
    \pi_{T_\theta}(z) = \pi(T_\theta(z))  \left| \det \nabla_z T_\theta(z) \right| \quad \text{for } z \in \X \; .
    \label{eq:03-pullback}
\end{align}
By construction, if $\rtheta$ is a good approximation of $\pi$, then the pullback distribution $\pi_{T_\theta}$ is close to the base distribution $\rzero$, which is usually chosen to be a distribution with easy geometry (e.g. standard Gaussian). Sampling from $\pi$ can therefore be achieved by first sampling from the simpler distribution $\pi_{T_\theta}$ and then pushing forward the samples through $T_\theta$. For instance, if indeed $\rzero=\N(0, I_d)$, following the intuition that $\pi_{T_\theta} \approx \rzero$, a natural choice is to resort to Langevin dynamics for sampling in the latent space, either with a Metropolis accept reject (MALA for Metropolis adjusted Langevin algorithm) or without (ULA for unadjusted Langevin algorithm).\footnote{See \cref{app:local-samplers} for reminders on these common local samplers.}
However, all other types of samplers can be employed: \cite{hoffmanNeuTralizingBadGeometry2019} examined Hamiltonian Monte Carlo (HMC), \cite{cabezasTransportEllipticalSlice2023} elliptical slice sampling and \cite{schonleOptimizingMarkovChain2023} Metropolis-within-Gibbs sampling. Provided an unbiased sampler is used to target $\rzero=\N(0, I_d)$, transporting its output through $T_\theta$ provides unbiased samples from $\pi$.
Sketching an algorithm to summarize the overall method yields:

\begin{minipage}{\textwidth}
\captionof{algorithm}{\textbf{neutra-MCMC}}
\label{alg:03-neutra-MCMC}
\vspace{-0.4em}
\begin{algorithmic}
    \State Initialize $z^{(0)}$,
    \For {$k = 1 \ldots K$}
        \State $z^{(k+1)} \sim M_{T_\theta, \pi}(\cdot|z^{(k)})$ with $M_{T_\theta, \pi}$ Markov kernel leaving $\pi_{T_\theta}$ invariant,
        \State $x^{(k+1)} = T_\theta(z^{(k+1)})$
    \EndFor
    \Return samples $\{x^{(k)}\}_{k=1}^K$
\end{algorithmic}
\setlength{\parskip}{\savedparskip}
\vspace{0.5em}
\end{minipage}

Note that Markov kernels $ M_{T_\theta, \pi}(\cdot|z^{(k)})$ based on gradient information of the target distribution --such as MALA, ULA and HMC-- require computing gradients of $\pi_{T_\theta}$ at each iteration of the algorithm. This involves automatic differentiation through $T_\theta$ making each update significantly more costly than when directly targeting $\pi$ and hence the overall efficiency of the sampler is not guaranteed to improve. However, the advantage of neutra-MCMCs is that they can be more robust to ill-conditioning of the target distribution than standard MCMC methods, as we will discuss in \cref{subsec:03-comparison}.

\subsection{A first comparison of NF powered samplers}
\label{subsec:03-comparison}
Faced with the variety of ways of easing sampling leveraging a transport map $T_\theta$ sending $\rzero$ close to the target $\pi$, a comparison is in order. A detailed numerical benchmark was conducted in \cite{greniouxSampling2023}, considering different sources of difficulties in sampling, namely ill-conditioning, multimodality and high-dimensionality. We rediscuss the results of this study in the following paragraphs.

\paragraph{Neutra-MCMCs are the most robust to ill-conditioning} If regions of high-probability of the target have significantly different scales in different directions of the state space, the target distribution is said to be ill-conditioned and this typically causes trouble to the convergence of MCMC based samplers. To assess the ability of the different flow-based approaches to alleviate this issue, one needs to also consider the quality of the training of $T_\theta$. For a perfectly trained flow $\rtheta = \pi$, and IS and IMH produce i.i.d samples from $\pi$ and are therefore optimal. The picture becomes more complicated if a residual mismatch persists between $\rtheta$ and $\pi$. To systematically assess the effect of an imperfect learning on sampling an ill-conditioned target, the following experiment was conducted. 

\begin{figure*}[t]
    \begin{subfigure}{0.40\linewidth}
        \centering
        \raisebox{-2cm}{\includegraphics[width=\linewidth]{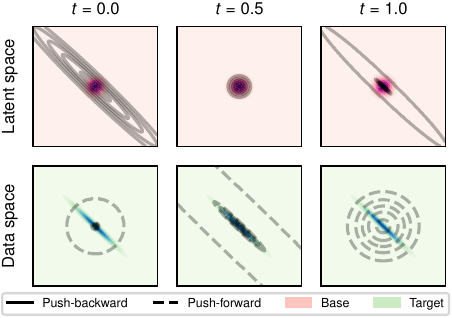}}
        \label{fig:3flows:t}
    \end{subfigure}
    \begin{subfigure}{0.55\linewidth}
        \centering
        \raisebox{-2cm}{\includegraphics[width=\linewidth]{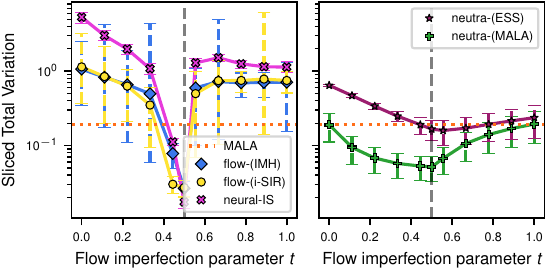}}
        \label{fig:3flows:sliced_tv_dim_128}
    \end{subfigure}
    \caption{\textbf{(Left)} \textbf{Pullback $\pi_{T_t}$ and pushforward
    $\rho_t$ as a function of the flow imperfection parameter $t$.}
    \textbf{(Right) Sliced TV distances to target of NF-powered Monte Carlo samples compared as a function of the quality of the flow $t$}, using 256 chains of length 1400 initialized with draws from NF with $T_t$. neural-IS~was evaluated with 14000 samples. Results were qualitatively unchanged for $d=16,32,64,256$. ESS stands for elliptical sliced-sampling in this figure. Reproduced from \cite{greniouxSampling2023}.}
    \label{fig:03-ill-condition}
\end{figure*}

The target distribution is a high-dimensional Gaussian ($d=128$) with an ill-conditioned covariance (see appendix D.2 of \cite{greniouxSampling2023} for complete experimental setting). An analytical flow is defined with a scalar quality parameter $t\in[0,1]$ as illustrated on the left panel of \cref{fig:03-ill-condition}. For $t=0.5$ the flow is perfect. For $t=0$ the flow density $\rho_t$ is isotropic and over-concentrated, with variance matching the smallest directional variance of the target distribution. Conversely, for $t=1$, $\rho_t$ is isotropic underconcentrated, matching the largest directional variance of the target distribution. When $t\neq0.5$, the flow fails at perfectly preconditioning the pullback $\pi_{T_t}$. Results of the experiment reproduced on the right panel of \cref{fig:03-ill-condition} show that neutra-MCMC methods are more robust to an imperfect map than approaches based on independent proposals (neural-IS, flow-IMH, flow-iSIR) whose performances degrade quickly as the flow training parameter deviates from $t=0.5$. This advantage is however mitigated with the cost per iteration of the method (higher than for independent proposal methods by a factor of 2 to 30 for the experiments reported in \cite{greniouxSampling2023}) and the challenges posed by multimodality for neutra-MCMCs, as discussed next.

\begin{figure}[t]
    \centering
    \includegraphics[width=0.75\linewidth]{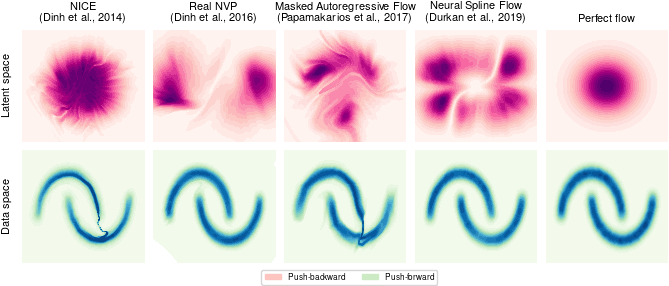}
    \caption{Pullback and pushforward of popular deep normalizing flows architectures targeting the two moons distribution. Reproduced from \cite{greniouxSampling2023}}
    \label{fig:03-pullbacks}
\end{figure}

\paragraph{Neutra-MCMCs struggle to mix between modes} When the target $\pi$ is multimodal, the intuition that the pullback $\pi_{T_\theta}$ has a simplified geometry breaks down. The reason is that common NF architecture based on transporting a standard Gaussian $\rho$ through coupling layers have troubles separating the mass of $\rzero$ into truly disconnected modes and respectively do not generate pullbacks with reunited mass. The difficulty of NFs to match multimodal distribution has been discussed in \cite{cornishRelaxingBijectivityConstraints2020,salmonaCanPushforwardGenerative2022}.
\Cref{fig:03-pullbacks} illustrates well this phenomenon on a 2$d$ example. 

\begin{figure*}
    \centering
    \begin{subfigure}{0.40\linewidth}
        \centering
        \raisebox{-2cm}{\includegraphics[width=\linewidth]{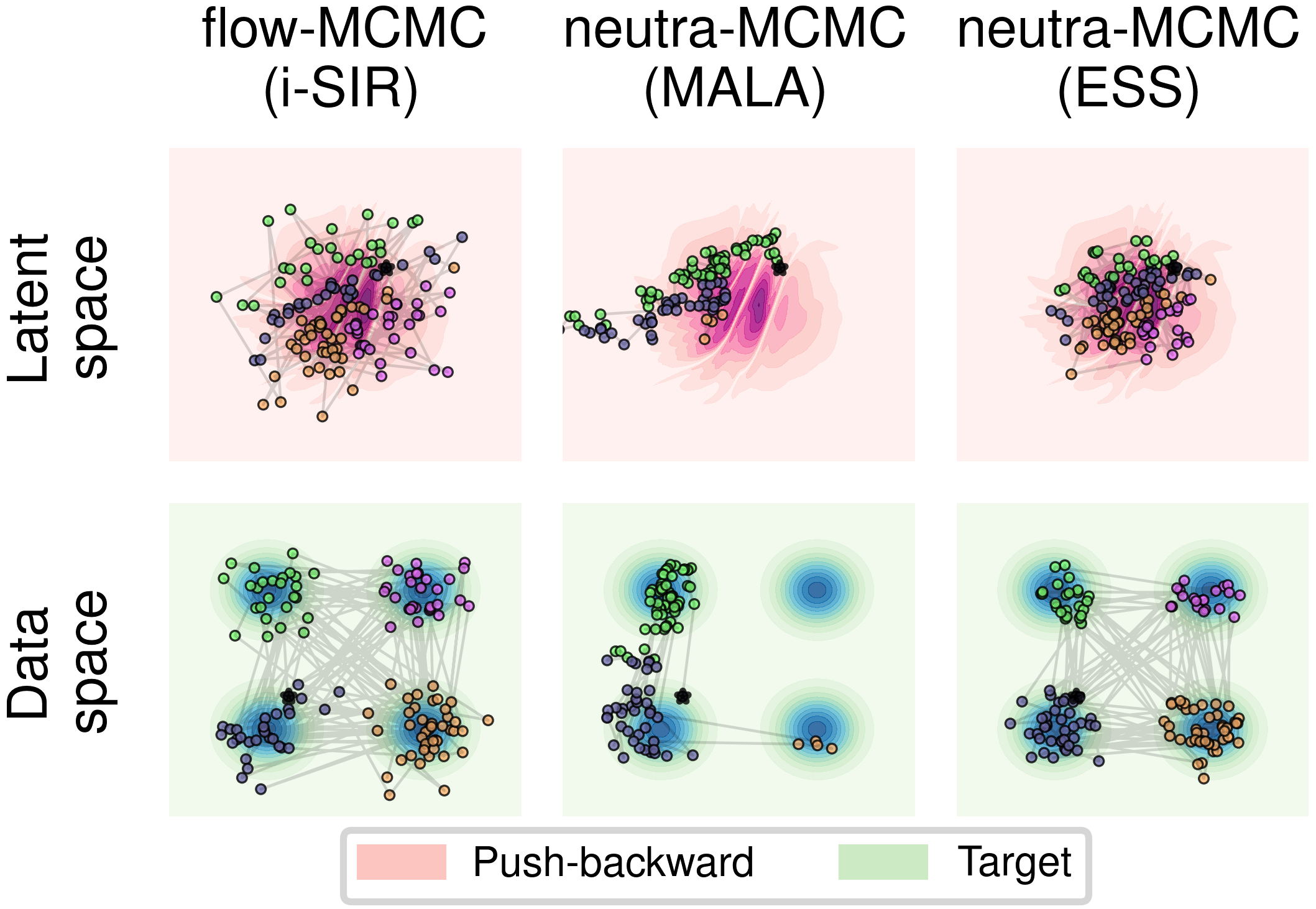}}
        \label{fig:highdim_mog:samples_neutra}
    \end{subfigure}%
    \begin{subfigure}{0.30\linewidth}
        \centering
        \raisebox{-2cm}{\includegraphics[width=\linewidth]{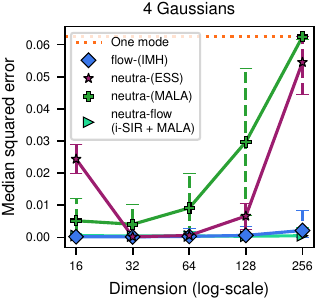}}
        \label{fig:highdim_mog:l2_dist}
    \end{subfigure}%
    \begin{subfigure}{0.30\linewidth}
        \centering
        \raisebox{-2cm}{\includegraphics[width=\linewidth]{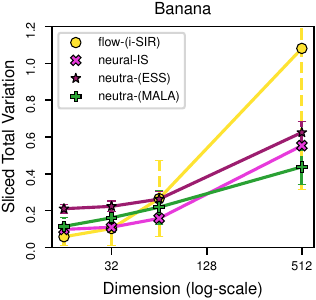}}
        \label{sec:dim_scaling}
    \end{subfigure}
    \caption{\textbf{(Left) Example chains of NF-enhanced walkers with a 2d target mixture of 4 Gaussians.} 
    The 128-step MCMC chain is colored according to the closest mode in the data space (bottom row) with corresponding location in the latent space (top row). The complex geometry of the pullback $\pi_{T_\theta}$ hinders the mixing of local-update algorithms. MALA's step-size was chosen to reach 75\% of acceptance. \textbf{(Middle)} \textbf{Median squared error of the histograms of visited modes of 4 Gaussians per chain against the perfect uniform histogram as a function of dimension.} 512 chains of 1000-steps on average were used. \textbf{(Right) Sliced total variation in sampling the Banana distribution in increasing dimension} using a RealNVP. 128 chains of 1024-steps were used. Reproduced from \cite{greniouxSampling2023}}
    \label{fig:03-multimod-neutra}
\end{figure*}

As a result, neutraMCMCs relying on MCMC kernels struggling with multimodality are not out of trouble by targeting the pullback. In this context, methods relying on independent proposals such as flow-IMH are better off. This can be seen on the results presented on the left and center panel of \cref{fig:03-multimod-neutra} considering the task of sampling from a Gaussian mixture. In particular, as the dimension increases, methods based on reparametrization increasingly fail to discover all the modes of the distribution despite the fact that the pushfoward $\rtheta$ did have significant mass overlap with all the modes. However, this success of independent proposal methods for multimodal distribution is limited by their sensitivity as the dimension further increases, which we examine next. 

\paragraph{Independent proposal methods are most affected by high-dimensionality}
Methods relying on independent proposals, either in the context of importance sampling (neural-IS) or of MCMCs (flow-IMH and flow-iSIR) suffer similarly from the increase of the dimensionality. A classical elementary calculation by \cite{agapiouImportanceSamplingIntrinsic2017} allows to catch the source of the difficulty considering the example case where the considered target and proposal distributions are the product of 1$d$ measures. But before we cover this example, let us define an
important quantity to assess the efficiency of IS and IMH that has not yet been introduced. The effective sample size (ESS) \cite{kishSamplingOrganizationsGroups1965} is defined, for a batch of $N$ samples $\{x_i\}_{i=1}^N$ drawn from the proposal distribution $\rho$, with importance weights $w(x_i) = \pi(x_i)/\rho(x_i)$, as
\begin{align}
    \text{ESS} = \frac{\left(\sum_{i=1}^N w(x_i)\right)^2}{\sum_{i=1}^N w(x_i)^2}.
    \label{eq:03-ess}
\end{align}
If all the weights are equal, meaning that the proposal perfectly matches the target, then $\text{ESS}=N$. Conversely, if one weight dominates all the others, then $\text{ESS} \approx 1$. The ESS therefore provides a measure of the number of independent samples from the target distribution that can be extracted from the weighted samples drawn from the proposal. Note that both unormalized and self-normalized IS weights can be used to compute the ESS, as the normalization constant cancels out in the ratio.

Going back to the example of product of 1$d$ distributions, consider target and proposal distributions defined on $\X=\R^d$ as
\begin{align}
    \pi(x) = \prod_{i=1}^d \pi_{1d}(x_i) \; \text{ and } \; \rho(x) = \prod_{i=1}^d \rho_{1d}(x_i).
\end{align}
Then the importance weight function itself factorizes $w(x)=\prod_{i=1}^d w_{1d}(x_i)$ with $w_{1d}(x_i) = \pi_{1d}(x_i)/\rho_{1d}(x_i)$. In the limit of large $N$, thanks to the independence of all the coordinates, the ESS can be approximated as
\begin{align}
    \text{ESS} \approx \frac{(N\E_\rho[w_{1d}]^{d})^2}{N \E_\rho[w_{1d}^2]^d}  = N \left(\frac{\E_{\rho_{1d}}[w_{1d}]^2}{\E_{\rho_{1d}}[w_{1d}^2]}\right)^d \approx N e^{-d \log \left( \E_{\rho_{1d}}[w_{1d}]^2 / \E_{\rho_{1d}}[w_{1d}^2] \right)} \, 
\end{align}
where the logarithm in the exponent is positive unless the proposal perfectly matches the target in 1$d$ in which case it is zero. This calculation shows that unless a perfect match is achieved, the ESS decays exponentially fast with the dimension $d$. Although this calculation is made under restrictive assumptions, it captures well the curse of dimensionality affecting IS and IMH methods. For instance, note that in statistical mechanics applications of sampling, the target distribution takes the form of a Boltzmann-Gibbs distribution $\pi(x) \propto e^{-U(x)}$ where the potential energy $U(x)$ is often extensive in the number of degrees of freedom, meaning scaling linearly with $d$. In such situations, even small relative errors in the approximation of $U(x)$ by the NF transport map $T_\theta$ can lead to exponentially small ESS as $d$ increases.

The right panel of \cref{fig:03-multimod-neutra} illustrates this curse of dimensionality affecting independent proposal methods, considering the task of sampling from a banana-shaped distribution in increasing dimension. Similar results have been reported by other works from which the systematic study of \cite{deldebbioEfficientModellingTrivializing2021} stands out. This work focuses on sampling from the Boltzmann distribution of a lattice field theory, the 2d $\phi^4$ model, of which the size of the lattice directly controls the dimension of the state space. Additionally, a temperature parameter controls the difficulty of the sampling task, the model going through a phase transition where the correlation length between lattice sites diverges, making the sampling particularly challenging. Despite an extensive hyperparameter search and the use of advanced NF architectures, the authors find that the acceptance rate of flow-IMH samplers decreases invariably as the lattice size increases and as the temperature approaches the critical point.

\paragraph{Where do the elementary NF enhanced samplers stand?} Summarizing the above discussion, we can sketch the state of affairs after initial research efforts on leveraging NFs for sampling. Independent proposal methods such as neural-IS and flow-IMH are well suited for sampling multimodal distributions thanks to their ability to explore globally the state space. However, they suffer from the curse of dimensionality, with performances degrading quickly as the dimension of the state space increases, unless a near perfect flow approximation is available. To provide an order of magnitude, the highest dimension admissible for independent proposals lies somewhere between a hundred to a few thousands depending on the regularity of the target distribution. This range covers many practical applications such as Bayesian inference for medium size models or molecular systems with a few tens of atoms, but falls short for larger applications such as simulating liquids, large biomolecules or statistical mechanics field theories toward the continuum limit.

Conversely, neutra-MCMCs leveraging local samplers in the latent space are more robust to high-dimensional settings and ill-conditioning. However, they struggle to mix between modes when the target distribution is multimodal, as the pullback distribution often retains a complex geometry hindering the convergence of local samplers. 

Taken together, these observations motivate the development of more advanced NF-powered samplers that leverage modern advances in the sampling literature, which has progressed considerably beyond the elementary tools of IS and MH. In the next two sections we review how NF can be combined with the ideas of annealing and of exploiting coarse-grainings to enhance sampling. While some of the methods covered next can be adapted to ARMs, many of them rely on the transport construction of NFs and we therefore focus on NFs in the following sections.

\subsection{Combining NFs with tempering}
\label{subsec:03-tempering}
Recall that annealed based samplers, such as annealed importance sampling (AIS), parallel tempering (PT) and sequential Monte Carlo (SMC), already mentioned in the introduction, rely on a bridging path between a simple reference measure and the target measure.
The philosophy behind combining tempering with NFs is no longer to expect that a single NF transport map directly reaches the target distribution, but instead to delegate part of the sampling effort to tempering-based schemes.
Below, we highlight several key examples, without attempting an exhaustive survey given the rapid pace of recent developments in the literature.

\subsubsection{Adapting the base distribution to the target} 
A first step in the direction of tempering is to adapt the base distribution $\rzero$ of the NF to be closer to the target distribution $\pi$, easing the training of the NF and improving its final accuracy. The choice of $\rzero$ is highly task dependent as it needs to remain relatively easy to sample. The idea was used on statistical physics fields systems using a Gaussian base with 
matching small scale correlations \cite{Gabrie2021}, in sampling liquids using a less dense and/or higher temperature fluid as base
\cite{schebekEfficientMappingPhase2024a,jungNormalizingFlowsEnhanced2024a,corettiLearningMappingsEquilibrium2025}, as well as for sampling crystals and materials using Einstein's harmonic crystal as base \cite{wirnsbergerTargetedFreeEnergy2020,wirnsbergerNormalizingFlowsAtomic2022,schebekScalableBoltzmannGenerators2025}. This is a natural way to incorporate physics knowledge into NF-based sampling schemes that can considerably ease the learning task. It can be further combined with tempering schemes described next.

\subsubsection{Combining NFs with AIS: stochastic normalizing flows}
\cite{wuStochasticNormalizingFlows2020a} introduced a stochastic type of NFs amenable to enhanced sampling using the AIS framework. Instead of relying on a deterministic transport map $T_\theta$, the flow alternates deterministic NF layers and stochastic Markovian layers. The advantage of stochastic NFs lies in the ability of stochastic layers to represent a multimodal target distribution given an unimodal base distribution $\rzero$, which traditional deterministic coupling layers struggle to achieve. 

A stochastic NF with $L$ layers is defined as a sequence of random variables $(x^0, \dots, x^L)$ with $x^0 \sim \rzero$ and $x^L$ being the output of the flow. The transition from layer $\ell-1$ to layer $\ell$ is defined by a transition kernel $q_{\theta_\ell}(x^{\ell}|x^{\ell-1})$ parametrized by $\theta_\ell$. Deterministic NF layers are implemented via a bijective map $T_{\theta_\ell}$ such that $q_{\theta_\ell}(x^{\ell}|x^{\ell-1}) = \delta(x^{\ell} - T_{\theta_\ell}(x^{\ell-1}))$.
Stochastic layers are characterized through a $q_{\theta_\ell}(x^{\ell}|x^{\ell-1})$ with explicit probability density and easy to sample from. 
The overall joint probability density over $(x_0, \dots, x_L)$ and the marginal output distribution read respectively:
\begin{align}
    \rho_\theta(x^0, \dots, x^L) = \rzero(x^0)\prod_{\ell=1}^L q_{\theta_\ell}(x^{\ell}|x^{\ell-1}),  \; \quad \rho_\theta(x^L) = \int \prod_{\ell=0}^{L-1} dx^{\ell} \, \rho_\theta(x^0, \dots, x^L) .    
\end{align}
The marginal output distribution $\rho_\theta(x^L)$ is intractable as it involves marginalizing over all stochastic layers. As such, proposals from the stochastic NF cannot be used directly in Neural-IS (\cref{alg:03-neural-IS}) or flow-IMH (\cref{alg:03-flow-IMH}) to estimate expectations under the target distribution $\pi$. Resorting instead to AIS-like scheme, a set of backward transition kernels needs to be introduced. The backward kernels, with parameters $\tilde \theta_\ell$, are denoted by $\tilde q_{\tilde \theta_\ell}(x^{\ell-1}|x^{\ell})$ and allow to define a backward joint distribution with $\pi$ as last marginal:
\begin{align}
    \tilde \rho_{\tilde \theta}(x^0, \dots, x^L) = \pi(x^L)\prod_{\ell=1}^L \tilde q_{\tilde \theta_\ell}(x^{\ell-1}|x^{\ell}), \quad \pi(x^L) = \int \prod_{\ell=1}^{L} dx^{\ell-1} \,  \tilde \rho_{\tilde \theta}(x^0, \dots, x^L)  \; .
\end{align}
An importance sampling estimator is constructed at the level of the joint distributions based on the equality
\begin{align}
    \E_{\pi}[f(x)] = \E_{\rho_\theta}\left[f(x^L) \times w(x^0, \dots, x^L)\right], \quad \text{with } w(x^0, \dots, x^L) = \frac{\tilde \rho_{\tilde \theta}(x^0, \dots, x^L)}{\rho_\theta(x^0, \dots, x^L)} \; ,
\end{align}
for any test function $f:\X \to \R$. The resulting self-normalized importance sampling estimator of $\E_{\pi}[f(x)]$ reads
\begin{align}
    \hat f_{\text{SNF}} = \frac{\sum_{i=1}^N f(x^L_i) \, w(x^0_i, \dots, x^L_i)}{\sum_{i=1}^N w(x^0_i, \dots, x^L_i)} \quad \text{with } (x^0_i, \dots, x^L_i) \sim \rho_\theta(\cdot) \; .
\end{align}
The efficiency of the resulting estimator relies on the agreement between the forward and backward joint distributions $\rho_\theta$ and $\tilde \rho_{\tilde \theta}$, which in turn depends on the specification of the forward and backward kernels. 
In practice, \cite{wuStochasticNormalizingFlows2020a} suggests designing deterministic and stochastic layers such that the marginal distributions at the successive layers follow as closely as possible an interpolation path between the base distribution $\rzero$ and the target distribution $\pi$ (e.g. $\pi_\ell= \rzero^{(L-\ell)/L}\pi^{{\ell/L}}$), as traditionally done in AIS.

This approach has also strong connections with non-equilibrium statistical mechanics equalities by Jarzynski and Crooks \cite{jarzynskiNonequilibriumEqualityFree1997,crooksNonequilibriumMeasurementsFree1998}, and their repurposing for Monte Carlo \cite{nilmeierNonequilibriumCandidateMonte2011}. The advantage of learning NF layers on top of vanilla non-equilibrium Monte Carlo moves was validated in particular in high-energy physics applications \cite{caselleStochasticNormalizingFlows2022,caselleSamplingLatticeNambuGoto2024,caselleNumericalDeterminationWidth2025}. 
In the next section, we discuss a very related line of work combining NFs with SMC.

\subsubsection{Combining NFs with SMC} 
The additional ingredient of SMC methods compared to AIS is the intermediate resampling among parallel draws (referred to as particles). For this, SMC relies on a sequence of intermediate distributions $\left(\pi_\ell\right)_{\ell=0}^L$ with $\pi_0=\rzero$ and $\pi_L=\pi$ and $\pi_\ell=e^{-U_\ell}/Z_\ell$. The resampling steps aim at guaranteeing a good agreement between the distribution of the particles and the current target distribution $\pi_\ell$.

Combining SMC with NFs, \cite{arbelAnnealedFlowTransport2021,matthewsContinualRepeatedAnnealed2022a} assume to have a set of transport maps $\left(T_{\theta_\ell}\right)_{\ell=1}^{L}$ respectively trained to approximately transport $\pi_\ell$ to $\pi_{\ell+1}$. The SMC procedures then combines these maps, with the standard SMC steps of mutation, weighting and resampling. 
The algorithm can be summarized as follows: 

\begin{minipage}{\textwidth}
\captionof{algorithm}{\textbf{(Continuously Repeated) Annealed Flow Transport (AFT/CRAFT)}}
\label{alg:03-AFT}
\vspace{-0.4em}
\begin{algorithmic}
    \State Initialize particles $\{x_i^{(0)}\}_{i=1}^N \sim \pi_0$, 
    \State Initialize weights $\hat w_i^{(0)} = 1/N$ for $i=1 \ldots N$,
    \State Initialize normalization constant of $\rho_0$ $\hat \Z_0 = 1$,
    \For {$\ell = 1 \ldots L$}
        \For {$i = 1 \ldots N$}
            \State Transport particles $y_i^{(\ell)} = T_{\theta_\ell}(x_i^{(\ell-1)})$,
            \State Update weights
            $ w_i^{(\ell)} = \hat w_i^{(\ell-1)}\left| \det \nabla_x T_{\theta_\ell}(x_i^{(\ell-1)}) \right| {e^{-U_{\ell}(y_i^{(\ell)})}}/{e^{-U_{\ell-1}(x_i^{(\ell-1)})}} \, ,$
        \EndFor
        \State Estimate normalization constant $\hat \Z_\ell = \hat \Z_{\ell-1} \sum_{i=1}^N w_i^{(\ell)}$,
        \State Normalize weights $\hat w_i^{(\ell)} = w_i^{(\ell)} / \sum_{j=1}^N w_j^{(\ell)}$ for $i=1 \ldots N$,
        \State Compute the effective sample size $\text{ESS}^{(\ell)}$ for $\{\hat w_i^{(\ell)}\}_{i=1}^N$ as defined in \cref{eq:03-ess}
        \If{ESS$^{(\ell)} < \alpha N$}
            \For {$i = 1 \ldots N$}
                \State Resample index $a_i$ from categorical distribution with probabilities $\{\hat w_j^{(\ell)}\}_{j=1}^N$ %for $i=1 \ldots N$,
                \State Reset weight $\hat w_i^{(\ell)} = 1/N$ %for $i=1 \ldots N$,
            \EndFor
        \EndIf
        \For {$i = 1 \ldots N$}
            \State $\tilde y_i^{(\ell)} = y_{a_i}^{(\ell)}$,
            \State ``Mutate'' $x_i^{(\ell)} \sim M_{\ell}(\cdot|\tilde y_i^{(\ell)})$ with $M_{\ell}$ a $\pi_{\ell}$-invariant Markov kernel,
        \EndFor
    \EndFor
    \Return particles $\{x_i^{(L)}\}_{i=1}^N$ and normalization constant estimate $\hat \Z_L$.
\end{algorithmic}
\setlength{\parskip}{\savedparskip}
\vspace{0.5em}
\end{minipage}

As is now common practice in SMC, the resampling step is triggered only when the effective sample size (ESS) drops below a threshold
$\alpha$ (e.g. $\alpha=0.5$). Also note that the method allows to estimate the normalizing constant of the target distribution $\pi$ as a by-product (which is also the case for AIS but was omitted in the previous section for brevity). 

By dividing the task of sampling from $\pi$ into smaller subtasks of sampling sequentially from successive intermediate distributions $\pi_\ell$, AFT/CRAFT eases the learning of each transport map $T_{\theta_\ell}$ compared to elementary NF-powered samplers.  Additionally, the resampling step helps focus computational efforts on particles in high probability regions of the target distribution, which is particularly useful when the target is multimodal. Finally, the mutation step relying on local MCMC kernels $M_\ell$ helps improve the diversity of the particles and correct for residual errors in the flow approximations. While combining all these ingredients increases the computational cost of the sampler compared to flow-IMH or Neural-IS, it also logically improves its robustness to challenging sampling tasks. Compared to regular SMC, the addition of transport maps allows to use a coarser interpolation ladder: the requirement that $\pi_{\ell-1}$ and $\pi_{\ell}$ be close enough, to avoid weights degeneracy, becomes that the pushforward of $\pi_{\ell-1}$ through $T_\ell$ be close enough to $\pi_{\ell}$. As a result, fewer intermediate steps are necessary to reach a good sampling performance.

In a related proposition, \cite{karamanisAcceleratingAstronomicalCosmological2022} combines NFs with SMC using a single transport map that is trained progressively to transport the base distribution $\rzero$ to the current intermediate target $\pi_\ell$ at each iteration of SMC. In this work, the mutation step takes also advantage of the NF transport map to employ a neutra-MCMC kernel as described in \cref{alg:03-neutra-MCMC}. The performance of this approach was validated on cosmological and astronomical Bayesian inference tasks.

\subsubsection{Combining NFs with PT}
Following the propositions made for AIS and SMC, NFs can also be used to ease sampling in the PT framework.  As a reminder PT relies on a similar interpolation path $\left( x_\ell\right)_{\ell=0}^L$ and consists in simulating in parallel a set of $L+1$ replicas $\left( x_\ell\right)_{\ell=0}^L$ sampling respectively from the intermediate distribution $\pi_\ell$. Periodic swaps are proposed between replicas at adjacent levels $\ell$ and $\ell+1$. To ensure a reasonable acceptance rate for swap proposals, neighboring distributions $\pi_\ell$ and $\pi_{\ell+1}$ must be close enough, which often leads to the necessity of maintaining many replicas at high computational cost.

In this context, \cite{invernizziSkippingReplicaExchange2022} proposed to leverage NFs to skip all the intermediate steps altogether. The idea is to approximately transport an easy replica $x_0$ sampling from $\pi_0=\rzero$ to the target $\pi_L=\pi$ while transporting a target replica $x_L$ to $\rzero$, before attempting their swap. Thereby, only two replicas are strictly necessary. Following standard practice in PT, the base distribution is typically adapted to the target distribution, choosing $\pi_0 \propto \pi^\beta$, with $\beta<1$, a high-temperature version of the target distribution $\pi$. As a result, $\rzero=\pi_0$ is not as trivial to sample from than when choosing a standard Gaussian, but the assumption is that a local MCMC sampler can still efficiently explore $\pi_0$.

Denoting by $T_\theta$ a NF transport map trained to transport $\rzero$ to $\pi$, and $M_0$ and $M_\pi$ local Markov kernels leaving respectively $\rzero$ and $\pi$ invariant, the PT procedure is modified as follows:

\begin{minipage}{\textwidth}
\captionof{algorithm}{\textbf{Learned replica exchange (LREX)}}
\label{alg:03-LREX}
\vspace{-0.4em}
\begin{algorithmic}
    \State Initialize replicas $x_0^{(0)}$ and $x_\pi^{(0)}$ at random,
    \For{k=1\dots K}
        \State $x_0^{(k)} \sim M_0(\cdot|x_0^{(k-1)})$,
        \State $x_\pi^{(k)} \sim M_\pi(\cdot|x_\pi^{(k-1)})$,
        \State Draw $u\sim \text{Uniform}(0,1)$
        \If {$u < \gamma$}
            \State Attempt swap:
            \State \quad $x_0' = T_\theta^{-1}(x_\pi^{(k)})$ 
            \State \quad $x_\pi' = T_\theta(x_0^{(k)})$,
            \State \quad Compute
        $
            \alpha = \min\left(1, 
            \frac{\pi(x_\pi') \rho_\theta(T_\theta(x_0'))}{\pi(x_\pi^{(k)}) \rho_\theta(T_\theta(x_0^{(k)}))} \right) 
            = \min\left(1, 
            \frac{\pi(x_\pi') \rzero(x_0')}{\pi(x_\pi^{(k)})\rzero(x_0^{(k)})} 
            \times \left|\frac{\det \nabla T_\theta(x_0^{(k)})}{\det \nabla T_\theta(x_0')}\right| \, \right)
        $,
        \State \quad Set $x_0^{(k)}\leftarrow x_0'$, $x_\pi^{(k)} \leftarrow x_\pi'$ with probability $\alpha$
        \EndIf
    \EndFor
\end{algorithmic}
\setlength{\parskip}{\savedparskip}
\vspace{0.5em}
\end{minipage}

The swap is attempted with probability $\gamma$ at each iteration and $\rho_\theta$ denotes the density of the pushforward distribution induced by $T_\theta$, given by the change-of-variables formula (\cref{eq:02-nf-change-of-variable}). The acceptance probability $\alpha$ ensures that the overall Markov chain over the joint state space of the two replicas leaves the joint distribution $\rho(x_0, x_\pi)=\rzero(x_0)\pi(x_\pi)$ invariant. 

In spirit, the proposition of \cite{invernizziSkippingReplicaExchange2022} is similar to flow-IMH (\cref{alg:03-flow-IMH}) except that the base distribution is not sampled independently and the non-local moves are mediated by swaps between the base and target replicas. As a result, one can expect similar performances of this approach compared to flow-IMH.

A subsequent proposal by \cite{zhangAcceleratedParallelTempering2025} places weaker requirements on the NF and retains a ladder of intermediate distributions $\left(\pi_\ell\right)_{\ell=1}^L$. As in SMC-based (CR)AFT, this ladder can be chosen coarser than in traditional PT, thanks to a collection of transport maps $\left(T_\ell\right)_{\ell=1}^L$ that approximately push forward $\pi_{\ell-1}$ to $\pi_\ell$.
Extending the two-replica version of \cref{alg:03-LREX}, local MCMC kernels $M_\ell$ targeting each $\pi_\ell$ are run for every replica $x_\ell$, and swaps between neighboring levels are periodically proposed using the corresponding transport maps. Although this strategy requires maintaining more replicas, it is expected to handle more complex target distributions than the two-replica scheme, as it decomposes the global transport from $\rho$ to $\pi$ into a sequence of simpler transitions from $\pi_{\ell-1}$ to $\pi_\ell$.

\subsubsection{Recalibrating NF samples with AIS or SMC} Returning to the setting where a single transport map $T_\theta$ is trained to push forward $\rzero$ to $\pi$, it has also been proposed to leverage AIS or SMC in order to recalibrate the NF samples. The idea is to introduce a sequence of intermediate distributions that, here, gradually transports $\rho_\theta$ to $\pi$, for instance $\pi_\ell = \rho_\theta^{(L-\ell)/L} \pi^{\ell/L}$. Then, an AIS or SMC procedure is run targeting $\pi$, initialized from samples drawn from $\rho_\theta$. This strategy was first proposed by \cite{midgleyFlowAnnealedImportance2022b} in the context of AIS and more recently revisited by \cite{tanScalableEquilibriumSampling2025} using SMC.

In contrast to elementary NF-based samplers, this approach explicitly compensates for imperfections in the learned transport map. In particular, since $\rho_\theta$ may not be sufficiently close to $\pi$ to serve directly as a proposal distribution for IS or IMH, the tempering procedure corrects for this mismatch. As with other combinations of tempering and normalizing flows, the computational cost is higher than that of flow-IMH or Neural-IS, but this additional effort is expected to translate into improved sampling performance. Although other innovations are presented in \cite{midgleyFlowAnnealedImportance2022b} and \cite{tanScalableEquilibriumSampling2025}, the former reports good performance of the AIS recalibration strategy on sampling from the Boltzmann measure of clusters of repulsive particles and the latter validates the SMC recalibration strategy on the sampling of small biomolecules.\pbreak

Overall, the various proposals for combining NFs with tempering-based sampling schemes expand the capabilities of traditional tempering methods and ease the learning burden compared to the simple NF-powered samplers discussed above. However, a systematic comparison of these algorithms is still lacking. Moreover, while these approaches typically moderate the demands placed on learning—and are therefore likely to handle more complex targets than simple NF-enhanced samplers—it remains unclear whether they exhibit more favorable scaling with dimension. Indeed, transport is still performed in the full ambient space. The alternative strategy introduced next addresses this limitation directly by reducing the effective dimensionality of the problem.

\subsection{Combining NFs with coarse-grainings}
\label{subsec:03-coarse-graining}
One emerging limitation of samplers relying on NF-based proposals is their poor scaling with dimension. Expecting a generative model to learn how to resample an entire high-dimensional variable in a coherent manner is, in a sense, overly ambitious. This observation motivates another line of research seeking to leverage NFs while avoiding a full resampling. This is the underlying principle of the two methods described below. 

\subsubsection{Partial resampling in the latent space}
A first strategy to mitigate the deterioration of acceptance rates with increasing dimension is to perform partial updates informed by the generative model trained to approximate $\pi$. With autoregressive models, this operation is straightforward provided that one updates variables appearing last in the autoregressive ordering \cite{wuUnbiasedMonteCarlo2021}.
However, in phases with strong correlations, updating a finite number of variables at the end of the ordering may not be sufficient to connect distant modes of the target distribution. 

In this context, \cite{schonleOptimizingMarkovChain2023} proposed an adaptation of this idea exploiting the latent space of a normalizing flow (with base distribution $\rzero$ and transport map $T_\theta$) where the variables are less correlated.
The method, referred to as Gibbs flow-Monte Carlo (GFlowMC), uses a Metropolis-within-Gibbs scheme on the pullback $\pi_{T_\theta}$ of $\pi$ through $T_\theta$ (see \cref{eq:03-pullback}) using the conditional distributions of $\rzero$ as proposal distributions, and subsequently accepting or rejecting the proposal according to a Metropolis acceptance rule.

More precisely, let $d_u \leq d$ denote the number of degrees of freedom to be updated, $S = (i_1, \dots, i_{d_u})$ a selected set of distinct indices, and $q(S)$ a probability distribution over such subsets. Denote by $z_S$ and $z_{\setminus S}$ the components of the base variable $z$ indexed by $S$ and by its complement, respectively. The partial resampling  MCMC procedure of \cite{schonleOptimizingMarkovChain2023} consists in:

\begin{minipage}{\textwidth}
\captionof{algorithm}{\textbf{Gibbs flow-Monte Carlo (GflowMC)}}
\label{alg:03-GflowMC}
\vspace{-0.4em}
\begin{algorithmic}
    \State Initialize $x^{(0)}$,
    \State Compute $z^{(0)} = T_\theta^{-1}(x^{(0)})$,
    \For {$k = 1 \ldots K$}
        \State Sample set $S \sim q(S)$,
        \State Propose partial refreshing $z_S' \sim \rzero(z_S|z^{(k-1)}_{\setminus S})$, 
        \State Set $z'_{\setminus S} = z^{(k-1)}_{\setminus S}$,
        \State Compute acceptance rate
        $\displaystyle \alpha = \min\left(1 \,, \; \frac{\pi_{T_\theta}(z')}{\pi_{T_\theta}(z^{(k-1)}) } 
        \times \frac{\rzero(z_S|z^{(k-1)}_{\setminus S})}{\rzero(z_S'|z^{(k-1)}_{\setminus S})}\right) \, ,$
        \State Set $z^{(k)} = z'$ with probability $\alpha$, else $z^{(k)} = z^{(k-1)}$,
    \EndFor
    \Return samples $\{x^{(k)} = T_\theta(z^{(k)})\}_{k=1}^K$
\end{algorithmic}
\setlength{\parskip}{\savedparskip}
\vspace{0.5em}
\end{minipage}

Here the density of the pullback $\pi_{T_\theta}$ is again needed in the acceptance ratio.
The conditional base distributions of the form $\rzero(z_S|z_{\setminus S})$ are assumed to be explicit and easy to sample from. For a factorized $\rzero$, this is trivially the case as the conditional distribution $\rzero(z_S|z_{\setminus S})$ is equal to the marginal $\rzero(z_S)$, itself factorized -- e.g. for the common choice $\rzero = \N(0_d,I_d)$, $\rzero(z_S|z_{\setminus S})=\N(0_{d_u},I_{d_u})$. 

\begin{figure}[t]
    \centering
    \includegraphics[width=0.95\linewidth]{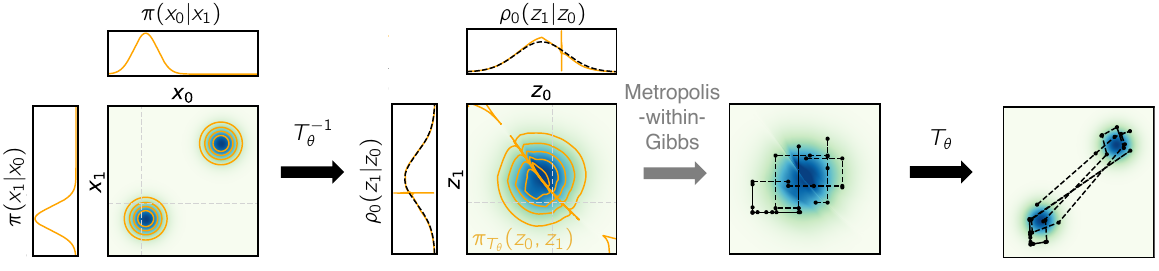}
    \caption{\textbf{Illustration of the Gibbs flow-Monte Carlo (GFlowMC) algorithm.} The target distribution $\pi$ is a mixture of two Gaussians (left pannel). The NF inverse transport map $T_\theta^{-1}$ approximately transports $\pi$ to a standard Gaussian base distribution $\rzero$ (middle left panel). The GFlowMC algorithm (middle right panel) performs partial resampling in the latent space of the NF targetting the pullback $\pi_{T_\theta}$ (in orange), here updating one variable at a time ($d_u=1$). Sending the samples back to the original space (right panel) evidence the ability of GFlowMC to jump between the two modes of $\pi$.
    Adapted from the poster of C. Schönle on \cite{schonleOptimizingMarkovChain2023}.}
    \label{fig:03-gflowmc}
\end{figure}

The algorithm shares similarities with neutra-MCMC (\cref{alg:03-neutra-MCMC}), but replaces the local MCMC updates in the latent space with partial global updates that can more easily jump across modes, as illustrated by the two-dimensional example in \cref{fig:03-gflowmc}. In the context of statistical field theories, \cite{schonleOptimizingMarkovChain2023} report that GFlowMC can outperform both neutra-MCMC and flow-IMH. In particular, the number of degrees of freedom updated, $d_u$, acts as a tunable parameter for optimizing sampler performance by trading off acceptance rate against the decorrelation in a single move.

In an application to biomolecular systems, \cite{pengFlowPerturbationAccelerate2025} exploit a similar idea using continuous normalizing flows (CNFs). They report that updating a single latent variable at a time ($d_u = 1$) is optimal, and demonstrate a significant acceleration in equilibrating samples of the small, fast-folding protein Chignolin.

In the strategy described in the above paragraphs, the NF is still trained to approximate the full target distribution $\pi$. An alternative approach consists in leveraging generative models to sample a low-dimensional coarse-grained representation of the system, which is the idea developed next.

\subsubsection{Collective-variable guided sampling}
As already mentioned in the introduction, apart from annealing-based methods, another key strategy to address metastability in high dimension relies on the existence of a low-dimensional representation (or coarse-graining), referred to as collective variables (CVs), that effectively separates the different modes of the target measure. In this context, generative models provide a flexible and scalable framework to estimate the marginal law of the CVs. In contrast to traditional non-parametric approaches — which typically restrict CV-guided algorithms to very low-dimensional CV spaces due to the difficulty of in keeping track of their marginal density — generative models make it possible to handle significantly higher-dimensional CV spaces. Generative models therefore extend the range of admissible dimensions for the CV representation in coarse-graining-based sampling strategies.

Here, we review the approach proposed by \cite{tamagnoneCoarseGrainedMolecularDynamics2024,schonleSamplingMetastableSystems2025,schonleEfficientMonteCarloSampling2025}, revisiting the concept of CV-guided sampling using a NF trained to sample the distribution of the CVs.
The approach is based on non-equilibrium samplers previously introduced without resorting to NFs, notably in \cite{nealSamplingMultimodalDistributions1996,athenesComputationChemicalPotential2002,
nilmeierNonequilibriumCandidateMonte2011,chenGeneralizedMetropolisAcceptance2015}.

For simplicity, we discuss the separable case where the CVs are a subset of the full variables, i.e. $x=(z, x^\perp) \in \R^{d}$, with $z \in \R^{d_z}$ the CVs and $x^\perp \in \R^{d - d_z}$ the remaining degrees of freedom. The strategy presented below readily generalizes to an arbitrary CV mapping  $z = \xi(x)$ following \cite{schonleEfficientMonteCarloSampling2025}. In the separable case, the marginal distribution over CVs of the target distribution $\pi(x)$ is given by
\begin{align}
    \pi_z(z) = \int \pi(z, x^\perp) \, dx^\perp \,.  
\end{align}
The sampling algorithm then relies on two main ingredients: 
\begin{itemize}
    \item[(i)] a NF trained to sample approximately from the marginal distribution of the CVs, its probability density function is noted $\rtheta(z)$,
    \item[(ii)] a non-equilibrium sampling procedure updating the value of $x^\perp$ conditional on an updated value of the CVs $z$. 
\end{itemize} 
Let us describe the second ingredient. 

Given the current state $x = (z, x^\perp)$ and a proposed new value $z'$ for the CVs, we introduce a $T$-step path $S(z,z') = (z_t)_{t=0}^T$ connecting $z$ to $z'$, with $z_0 = z$ and $z_T = z'$. A simple choice is the linear interpolation $z_t = z + (z' - z)\, t/T$. Starting from $x_0^\perp = x^\perp$, the orthogonal degrees of freedom are then updated through a sequence of $T$ Markov kernels associated with the intermediate values $z_t$:

\begin{minipage}{\textwidth}
\captionof{algorithm}{\textbf{Non-equilibrium update of orthogonal degrees of freedom}}
\label{alg:03-non-eq-update}
\vspace{-0.4em}
\begin{algorithmic}
    \State {\textbf{Input:}} Current state $x = (z, x^{\perp})$, target CVs $z'$, Markov kernels $\left(M_{z}\right)_{z \in \R^{d_z}}$, \\
    {\color{white}\textbf{Input:}}  number of steps $T$, schedule $S(z, z') = (z_t)_{t=0}^T$
    \State Initialize $x^{\perp}_0 = x^{\perp}$,
    \For {$t = 0 \ldots K-1$}
        \State Sample $x^{\perp}_{t+1} \sim M_{z_t}(\cdot|x^{\perp}_t)$,
    \EndFor 
    \Return Final chain of states $\left(z_t, x^{\perp}_t\right)_{t=0}^T$
\end{algorithmic}
\setlength{\parskip}{\savedparskip}
\vspace{0.5em}
\end{minipage}

This elementary algorithm defines a procedure for proposing a new value of the full state
$x' = (z', {x^\perp}') = (z_T, x_T^\perp)$ given a proposed CV update $z \rightarrow z'$. Although, in principle, any Markov kernel $M_z$ may be used, efficiency of the algorithm described below requires choosing $M_z$ such that the resulting proposal $(z', {x^\perp}')$ is likely under the target distribution $\pi$. A common and effective choice is a single step of overdamped Langevin dynamics targeting the conditional distribution $\pi(x^\perp \mid z)$ at fixed $z$:
\begin{align}
    M_z(x^{\perp}_{t+1}|x_t^{\perp}) = \N\left(x^{\perp}_{t+1}; x^{\perp}_t - \delta t \nabla_{x^\perp} U(z, x^{\perp}_t), 2 \delta t I_{d - d_z} \right) \, ,
\end{align}
where $\delta t$ is a small time-step and $U(x) = -\log \pi(x)$ (up to an additive constant) is the potential energy function associated with the target distribution.

Using a Markov kernel $Q$ to propose an update of the CV, an accept-reject criterion can be written à la Metropolis-Hastings to design an MCMC with stationary distribution $\pi$.
For this, the CV-schedule needs to be reversible, meaning that $S(z,z')= (z_t)_{t=0}^T$ implies that $S(z', z) = (z_{T-t'})_{t'=0}^T$. The probability density of the generated path and the probability density of following the reverse path with the same protocol are respectively given by
\begin{align}
    Q(z'|z) \prod_{t=0}^{T-1} M_{z_t}(x^{\perp}_{t+1}|x^{\perp}_t) \, , \quad \text{and} \quad Q(z|z') \prod_{t'=0}^{T-1} M_{z_{T-t'}}(x^{\perp}_{T - t' - 1}|x^{\perp}_{T - t'}) \, .
\end{align}
Then a sufficient criterion for ensuring that the overall Markov chain leaves $\pi$ invariant is to accept the proposed move $(z, x^{\perp}) \rightarrow (z', {x^\perp}')$ with probability:
\begin{align}
    \alpha = \min\left(1, \frac{\pi(z', {x^\perp}')} {\pi(z, x^{\perp})} \times \frac{Q(z|z')}{Q(z'|z)} \times \frac{\prod_{t'=0}^{T-1} M_{z_{T-t'}}(x^{\perp}_{T - t' - 1}|x^{\perp}_{T - t'})}{\prod_{t=0}^{T-1} M_{z_t}(x^{\perp}_{t+1}|x^{\perp}_t)} \right) \,.
    \label{eq:03-cv-flow-mc-acceptance}
\end{align} 
This acceptance rule satisfies a property stronger than detailed balance, sometimes referred to as super-detailed balance \cite{frenkelSpeedupMonteCarlo2004,nilmeierNonequilibriumCandidateMonte2011}. By using a NF trained to approximate $\pi_z$ to propose updates in the CV space — that is, by taking $Q = \rtheta$ — one can promote transitions between metastable modes. In turn, this mechanism yields an efficient sampler for complex multimodal distributions. The complete algorithm reads as follows:

\begin{minipage}{\textwidth}
    \captionof{algorithm}{\textbf{Collective-variable guided flow-Monte Carlo (CV-FlowMC)}}
\label{alg:03-cv-flow-mc}
\vspace{-0.4em}
\begin{algorithmic}
    \State Initialize $x^{(0)} = (z^{(0)}, x^{\perp (0)})$,
    \For {$k = 1 \ldots K$}
        \State Propose new CVs $z' \sim \rtheta(z)$,
        \State Generate non-eq. path $\left(z_t, x^{\perp}_t\right)_{t=0}^T$ using \cref{alg:03-non-eq-update} with input $(z^{(k-1)}, x^{\perp (k-1)})$ and $z'$,
        \State Set $x' = (z', x^{\perp}_T)$,
        \State Compute acceptance rate $\alpha$ using \cref{eq:03-cv-flow-mc-acceptance},
        \State Set $x^{(k)} = x'$ with probability $\alpha$, else $x^{(k)} = x^{(k-1)}$,
    \EndFor
    \Return samples $\{x^{(k)}\}_{k=1}^K$
\end{algorithmic}
\setlength{\parskip}{\savedparskip}
\vspace{0.5em}
\end{minipage}

Many variants of this algorithm can be envisioned, ranging from technical choices in the definition of the reverse protocol to different selections of the Markov kernels $M_z$ and associated hyperparameters (e.g., the number of steering steps $T$, or the time step $\delta t$ when using overdamped Langevin dynamics). We refer to \cite{schonleSamplingMetastableSystems2025,schonleEfficientMonteCarloSampling2025} for a detailed analysis of these design choices and their impact on sampler performance. As mentioned above, the method can moreover be extended to arbitrary CV mappings $z = \xi(x)$ by evolving the orthogonal degrees of freedom $x^\perp$ at fixed CV values via constrained dynamics, as described in \cite{schonleEfficientMonteCarloSampling2025}.

\pbreak

The core idea of combining coarse-graining with flows for sampling is the use of a NF to propose updates in the CV space that traverse low-probability regions and connect metastable modes. Crucially, the flow is trained only in a coarse-grained space ($d_z < d$), whose marginal distribution is typically smoother than the full target distribution $\pi$. This substantially alleviates the curse of dimensionality affecting NF-based samplers in the full space, while preserving their ability to generate large, mode-crossing moves. In particular, the admissible dimensionality of the CV space is substantially higher than in traditional CV-based enhanced sampling methods: values on the order of $d_z \approx 10\text{s}-100\text{s}$ appear feasible, whereas standard approaches often struggle beyond $d_z \approx 3\text{-}4$ due to the difficulty of estimating the marginal distribution in higher dimensions with non-parametric methods. 

Finally, for all the exact samplers leveraging NFs discussed in this \cref{sec:03-exact_samplers}, we have so far assumed that the generative model (either a normalizing flow or an autoregressive model) is trained beforehand to approximate either the full target distribution or its marginal over the CV space. The question of how to train such models efficiently for sampling purposes is, however, a substantial research direction in its own right, to which we now turn.

\pagebreak

\pagebreak
\section{Training generative models without data}
\label{sec:04-training}
In their conventional setting, generative models are trained on large datasets consisting of samples drawn from an unknown target distribution. In scientific computing, the situation is markedly different: the target distribution $\pi$ is typically known—at least up to a normalization constant—and the central challenge is precisely to generate samples from it. This fundamental distinction calls for a rethinking of the training paradigm. Rather than learning from empirical data, the model must instead exploit direct access to the unnormalized density of the target distribution. \emph{In this section, we assume that $\pi(x)$ is easy to evaluate up to normalization at any point $x$ in $\X$ but hard to sample, and we discuss how to leverage this information to train different generative models to approximate $\pi$.}

We organize the discussion according to the main classes of generative models introduced in \cref{sec:02-tuto_gm}. Throughout, we use the notation $\pi = e^{-U}/\Z$ for the target distribution, where $U$ denotes the potential energy function and $\Z$ the (unknown) normalization constant.

\subsection{Training exact likelihood generative models: normalizing flows and autoregressive models}
\label{subsec:04-nfs}
\subsubsection{Variational inference}

\paragraph{Minimizing the \emph{reverse} KL divergence}
Given their tractable likelihood, normalizing flows (NFs) and autoregressive models (ARMs) are typically trained for generative modeling by maximizing the likelihood of the training data. Equivalently, this corresponds to minimizing the Kullback-Leibler (KL) divergence between the data distribution $p_{\mathrm{data}}$ and the model's distribution $\rtheta$:
\begin{align}
    \label{eq:03-fwd-kl}
D_{\mathrm{KL}}(p_{\mathrm{data}} \| \rtheta) = \E_{p_{\mathrm{data}}} \left[ \log \frac{p_{\mathrm{data}}(x)}{\rtheta(x)} \right]
= - \mathcal{H}(p_{\mathrm{data}}) - \E_{p_{\mathrm{data}}} \left[ \log \rtheta(x) \right],
\end{align}
where $\mathcal{H}(p_{\mathrm{data}})$ is the entropy of the data distribution, independent of the model parameters $\theta$. 
In the context of sampling, however, the ``data'' distribution is replaced by a target distribution $\pi$, and computing expectations with respect to $\pi$ is typically challenging. There, the \emph{reverse} KL divergence can serve as a practical training objective instead:
\begin{align}
D_{\mathrm{KL}}(\rtheta \| \pi) = \E_{\rtheta} \left[ \log \frac{\rtheta(x)}{\pi(x)} \right]
= \E_{\rtheta} \left[ \log \rtheta(x) + U(x) \right] + \log \Z \,.
\end{align}
Expectations with respect to $\rtheta$ are cheap to estimate by drawing independent samples from the NF or ARM, $\log \rtheta$ is available in closed form, and $\log \Z$ is unknown but independent from $\theta$. Minimizing this objective corresponds to variational inference (VI), a well-established technique in Bayesian statistics \cite{jordanIntroductionVariationalMethods1999,bleiVariationalInferenceReview2017}. 

Variational families for $\rho_\theta$ must maintain tractable sampling and density evaluation. Traditional choices, such as factorized or Gaussian distributions, are often poorly suited to approximate complex target distributions because they lack flexibility. NFs and ARMs, in contrast, are highly expressive while preserving the required ease of sampling and density evaluation.
Using NFs for VI was first proposed by \cite{rezendeVariationalInferenceNormalizing2015}, who leveraged the reparameterization trick to avoid backpropagating through the sampling process. Writing $x \sim \rtheta$ as $x = T_\theta(z)$ with $z \sim \rzero$ the base distribution, the effective loss for a batch of $N$ samples $z_i \sim \rzero \; \text{i.i.d.}$ becomes
\begin{align}
\ell_N(\theta) = \frac{1}{N} \sum_{i=1}^N \left[ \log \rzero(z_i) - \log \det \left|\nabla_{z} T_\theta(z_i)\right| + U(T_\theta(z_i)) \right] \,.
\end{align}
Following this foundational work, simultaneous applications in statistical physics emerged: \cite{albergoFlowbasedGenerativeModels2019} applied NFs to field theories, \cite{noeBoltzmannGeneratorsSampling2019} to molecular systems -- where reverse-KL training is called ``training by energy''--, and \cite{wuSolvingStatisticalMechanics2019} to spin systems using ARMs\footnote{Here to avoid the issue of having to backpropagate through the sampling process, one compute a Monte Carlo estimator of the training gradients to be plugged into the optimization algorithm rather than applying autodiff directly to the loss estimator (see \cite{wuSolvingStatisticalMechanics2019}).}. However, this seemingly data-free training approach has an important limitation when applied to multimodal target distributions, a phenomenon often referred to as mode collapse.

\paragraph{Mode collapse in VI}
Mode collapse refers to the tendency of the learned distribution $\rtheta$ to cover only a subset of the modes of the target distribution $\pi$. This phenomenon stems from the well-known mode-seeking behavior of the reverse KL divergence: assigning probability mass to regions where $\pi$ has low density is penalized much more strongly than failing to cover some modes of $\pi$ altogether \cite{jordanIntroductionVariationalMethods1999,minkaDivergenceMeasuresMessage2005}. Crucially, this behavior is intrinsic to the objective function itself and is therefore not resolved by increasing model expressiveness. As a result, even highly flexible models such as NFs and ARMs can suffer from mode collapse, despite their capacity to represent multimodal distributions, as reported by several authors \cite{nicoliDetectingMitigatingModecollapse2023,blessingELBOsLargeScaleEvaluation2024a,greniouxImprovingEvaluationSamplers2025b}. This limitation is particularly severe in sampling applications of generative models, since multimodal target distributions are precisely those for which traditional sampling methods struggle the most. 

A number of heuristic strategies have been proposed to alleviate mode collapse in VI when training exact likelihood models for sampling purposes. For example, \cite{noeBoltzmannGeneratorsSampling2019} introduced a hybrid training procedure that augments the reverse KL objective with a maximum-likelihood term based on short MCMC trajectories initialized near distinct modes. Alternatively, \cite{midgleyFlowAnnealedImportance2022b} proposed estimating the $\chi^2$ divergence between $\rtheta$ and $\pi$ using AIS, and minimizing this estimate instead. Another broadly applicable approach consists in combining VI with annealing, which we discuss next.

\paragraph{Annealed VI to mitigate mode collapse} A natural strategy to mitigate mode collapse when training NFs is to introduce annealing, in a manner reminiscent of annealed sampling schemes. The key idea is to define a sequence of intermediate distributions $\left(\pi_k\right)_{k=0}^K$ interpolating between a simple unimodal distribution $\pi_0$ and the target distribution $\pi_K = \pi$. The model is then trained sequentially: 
at each stage $k$, variational inference (VI) is used to approximate $\pi_k$, initializing the parameters of ${\rtheta}_k$ from those obtained at stage $k-1$.

In practice, the intermediate distributions are often defined through tempering of the target distribution, $\pi_k \propto e^{-\beta_k U}$, with a monotonically increasing sequence of inverse temperatures $\beta_0 < \beta_1 < \ldots < \beta_K = 1$ or through a geometric interpolation toward a chosen $\pi_0$, $\pi_K \propto \pi^{k/K}\pi_0^{(K-k)/K}$. The progressive training strategy along the interpolation path enables the NF to gradually discover and cover distinct modes of the target distribution, thereby reducing the risk of mode collapse. 

Annealed VI has demonstrated empirical success across a range of applications: the exploration of complex Bayesian posteriors \cite{huangImprovingExplorabilityVariational2018,wangMitigatingModeCollapse2025}, the sampling of Boltzmann distributions in statistical mechanics \cite{wuSolvingStatisticalMechanics2019,hackettFlowbasedSamplingMultimodal2021,schopmansTemperatureAnnealedBoltzmannGenerators2025}, and in optimization tasks ranging from locating multiple modes \cite{leminhNaturalVariationalAnnealing2025} to identifying global minima \cite{hibat-allahVariationalNeuralAnnealing2021a,khandokerLatticeProteinFolding2025,bonoDemonstratingRealAdvantage2025}. A closely related strategy is employed in the previously discussed (CR)AFT samplers \cite{arbelAnnealedFlowTransport2021,matthewsContinualRepeatedAnnealed2022a}, where the flow architecture itself is expanded at each annealing stage by adding new layers, rather than retraining a fixed model. 

Despite these encouraging results, performance remains highly sensitive to the annealing schedule—most notably the choice of the initial distribution $\pi_0$ and the rate at which the transition is operated toward the target distribution. To date, theoretical guidance on these choices within the VI framework remains limited. Recent work \cite{soletskyiTheoreticalPerspectiveMode2025,foglianiAnnealingVariationalInference2026}, however,
has begun to analyze the behavior of (annealed) VI in simplified settings, in particular Gaussian mixture models. 

In \cite{soletskyiTheoreticalPerspectiveMode2025}, the authors identify a critical threshold for the separation between modes—dependent on their relative weights but independent of the ambient dimension—that determines whether mode collapse occurs. This behavior is also observed empirically when training NFs on Gaussian mixtures.
Building on this analysis, \cite{foglianiAnnealingVariationalInference2026} derive a closed-form expression for the probability of mode collapse in the bimodal Gaussian mixture case, enabling a more detailed study of the impact of the annealing schedule. Their results confirm the mitigating role of annealing and provide concrete guidance for its practical implementation: starting from a sufficiently high temperature is not, by itself, sufficient—one must also ensure that the annealing schedule is slow enough. These theoretical predictions are again supported by experiments with NFs trained on Gaussian mixtures. 

These works constitute an important first step toward a more rigorous understanding of (annealed) VI with expressive generative models, such as ARMs and NFs, and offer valuable insight for its use in challenging sampling applications.

\subsubsection{Adaptive strategies}
\label{subsubsec:04-adaptive-strategies}
While training exact likelihood models via the reverse KL divergence requires explicit safeguards against mode collapse—such as annealing—an alternative line of work has focused on retaining the standard and notably stable maximum-likelihood training paradigm, equivalent to \cref{eq:03-fwd-kl}. Because direct samples from the target distribution $\pi$ are not available, these approaches rely on \emph{adaptive strategies} in which data collection and model training are interleaved. This idea was first introduced by \cite{parnoTransportMapAccelerated2018} in the context of triangular transport maps, independently extended to general variational families by \cite{naessethMarkovianScoreClimbing2020}, and later rediscovered for modern NFs in \cite{Gabrie2021a,Gabrie2021}. We summarize the latter approach below (which is straightforwardly applicable to ARMs as well).

The adaptive training procedure runs an MCMC chain targeting $\pi$, which alternates between two types of Markov update kernels:
\begin{itemize}
    \item[(i)] A \emph{local} MCMC kernel (e.g., Langevin dynamics or Hamiltonian Monte Carlo) that leaves $\pi$ invariant and explores the target distribution, denoted $M_{\pi}$ below;
    \item[(ii)] A \emph{global} NF-based kernel that also leaves $\pi$ invariant -- such as the flow-independent Metropolis-Hastings kernel described in \cref{alg:03-flow-IMH} --and enables long-range, mode-crossing updates, denoted $M_{\rtheta}$ below. 
\end{itemize}
The NF is periodically retrained by maximum likelihood on the samples collected so far, leading to the following overall scheme:

\begin{minipage}{\textwidth}
\captionof{algorithm}{\textbf{Adaptive flow Monte Carlo (A-flowMC)}}
\label{alg:04-ada-flow-MC}
\begin{algorithmic}
    \State {\textbf{Input:}} Initial NF parameters $\theta$, learning rate $\eta$, frequency of global jumps $K_g$, frequency of retraining $K_t$
    \State Initialize $x^{(0,m)}$ for $m=1\ldots M$ in basins of attraction of the distinct modes of $\pi$
    \For {$k = 1 \ldots K$}
        \If{$k \mod K_g = 0$}
            \State Global NF-based move $x^{(k, m)} \sim M_{\rtheta}(\cdot|x^{(k-1, m)})$ 
        \Else
            \State Local MCMC move $x^{(k, m)} \sim M_{\pi}(\cdot|x^{(k-1, m)})$ 
        \EndIf
        \If{$k \mod K_t = 0$}
            \State $\displaystyle \theta \leftarrow \theta + \frac{\eta} {M k} \sum_{m=1}^M \sum_{\ell=1}^k \nabla \log \rtheta(x^{(\ell, m)}) $
        \EndIf
    \EndFor
    \Return trained flow $\rho_\theta$, samples $\left(x^{(k, m)}\right)_{m,k=1}^{M,K}$
\end{algorithmic}
\setlength{\parskip}{\savedparskip}
\vspace{0.5em}
\end{minipage}
Variants of this basic scheme include discarding the oldest samples to limit memory usage and improve equilibration, subsampling mini-batches from previously collected samples, performing multiple gradient steps at each retraining stage, and using more advanced optimizers such as Adam instead of simple stochastic gradient descent (SGD). 

The key feature of this approach is that it simultaneously achieves two objectives: it produces unbiased samples from the target distribution $\pi$, and it trains a NF  (or an ARM) that can subsequently be used within an efficient standalone sampler from \cref{sec:03-exact_samplers}. From a theoretical standpoint, however, proving convergence of such adaptive MCMC schemes—where the transition kernel evolves over time based on the history of the chain and thus violates the Markov property—is non-trivial \cite{brofosAdaptationIndependentMetropolisHastings2022}. In particular, convergence of the Markov chain to the target measure $\pi$ typically requires the adaptation to vanish asymptotically. In practice, this issue can be sidestepped by stopping the training once the NF is deemed sufficiently accurate, and then continuing the simulation with a fixed kernel to generate exact samples from $\pi$. 

The benefit of alternating between local and global moves is that the two components reinforce each other. Local MCMC updates explore individual modes in detail, providing training data that allow the NF to capture local structure, while the NF enables global proposals that help the chain escape metastable states. Together, these mechanisms facilitate accurate estimation of the relative weights of the different modes. Rigorous analyses demonstrating the advantages of combining local and global sampling have been established for NFs in \cite{samsonov2022localglobal} and for ARMs in \cite{delbonoPerformanceMachinelearningassistedMonte2025,bonoDemonstratingRealAdvantage2025}.  

Provided that the MCMC chains can be initialized in all modes of interest of the target distribution, the maximum-likelihood objective naturally prevents mode collapse, since all modes present in the training data must be represented by the generative model. A notable exception arises for modes that are significantly rarer than others: in this case, some training batches may contain no samples from these rare modes, causing the NF to accidentally forget them. This issue can be mitigated by enforcing that training batches include samples from all relevant modes, or by using a mixture of models in which each component is trained on a specific mode and the mixture weights are lower bounded \cite{Gabrie2021a,hackettFlowbasedSamplingMultimodal2021,molina-tabordaActiveLearningBoltzmann2024a}. Aside from these rare-mode scenario—which requires special care—the main limitation of adaptive strategies based on maximum-likelihood training remains the need to initialize chains in all relevant modes. As for mode collapse, this limitation can be alleviated by resorting to annealing, as discussed next.

\subsubsection{Sequential tempering}
The idea of adaptive maximum-likelihood training can be easily combined with annealing in order to relax the assumption that all relevant modes of the target distribution are known in advance. This sequential tempering is therefore built upon an annealing schedule $\left(\pi_k\right)_{k=0}^K$ defined, as before, with $\pi_0$ a simple, typically unimodal distribution and $\pi_K = \pi$ the target distribution and follows an iterative procedure.

The algorithm is initialized by sampling from $\pi_0$ using a suitable MCMC method, a task that is assumed to be straightforward. An exact likelihood model is then trained by maximum likelihood on the resulting samples from $\pi_0$. Once trained, the generative model is used to facilitate sampling from the slightly more challenging distribution $\pi_1$, for instance via flow/ARM-independent Metropolis-Hastings (\cref{alg:03-flow-IMH}). The samples collected at this stage are used in turn to retrain the generative model by maximum likelihood, initializing the optimization from the parameters learned at the previous level.
This procedure is repeated along the annealing path until the target distribution $\pi_K$ is reached, leading to the following overall scheme:

\begin{minipage}{\textwidth}
\captionof{algorithm}{\textbf{Sequential tempering}}
\label{alg:04-seq-tempering}
\begin{algorithmic}
    \State {\textbf{Input:}} Initial NF parameters $\theta$, learning rate $\eta$
    \State  $x^{(0, m)}\sim \pi_0$ using MCMC for $m=1\ldots M$
    \For {$k = 0 \ldots K-1$}
        \Repeat 
            \State $\displaystyle \theta \leftarrow \theta + \frac{\eta} {M} \sum_{m=1}^M  \nabla_\theta \log \rtheta(x^{(k, m)}) $
        \Until{convergence}
        \State $x^{(k+1, m)} \sim$ flow/ARM-IMH targeting $\pi_{k+1}$ using current $\rtheta$ for $m=1\ldots M$
    \EndFor
    \Return trained flow $\rtheta$, samples $\left(x^{(K, m)}\right)_{m=1}^{M}$
\end{algorithmic}
\setlength{\parskip}{\savedparskip}
\vspace{0.5em}
\end{minipage}

Training the flow to convergence at each annealing stage robustly prevents mode collapse, provided that the annealing schedule is sufficiently fine. This conservative strategy was termed ``adiabatic retraining'' by \cite{hackettFlowbasedSamplingMultimodal2021}. In practice, however, it may be computationally more efficient to perform only a limited number of training steps at each stage, in the same spirit as adaptive training without annealing (\cref{alg:04-ada-flow-MC}). Note that, in the same spirit, \cite{cabezasMarkovianFlowMatching2024} proposed an annealed adaptive training for flow matchings (\cref{subsec:02-stochastic-interpolants}), replacing the maximum-likelihood objective with the likelihood-free flow-matching loss \cref{eq:02-SI-loss}, and substituting the independent Metropolis-Hastings sampler with a sampler related to NeutraMCMC, which however requires the costly evaluation of the flow Jacobian determinant. 

This sequential tempering strategy has been primarily explored in the context of sampling disordered spin systems using autoregressive models, beginning with the work of \cite{mcnaughtonBoostingMonteCarlo2020} and subsequently adopted in \cite{ciarellaMachinelearningassistedMonteCarlo2023,delbonoPerformanceMachinelearningassistedMonte2025,bonoDemonstratingRealAdvantage2025}. Such settings are particularly well suited to sequential tempering, as the number and location of modes are typically unknown. Nevertheless, the approach may still fail to produce exact samples when the target distribution is too complex to be faithfully represented by the chosen model class. This limitation was illustrated in \cite{ciarellaMachinelearningassistedMonteCarlo2023}, where autoregressive models were found to be inadequate for sampling solutions of a graph-coloring constraint satisfaction problem.

Going further, \cite{ciarellaMachinelearningassistedMonteCarlo2023} analyze the mechanisms underlying these failures by comparing different training objectives, namely VI and maximum-likelihood training with tempering. VI tends to produce approximations $\rtheta$ that underestimate the entropy of $\pi$, in line with the mode collapse phenomenon, while still generating configurations with appropriate energies. In contrast, maximum-likelihood training accurately captures the entropy but often yields configurations with energies higher than those of the true equilibrium distribution. 
While this study focused on autoregressive models, similar phenomena are likely to arise when training normalizing flows under either objective.
% , since both model classes are constrained by the requirements of easy sampling and tractable density evaluation.
To move beyond these limitations, one can consider alternative classes of generative models amenable to different training objectives, as detailed in the next sections.

\subsection{Continuous annealed flows}
\label{subsec:04-annealed-cnfs}
Continuous normalizing flows (CNFs) based on ODEs are typically more expressive than their discrete counterparts. However, evaluating their likelihood (see \cref{eq:02-instantaneous-change-of-variables-b}) is computationally demanding and inevitably subject to discretization error. Consequently, the training methods described in the previous sections for "discrete-time" normalizing flows are not well suited to the continuous setting.

Instead, one can exploit the intrinsic continuity of CNFs and train them for sampling by matching the trajectories of the flow to a prescribed annealing path of distributions. Here we consider a path $\left(\pi_t\right)_{t \in [0,1]}$ indexed by a continuous time $t\in[0,1]$, with $\pi_0$ a simple distribution and $\pi_1 = \pi$ the target. This approach was first developed for deterministic continuous flows (as CNFs presented in \cref{subsec:02-cnfs}). It was subsequently extended to a stochastic variant, which we discuss afterward. A summary of the different training settings for continuous flows is provided in \cref{tab:04-continuous-overview}.

\paragraph{Deterministic flows following an annealing path}
The goal is to learn a velocity field $v_\theta(x,t)$ such that the time marginal $\rho_{\theta,t}$—i.e., the instantaneous distribution of the solution to
\begin{align}
\label{eq:04-annealed-ode}
\frac{\dd X_t}{\dd t} = v_\theta(X_t,t), \qquad X_0 \sim \rho_0 = \pi_0,
\end{align}
matches the prescribed family distributions $\pi_t$ for all $t\in[0,1]$.
Following \cite{mateLearningInterpolationsBoltzmann2023,tianLiouvilleFlowImportance2024,fanPathGuidedParticlebasedSampling2024}, one can derive a learning objective as follows. Write $\pi_t$ as a Boltzmann distribution
\begin{align}
    \label{eq:04-annealing-path}
    \pi_t(x)=e^{-U_t(x)}/\Z_t, 
\end{align}
where $U_t$ is a known time-dependent energy (e.g. $U_{t} = t U_1 + (1-t) U_0$) and $\Z_t$ is the associated (intractable) partition function. A velocity field $v$ that exactly transports $\pi_0$ to $\pi_t$ at time $t$, satisfies the continuity equation
\begin{align}
\label{eq:04-43-continuity}
\partial_t \pi_t + \nabla \cdot (\pi_t v) = 0
\;\Rightarrow\;
\partial_t \pi_t
= -\nabla \cdot (\pi_t v)
= -\pi_t\bigl(\nabla\cdot v - \nabla U_t \cdot v\bigr),
\end{align}
where in the last equality we used $\nabla \pi_t = -\pi_t \nabla U_t$.
On the other hand, we also have
\begin{align}
\label{eq:04-43-explicit-derivative}
\partial_t \pi_t
= -\pi_t\bigl(\partial_t U_t + \partial_t \log \Z_t\bigr).
\end{align}
Comparing \cref{eq:04-43-continuity,eq:04-43-explicit-derivative} shows that any velocity field $v$ that perfectly matches the reference path must satisfy on the support of $\pi_t$ for all $t$,
\begin{align}
\label{eq:04-43-velocity-equality}
\partial_t U_t + \partial_t \log \Z_t
= \nabla \cdot v - \nabla U_t \cdot v.
\end{align}
To turn \cref{eq:04-43-velocity-equality} into a tractable training objective for $v_\theta$, we must address the fact that $\log \Z_t$ is not easy to compute. However, one can show that any triplet $(v,U,\Z)$ satisfying \cref{eq:04-43-velocity-equality} for all $x$ and $t$ necessarily also satisfies $\Z_t=\int e^{-U_t(x)}\,\dd x$ for all $t$ (see, e.g., Lemma 1 in \cite{mateLearningInterpolationsBoltzmann2023}). This motivates treating $\log \Z_t$ as an additional time-dependent function to be learned; here we parameterize it as $-F_\theta(t)$, in analogy with the free energy.\footnote{Note that \cite{mateLearningInterpolationsBoltzmann2023} instead approximate $\partial_t \log \Z_t$ directly.}
Enforcing \cref{eq:04-43-velocity-equality} by minimizing the squared residual in expectation under a user-chosen sampling distribution $\rho_t$ yields the loss
\begin{align}
\label{eq:04-43-pinn-loss}
L(\theta) = \int_0^1 \dd t \int \dd x \, \rho_t(x)\,
\left\vert
\partial_t U_t(x) - \partial_t F_\theta(t)
- \nabla \cdot v_\theta(x,t)
+ \nabla U_t(x)\cdot v_\theta(x,t)
\right\vert^2.
\end{align}
This objective corresponds to a so-called physics-informed neural network (PINN) loss \cite{raissiPhysicsinformedNeuralNetworks2019}. It involves parameterizing both the velocity field $v_\theta$ and the free energy $F_\theta$ with neural networks, and evaluating the loss by computing the required derivatives in the residual via automatic differentiation.

In practice, the choice of where to evaluate the loss is critical, since the velocity field needs to be accurate in regions where $\pi_t$ has significant probability mass and which are therefore likely to be explored during sampling. A common choice is to set $\rho_t = \mathrm{StopGrad}(\rho_{\theta,t})$,\footnote{The dependency of the argument of the $\mathrm{StopGrad}$ operator on trainable parameters is ignored during training.} focusing efforts on regions discovered during training. It has also been observed that training is more stable when one begins with concentrating on small times $t$ and progressively extends the time horizon, in a manner reminiscent of sequential tempering or of the AFT/CRAFT algorithms previously discussed for discrete-time NFs.
Another strategy to avoid starting without guidance about where the probability mass lies is to complement the learned transport with an annealed Langevin dynamics, as discussed next.

\paragraph{Langevin-driven stochastic flows}
Given the reference path $\left(\pi_t\right)_{t\in[0,1]}$, consider the annealed Langevin dynamics with time-dependent diffusion coefficient $\gamma_t$:
\begin{align}
\label{eq:04-annealed-langevin}
\dd X_t = - \gamma_t \nabla U_t(X_t)\, \dd t + \sqrt{2\gamma_t}\, \dd W_t,
\qquad X_0 \sim \pi_0.
\end{align}
Because the potential $U_t$ evolves continuously in time, this dynamics is out- of-equilibrium and does not in general follow the prescribed path $\pi_t$. However, a remarkable result from stochastic thermodynamics shows that one can modify the dynamics by adding a control term $v$ to the drift so that the resulting controlled process tracks the reference path $\pi_t$ exactly at all times \cite{vaikuntanathanEscortedFreeEnergy2008}.
This idea was revisited by \cite{vargasTransportMeetsVariational2024,albergoNETSNonequilibriumTransport2025} using modern neural networks to learn the control drift for sampling purposes.

The derivation relies on the Fokker-Planck equation. In addition to \cref{eq:04-43-continuity,eq:04-43-explicit-derivative}, the density $\pi_t$ satisfies
\begin{align}
\nabla \pi_t = -\nabla U_t \,\pi_t
\quad \Rightarrow \quad
0 = \gamma_t \bigl(\nabla \pi_t + \nabla U_t \pi_t\bigr)
\quad \Rightarrow \quad
0 = \gamma_t \nabla \cdot \bigl(\nabla \pi_t + \nabla U_t \pi_t\bigr).
\end{align}
Adding this last identity to the continuity equation \cref{eq:04-43-continuity}, satisfied by any velocity field $v$ that perfectly generates $\pi_t$, yields
\begin{align}
\partial_t \pi_t
&= -\nabla \cdot (\pi_t v)
+ \gamma_t \nabla \cdot (\nabla \pi_t + \nabla U_t \pi_t) \
= - \nabla \cdot \bigl(\pi_t (v - \gamma_t \nabla U_t)\bigr)
+ \gamma_t \Delta \pi_t,
\end{align}
which is precisely the Fokker-Planck equation associated with the controlled Langevin dynamics
\begin{align}
\label{eq:04-43-controlled-langevin}
\dd X_t
= \bigl(v(X_t,t) - \gamma_t \nabla U_t(X_t)\bigr)\,\dd t
+ \sqrt{2\gamma_t}\,\dd W_t,
\qquad X_0 \sim \pi_0.
\end{align}
Therefore, learning a control $v_\theta$ by minimizing the PINN loss \cref{eq:04-43-pinn-loss} produces a controlled Langevin dynamics \eqref{eq:04-43-controlled-langevin} that approximately follows the reference path $\pi_t$. At the beginning of training, the control is expected to be inaccurate, so the dynamics will not yet track $\pi_t$ precisely. Nevertheless, thanks to the intrinsic Langevin drift $-\gamma_t \nabla U_t$, at the same stage of training, it remains closer to the reference path than the purely deterministic ODE discussed in the previous paragraph.

\paragraph{Learning the annealing path jointly with the dynamics}
Finally, we note that within this framework it has also been proposed to learn the annealing path $(\pi_t)_{t \in [0,1]}$ itself, in addition to the velocity field and the free energy \cite{mateLearningInterpolationsBoltzmann2023,albergoNETSNonequilibriumTransport2025}. This additional layer of learning is motivated by the observation that, as for any annealing based sampling or training approach, the choice of path can have a substantial impact on sampling performance.
% , and that designing an effective path for a given target distribution is far from straightforward. 
In particular, \emph{phase transitions} along the path—namely abrupt changes in the localization or relative weights of the modes—are expected to severely hinder sampling efficiency.

In \cite{mateLearningInterpolationsBoltzmann2023}, the energy $U_t$ is treated as a trainable function of the form
\begin{align}
\label{eq:04-learned-path}
    {U_{\theta,t} = t U_1 + (1-t) U_0 + t(1-t) \hat U_{\theta,t}}
\end{align}
where $\hat U_{\theta,t}$ is parameterized by a neural network. The authors minimize the same PINN loss \cref{eq:04-43-pinn-loss} jointly over $v_\theta$, $F_\theta$, and $U_\theta$. The resulting objective is likely to admit many solutions, and it remains unclear how to systematically favor those leading to favorable sampling properties. Nevertheless, this approach constitutes a first step toward learning annealing paths automatically, and encouraging numerical results are reported in simple settings where traditional annealing schemes of geometrical averaging or temperature scheduling struggle.

In contrast, the more recent work \cite{albergoNETSNonequilibriumTransport2025} 
% \cite{albergoNETSNonequilibriumTransport2025} mention the possibility of 
advocate for learning the energy function associated with the marginal distribution of a stochastic interpolant (see \cref{eq:02-interpolation-bridge}) deriving yet again another PINN loss.
% , although without providing theoretical guarantees or numerical evaluation. 
The appeal of this choice is not accidental: stochastic interpolant bridges are typically smooth and regular, thereby avoiding the sharp transitions and mode switching that plague many traditional annealing paths. This regularity is also characteristic of diffusion models and partly underlies the considerable attention they have received for sampling applications, as discussed in the next section. Namely, we now turn to examine how diffusion models can be trained for sampling purposes.

\subsection{Diffusion models}
\label{subsec:04-diffusion}
Due to the stochastic nature of the generative process, likelihood evaluation for diffusion models is even less tractable than for continuous normalizing flows (CNFs), for which the variational and adaptive training methods presented in \cref{subsec:02-cnfs} are already of limited practical suitability. Moreover, diffusion models inherently define their own bridging path between a simple reference distribution and the target, and therefore cannot be trained to follow a prescribed annealing path as CNFs. For these reasons, diffusion models have prompted the development of a distinct family of training objectives tailored to sampling.

The reader interested in disordered statistical physics system and high-dimensional statistics should refer to \cite{alaouiSamplingSherringtonKirkpatrickGibbs2022a,montanariSamplingDiffusionsStochastic2023d,montanariPosteriorSamplingHigh2024,ghioSamplingFlowsDiffusion2024}. In these special instances, the score can be computed using message passing algorithms. We do not discuss these approaches here and focus instead on general purpose strategies.

We begin by clarifying the challenges associated with likelihood estimation in diffusion models (\cref{subsubsec:04-likelihood-diffusion}) before turning to alternative training approaches that circumvent explicit likelihood computation (\cref{subsubsec:04-monte-carlo-score,subsubsec:04-soc}). The different approaches presented are summarized in \cref{tab:04-continuous-overview}.

\subsubsection{On estimating the likelihood of diffusion models}
\label{subsubsec:04-likelihood-diffusion}
We adopt the notation introduced in \cref{subsec:02-diffusion-models} and consider a diffusion model based on the OU noising process (\cref{eq:02-noising-sde,eq:02-denoising-sde}). The instantaneous probability density of the process is denoted by $\pi_t$.
After learning a score model $s_\theta(t,x) \approx \nabla \log \pi_t(x)$, the generative procedure approximates the denoising dynamics via the reverse-time SDE:
\begin{align}
\label{eq:04-denoising-sde-model}
\dd X_t = \big(X_t + 2 s_\theta(T-t,X_t)\big)\, \dd t + \sqrt{2}\, \dd W_t,\quad 
X_0 \sim \N(0,I),
\end{align}
with instantaneous density denoted by ${\rtheta}_{t}$. One seeks to estimate ${\rtheta}_T$. The Fokker-Planck equation governing the evolution of the density ${\rtheta}_t$,
\begin{align}
\partial_t {\rtheta}_t = - \nabla \cdot \big((x + 2 s_\theta(T-t,x)) {\rtheta}_t\big) + \Delta {\rtheta}_t \, 
\end{align}
is not tractable to solve directly. To circumvent this difficulty, one can instead consider a deterministic process that shares the same time marginals as the exact noising-denoising dynamics. Such a process can be derived by examining the Fokker-Planck equation associated with the OU SDE in forward time $\tau$:
\begin{align}
\partial_\tau \pi_\tau & = - \nabla \cdot \big(-x \pi_\tau\big) + \Delta \pi_\tau \nonumber\\
& = - \nabla \cdot \big(- x\pi_\tau\big) + \nabla \cdot \big(\nabla \log \pi_\tau(x) \pi_\tau\big)\nonumber\\
& = - \nabla \cdot \big((- x - \nabla \log \pi_\tau(x))\,\pi_\tau\big), \nonumber
\end{align}
which can be interpreted as a continuity equation with velocity field $v_\tau(x) = - x - \nabla \log \pi_\tau(x)$. Replacing the true score $\nabla \log \pi_\tau$ with its learned approximation $s_\theta(\tau,x)$ and switching to reverse time $t = T - \tau$ yields the deterministic denoising process:
\begin{align}
\label{eq:04-pf-ode}
\dd X^{\dd}_t = \left(X^\dd_t + s_\theta(T-t,X^\dd_t)\right)\, \dd t, \quad X^\dd_0 \sim \N(0_d,I_d), 
\end{align}
whose instantaneous density is denoted by ${\rtheta}^\dd_t$. At this stage, the diffusion model can be view as a CNF, for which the likelihood along a trajectory $(X^\dd_t)_{t\in[0,T]}$ can be estimated by the instantaneous change of variables formula (\cref{eq:02-instantaneous-change-of-variables-a,eq:02-instantaneous-change-of-variables-b}),
\begin{align}
\log {\rtheta}^\dd_T(X^\dd_T) = \log {\rtheta}^\dd_0(X^\dd_0) - \int_0^T \nabla \cdot \left(X^\dd_t +  s_\theta(T-t,X^\dd_t)\right) \, \dd t.
\end{align}
As discussed earlier for CNFs, estimating this integral is challenging. The divergence term requires backpropagating through the score model, with a computational cost that scales with the ambient dimension. In practice, the divergence is therefore often approximated using Hutchinson's trace estimator, introducing an additional source of error that compounds with the unavoidable discretization error of the time integral. Finally, this change-of-variables formula only yields the likelihood of the deterministic denoising process ${\rtheta}^\dd_T$, which itself merely approximates the density ${\rtheta}_T$ associated with the stochastic denoising SDE. 

Several works have sought to improve the accuracy of likelihood estimation for diffusion models, motivated by different objectives. A detailed discussion of these methods is here out of scope; we only briefly highlight two representative research directions. The approach proposed in \cite{choiDensityRatioEstimation2022} introduces an auxiliary model to estimate so-called time scores $\partial_t \log {\pi}_t(x)$, which can be used to bypass the explicit computation of the divergence term in the change-of-variables formula, while remaining focused on the deterministic denoising process. In contrast, \cite{skretaSuperpositionDiffusionModels2024} applies Itô's lemma to derive an instantaneous change-of-variables formula directly along the stochastic denoising dynamics. 
Although both approaches lead to improved likelihood estimates, their accuracy remains fundamentally limited when the dimension of the problem increases. We therefore turn next to likelihood-free (and data-free) strategies for training diffusion models in the context of sampling. 

\subsubsection{Monte Carlo estimation of the score} 
\label{subsubsec:04-monte-carlo-score}
A first class of methods has sought to exploit Monte Carlo estimators of the score function $\nabla \log \pi_t$ in order to obtain diffusion models from the target distribution $\pi$ without access to a training dataset. As already mentioned in the introduction on diffusion models (\cref{subsec:02-diffusion-models}), the score function can be expressed as a conditional expectation with respect to the noising process. For the OU process, recalling \cref{eq:02-score-conditional-expectation}, we have
\begin{align}
    \label{eq:04-score-conditional-expectation}
    \nabla \log \pi_\tau(x) 
     = - \E_{x_0}\left[ \left.\frac{x - x_0 e^{-\tau}}{1 - e^{-2\tau}} \right\vert \tilde X_\tau = x  \right] \,. 
\end{align}
Constructing a Monte Carlo estimator of the score therefore amounts to sampling from $q_{0|\tau}(\cdot|x)$, the conditional distribution of the noiseless state $x_0$ given that the noising process reaches $x$ at time $\tau$. This distribution is obtained via Bayes rule and, again for the OU process, is given by  ${q_{0|\tau}(\cdot|x) \propto \pi(x_0) \N(x\,; \, x_0e^{-\tau}, \, (1-e^{-2\tau})I_{d})}$. Several estimation strategies have been proposed.

\paragraph{On-the-fly estimation}
The sampler proposed by \cite{huangReverseDiffusionMonte2024} runs the denoising dynamics of a diffusion model starting from $t=0$ and $X_0\sim \N(0,I)$, but replaces the learned score function with an on-the-fly Monte Carlo estimate of the score. This estimate is obtained by running short Langevin trajectories targeting $q_{0|T-t}(\cdot|x)$ at each time step. At denoising time $t$, the algorithm to estimate the score $\nabla \log \pi_{T-t}(x)$ proceeds as follows:

\begin{minipage}{\textwidth}
\captionof{algorithm}{\textbf{On the fly score estimation by Langevin dynamics}}
\label{alg:04-langevin-score-estimation}
\begin{algorithmic}
    \State {\textbf{Input:}} Number of Langevin steps $L$, number of Langevin chains $N$, Langevin step size $\eta$, time $t$, conditioning point $x$
     \State Initialize Langevin chains $x_0^{(0,i)}$ for $i=1\ldots N$
        \For {$\ell = 1 \ldots L$}
        \For {$i=1\ldots N$}
            \State $x_0^{(\ell,i)} = x_0^{(\ell-1,i)} + \eta \nabla \log q_{0|T-t}(x_0^{(\ell- 1,i)}|x) + \sqrt{2\eta} \, \xi_{i,\ell}$, with $\xi_{i,\ell} \sim \N(0,I)$ 
        \EndFor
        \EndFor
    \Return Score estimate $ \hat s(t, x) = - \frac{1}{N} \sum_{i=1}^N \frac{x - x_0^{(L,i)} e^{-(T-t_k)}}{1 - e^{-2(T-t_k)}}$
\end{algorithmic}
\setlength{\parskip}{\savedparskip}
\vspace{0.5em}
\end{minipage}
Using a simple Euler-Maruyama discretization of the denoising SDE, the resulting sampling algorithm for the target distribution $\pi$ is:

\begin{minipage}{\textwidth}
\captionof{algorithm}{\textbf{Reverse diffusion Monte Carlo (RDMC)}}
\label{alg:04-RDMC}
\begin{algorithmic}
    \State {\textbf{Input:}} Number of denoising steps $K$, time horizon $T$, time scheduling $\{t_k\}_{k=1}^K$ with $t_K = T$
    \State Initialize $X_{0} \sim \N(0,I)$
    \For {$k = 1 \ldots K$} 
        \State $\Delta t_k = t_k - t_{k-1}$
        \State Estimate the score $ \hat s(t_{k}, X_{k-1})$ by running \cref{alg:04-langevin-score-estimation} 
        \State Update $X_k = X_{k-1} + \big(X_{k-1} + 2 \hat s(t_k, X_{k-1})\big) \Delta t_k + \sqrt{2\Delta t_k}\, \xi_k$, with $\xi_k \sim \N(0,I)$
    \EndFor
    \Return $X_K$
\end{algorithmic}
\setlength{\parskip}{\savedparskip}
\vspace{0.5em}
\end{minipage}

\begin{figure}
    \centering
    \includegraphics[width=\textwidth]{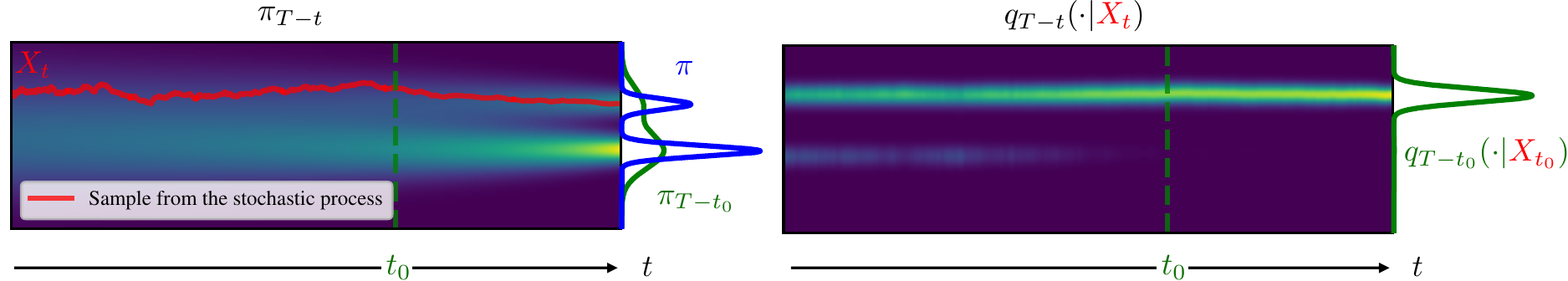}
    \caption{Illustration on a 1d multimodal target distribution $\pi$ of the transition of the marginal $\rho_{T-t}$ from being unimodal at small times $t$ to multimodal at larger times, and the reciprocal transition of the conditional distribution $q_{0|T-t}(\cdot|x)$ from being multimodal at small times to unimodal at larger times. Adapted from \cite{greniouxStochasticLocalizationIterative2024}.}
    \label{fig:04-slips}
\end{figure}

The inner loop of Langevin dynamics at each time step of the outer denoising process estimates the score on the fly, without requiring a training dataset. However, the quality of the score estimation is inherently limited by the ability of Langevin dynamics to sample accurately from $q_{0|T-t}(\cdot|x)$. When $\pi$ is multimodal, $q_{0|T-t}(\cdot|x)$ typically undergoes a transition: it is multimodal close to $t=0$ (since $q_{0|T}\approx \pi$) and becomes unimodal as $t$ increases (since the conditional distribution approaches a Gaussian centered at the conditioning $x$ rescaled by $e^{T-t}$, see illustration in \cref{fig:04-slips}). Since Langevin dynamics is known to mix poorly when targeting multimodal distributions, this algorithm must be modified when $\pi$ is multimodal.

Motivated by this observation, \cite{greniouxStochasticLocalizationIterative2024} proposed a modification of the previous algorithm. Since $q_{0|T-t}(\cdot|x)$ becomes increasingly log-concave as $t$ increases, the denoising process can be initialized at a time $t_0 > 0$ for which $q_{0|T-t_0}(\cdot|x)$ is already unimodal. However, this new starting time brings a difficulty. While starting the denoising process at $t = 0$ only requires sampling from the standard Gaussian base distribution, initializing at $t_0 > 0$ instead requires sampling from the intermediate noised marginal 
\begin{align}
    \pi_{T-t_0}(x) = \int \dd x_0 \, \pi(x_0) \, \N\left(x;\,x_0e^{-(T-t_0)}, \left(1- e^{-2(T-t_0)}\right)I\right),
\end{align}
which involves an intractable convolution of the target distribution $\pi$ with a Gaussian kernel. Nevertheless, the score of this distribution, $\nabla \log \pi_{T-t_0}(x)$, is given by the same conditional expectation as in \cref{eq:04-score-conditional-expectation} with $\tau = T-t_0$, and can therefore be estimated using \cref{alg:04-langevin-score-estimation}. Since the score of a distribution is sufficient to sample from it using Langevin dynamics, \cite{greniouxStochasticLocalizationIterative2024} proposes to run a Langevin-within-Langevin dynamics to initialize $X_{t_0}\sim\pi_{T-t_0}$, before proceeding with the same denoising procedure as in \cite{huangReverseDiffusionMonte2024}. The resulting algorithm named stochastic localization through iterative posterior sampling (SLIPS) is summarized in \cref{alg:04-SLIPS}. 

The original formulation of SLIPS uses the stochastic localization process as the outer denoising dynamics. This process can be viewed as a variant of the denoising dynamics that starts from a Dirac distribution at $t=0$ rather than from a Gaussian. Its name originates from the stochastic localization framework introduced in \cite{eldan2013thin, eldan2020taming, eldan2022analysis, chen2022localization}, where it was developed as a tool to establish results in geometric measure theory. However, the same idea can also be applied to the standard denoising dynamics used in diffusion models, leading to the algorithm described in \cref{alg:04-SLIPS}.

\begin{minipage}{\textwidth}
\captionof{algorithm}{\textbf{Stochastic localization through iterative posterior sampling (SLIPS)}}
\label{alg:04-SLIPS}
\begin{algorithmic}
    \State {\textbf{Input:}} Initialization time $t_0$, number of denoising steps $K$, time horizon $T$, time scheduling $\{t_k\}_{k=1}^K$ with $t_K = T$,
     number of Langevin steps $L$, number of Langevin chains $N$, Langevin step size $\eta$
    \State Initialize $X_{t_0}$ randomly
    \For {$\ell = 1 \ldots L$}
        \State Estimate the score $ \hat s(t_0, X_{t_0})$ by running \cref{alg:04-langevin-score-estimation} 
        \State $X_{t_0} \leftarrow X_{t_0} + \eta \big(X_{t_0} + 2 \hat s(t_0, X_{t_0})\big) + \sqrt{2\eta} \, \xi_{\ell}$, with $\xi_{\ell} \sim \N(0,I)$ 

    \EndFor
    \For {$k = 1 \ldots K$}
        \State $\Delta t_k = t_k - t_{k-1}$
        \State Estimate the score $ \hat s(t_k, X_{k-1})$ by running \cref{alg:04-langevin-score-estimation} 
        \State Update $X_k = X_{k-1} + \big(X_{k-1} + 2 \hat s(t_k, X_{k-1})\big) \Delta t_k + \sqrt{2\Delta t_k}\, \xi_k$, with $\xi_k \sim \N(0,I)$
    \EndFor
    \Return $X_K$
\end{algorithmic}
\setlength{\parskip}{\savedparskip}
\vspace{0.5em}
\end{minipage}

Starting at $t_0 > 0$ allows the algorithm to bypass the multimodality of $q_{0|T-t}(\cdot|x)$ at earlier times $t<t_0$, but it also introduces an additional source of error associated with the initialization of the denoising process. The quality of this initialization now depends on the ability of Langevin dynamics to sample from $\pi_{T-t_0}$. In contrast to the conditional distribution $q_{0|T-t}(\cdot|x)$, the distribution $\pi_{T-t_0}$ is unimodal for $t_0$ small enough (since $\pi_T \approx \N(0,I)$, see again illustration of \cref{fig:04-slips}). Hence, the approach of \cite{greniouxStochasticLocalizationIterative2024} accommodates multimodal target distributions $\pi$ at the cost of identifying a suitable starting time $t_0$ at which both $\pi_{T-t_0}$ and $q_{0|T-t_0}(\cdot|x)$ are unimodal, or more conservatively log-concave. This requirement is referred to as the \emph{duality of concavity}. Such a regime is not guaranteed to exist for all target distributions $\pi$. 
\cite{greniouxStochasticLocalizationIterative2024} identifies a class of distributions for which the duality of log-concavity condition can be satisfied and demonstrated in numerical experiments accurate sampling of multimodal distributions beyond this class. However, determining the appropriate starting time $t_0$ in general remains an open question.

As an alternative to the standard denoising score matching loss used to train diffusion models from data, \cite{huangReverseDiffusionMonte2024,greniouxStochasticLocalizationIterative2024} discussed above propose Monte Carlo estimators of the score. These approaches eliminate the need to train a neural network altogether by performing score estimation on the fly during the denoising sampling process, thereby yielding non-parametric methods. While this strategy avoids issues related to model misspecification, it precludes any form of amortization of the score estimation. A natural extension in this direction, discussed in the following paragraphs, is to regress a parametric score model onto the Monte Carlo estimates, enabling faster resampling once training is completed.  

\paragraph{Adaptive regression of a Monte Carlo estimator of the score}
Let $\hat s (x, \tau)$ be a Monte Carlo estimator of the score $\nabla \log \pi_\tau(x)$. In order to amortize the score estimation, one can train a parametric model $s_\theta(x, \tau)$ by regressing it onto $\hat s(x,\tau)$. Since it is unrealistic to expect $\hat s(x,\tau)$ to be accurate everywhere in the state space, a key question is how to select the points at which the regression loss should be evaluated. Ideally, the learned score should be accurate in regions of high probability under $\pi_{\tau}$ that are likely to be visited during the denoising process. In the absence of training data, however, achieving such localization is extremely challenging

In a philosophy closely related to the adaptive training strategies for normalizing flows described in \cref{subsubsec:04-adaptive-strategies}, the approach proposed by \cite{akhound-sadeghIteratedDenoisingEnergy2024a} leverages the availability of a small number of low-quality target samples, which in particular provide information about the location of the different modes. These initial samples are used to initialize a buffer of datapoints. Points from this buffer are used as initial conditions for noising trajectories to generate an initial set of points on which the regression loss is evaluated. The learned score model is subsequently used to generate new denoised samples, which are added to the buffer, and the procedure is then repeated. This iterative scheme allows the score model to progressively improve its accuracy in regions of high probability under $\pi_t$ that are most relevant for sampling from $\pi$. Using a mean squared regression loss, and a simple Euler-Maruyama discretization of the noising process, the resulting algorithm can be sketched as follows: 

\begin{minipage}{\textwidth}
\captionof{algorithm}{\textbf{Iterated score regression}}
\label{alg:04-iterated-score-regression}
\begin{algorithmic}
    \State {\textbf{Input:}} Initial score model parameters $\theta$, learning rate $\eta$, noising time step $\Delta \tau$, number of iterations $K$, size of batch of trajectories per iteration $B$, $x^{(0,m)}$ for $m=1\ldots M$ a few low-quality samples from $\pi$
    \State Initialize buffer $\mathcal{B} = \{x^{(0,m)}\}_{m=1}^M$
    \For {$k = 1 \ldots K$}:
        \State Sample $x^{(i)}\sim \mathcal{B}$ for $i=1\ldots B$
        \State Initialize $\tilde x^{(i)}_0 = x^{(i)}$ for $i=1\ldots B$
        \For{$\tau = 0 \ldots T$}
        \State $\tilde x^{(i)}_\tau = (1-\Delta \tau) \tilde x^{(i)}_{\tau-1}  + \sqrt{2 \Delta \tau} \, \xi_i$ with $\xi_i \sim \N(0,I)$for $i=1\ldots B$ (noising trajectories)
        \EndFor
        \State Evaluate regression loss 
        \vspace{-0.5em}
        \begin{align}
            \label{eq:04-mc-regression-loss}
            \mathcal{L}(\theta) = \frac{1}{B(T+1)} \sum_{i=1}^B \sum_{t=0}^T \|s_\theta(t, \tilde x^{(i)}_t) - \hat s(t, \tilde x^{(i)}_t)\|^2
        \end{align}
        \vspace{0.3em}
        \State $\theta \leftarrow \theta - \eta \nabla \mathcal{L}(\theta)$
        \State Sample $x^{(k,m)}$ from the diffusion model using the current score model $s_\theta$ for $m=1\ldots M$
        \State $x^{(k,m)} \leftarrow \mathrm{StopGrad}(x^{(k,m)})$ for $m=1\ldots M$ (prevent backprop. through sampling process)
        \State Add new samples to the buffer $\mathcal{B} \leftarrow \mathcal{B} \cup \{x^{(k,m)}\}_{m=1}^M$ 
    \EndFor
    \Return trained score model $s_\theta$
\end{algorithmic}
\setlength{\parskip}{\savedparskip}
\vspace{0.5em}
\end{minipage}

In their approach, \cite{akhound-sadeghIteratedDenoisingEnergy2024a} employ an importance sampling estimator of the score derived from the conditional expectation formula in \cref{eq:04-score-conditional-expectation}. In practice, however, importance sampling is not expected to scale well to high-dimensional settings. Nevertheless, the idea of amortizing score estimations through a regression model evaluated on an adaptive buffer of samples is appealing. Overall, the remaining open question in this line of research estimating the score by Monte Carlo concerns the choice of the sampler for the estimator, as both Langevin dynamics and importance sampling suffer from intrinsic limitations, as discussed above.  

\subsubsection{Revisiting variational inference with stochastic optimal control}
\label{subsubsec:04-soc}
Another line of work, which bypasses both the need to estimate likelihoods and the need for training samples when learning diffusion models, exploits the connection between diffusion models and stochastic optimal control problems \cite{tzenTheoreticalGuaranteesSampling2019,zhangPathIntegralSampler2021,bernerOptimalControlPerspective2023,vargasDenoisingDiffusionSamplers2023a,richterImprovedSamplingLearned2023a,zhangDiffusionGenerativeFlow2023a,nobleLearnedReferencebasedDiffusion2024a,havensAdjointSamplingHighly2025a}. This connection can also be interpreted through the lens of variational inference. Without aiming for full mathematical rigor, we provide below a high-level sketch of this relationship.

For simplicity, we restrict attention to the OU noising process of the target distribution $\pi$, denoted $\tilde X_\tau$ for $\tau \in [0,T]$, and to the associated exact denoising process $X_t$ for $t \in [0,T]$, respectively defined by the SDEs:
\begin{align}
    \dd \tilde X_\tau &= - \tilde X_\tau \, \dd \tau + \sqrt{2} \, \dd W_\tau, \quad \tilde X_0 \sim \pi \quad (\text{with time marginal} \pi_\tau) \\
    \dd X_t &= (X_t + 2 \nabla \log \pi_{T-t}(X_t)) \, \dd t + \sqrt{2} \, \dd W_t, \quad X_0 \sim \pi_T.
\end{align}
The trajectories $(X_t)_{t\in[0,T]}$ are distributed according to a path measure $\mathbb{P}_\pi$. The goal of sampling from $\pi$ can then be rephrased as sampling from $\mathbb{P}_\pi$ starting from $X_0 \sim \N(0,I)$ as $\pi_T \approx \N(0,I)$ for $T$ large enough. For this, 
seek to approximate $\mathbb{P}_\pi$ by a variational path measure $\mathbb{P}_\theta$, generated by an SDE similar to the denoising process but parametrized by a score model $s_\theta$:
\begin{align}
\label{eq:04-variational-path}
    \mathbb{P}_\theta \; \overset{\text{law}}{=} \; 
        \dd X_t = (X_t + 2 s_\theta(X_t, T-t)) \, \dd t + \sqrt{2} \, \dd W_t, \quad X_0 \sim \N(0,I).
\end{align}
The parameters of the score $s_\theta$ are learned, following the traditional variational inference (VI)objective, by minimizing the KL divergence between the two path measures $\mathbb{P}_\pi$ and $\mathbb{P}_\theta$. A tractable expression of this KL divergence can be derived by introducing a reference path measure $\mathbb{P}_{\text{ref}}$. This path measure is the law of the trajectories $(X^{\text{ref}}_t)_{t\in[0,T]}$ of the time-reverse of the OU noising process $\tilde X^{\text{ref}}_\tau$ initialized at $t=0$ with a distribution $\rho_\text{ref}$: 
\begin{align}
\label{eq:04-reference-path}
    % \mathbb{P}_{\text{ref}} \; \overset{\text{law}}{=} \; 
    % \begin{cases}
        \dd \tilde X^{\text{ref}}_\tau &= - \tilde X^{\text{ref}}_\tau \, \dd \tau + \sqrt{2} \, \dd W_\tau, \quad \tilde X^{\text{ref}}_0 \sim \rho_\text{ref} \quad (\text{with time marginal} \; \rho_{\text{ref},\tau})\\
        \dd X^{\text{ref}}_t &= (X^{\text{ref}}_t + 2 \nabla \log \rho_{\text{ref},T-t}(X^{\text{ref}}_t)) \, \dd t + \sqrt{2} \, \dd W_t, \quad X^{\text{ref}}_0 \sim \rho_{\text{ref},T}.
    % \end{cases}
\end{align}
The reference distribution $\rho_\text{ref}$ is chosen to be sufficiently simple so that the score of its noised marginals $\rho_{\text{ref},\tau}$ is tractable at all times, which is the case, for instance, when $\rho_\text{ref}$ is Gaussian. Reparametrizing the score of the variational path measure as ${s_\theta(x,t) = \nabla \log \rho_{\text{ref},T-t}(x) + u_\theta(x,t)}$, with $u_\theta$ being the residual learnable term, the KL divergence between $\mathbb{P}_\pi$ and $\mathbb{P}_\theta$ can then be expressed as
\begin{align}
    D_\mathrm{KL}(\mathbb{P}_\theta \| \mathbb{P}_\pi) & = \E_{\mathbb{P}_\theta} \left[\log \frac{\dd \mathbb{P}_\theta}{\dd\mathbb{P}_\pi}\right]
    = D_\mathrm{KL}(\mathbb{P}_\theta \| \mathbb{P}_{\text{ref}}) + \E_{\mathbb{P}_\theta} \left[ \log \frac{\dd \mathbb{P}_{\text{ref}}}{\dd \mathbb{P}_\pi} \right] \nonumber\\
    & = \E_{\mathbb{P}_\theta} \left[ \int_0^T  \|u_\theta(X_t, T-t)\|^2 \, \dd t + \log \frac {\rho_\text{ref}(X_T)}{e^{-U(X_T)}}\right] + \text{const.}
    \label{eq:04-kl-soc}
\end{align}
The derivation of this expression is detailed in \cref{app:soc-derivation} and involves a combination of the change-of-measure formula for SDEs and Girsanov's theorem.
The resulting training objective \cref{eq:04-kl-soc} directly reads as a stochastic optimal control problem, with $u_\theta$ playing the role of the control. The control is learned so as to minimize a cost functional balancing a running cost $\int_0^T \|u_\theta(X_t, T-t)\|^2 \, \dd t$, which penalizes deviations from the reference process, and a terminal cost $\log \frac {\rho_\text{ref}(X_T)}{e^{-U(X_T)}}$, which encourages the terminal state $X_T$ to follow the target distribution $\pi$. 

This objective can be optimized by sampling trajectories from the variational path measure $\mathbb{P}_\theta$ and evaluating the cost functional along these trajectories.  The tractability of the scores of the reference process ensures that $\mathbb{P}_\theta$ can be sampled efficiently, while the terminal cost only requires evaluating the unnormalized density of the target distribution $\pi$. This formulation provides a principled way to train diffusion models for sampling without access to a training dataset or to the likelihood of the model. 

However, this training objective is computationally expensive, as it requires re-sampling, and backpropagating through, trajectories from the variational path measure $\mathbb{P}_\theta$ 
at each training iteration to evaluate the gradients of the cost functional. To alleviate this cost, \cite{richterImprovedSamplingLearned2023a} proposes an alternative objective, termed the log-variance loss. In place of the KL--that is, the expectation under the variational measure of the logarithm of the Radon-Nikodym derivative between the variational and target path measures--, the log-variance loss is the variance of this same log-derivative with respect to an arbitrary path measure 
$\mathbb{\hat P}$:
\begin{align}
\label{eq:04-log-variance}
    D_{\mathrm{LV}}(\mathbb{P}_\theta \| \mathbb{P}_\pi) & = \mathbb{V}_{\mathbb{\hat P}} \left[\log \frac{\dd \mathbb{P}_\theta}{\dd\mathbb{P}_\pi}\right].
\end{align}
This quantity can also be re-expressed into a tractable form using the reference process (see \cref{app:soc-derivation}).
The possibility to use an arbitrary path measure $\mathbb{\hat P}$ to evaluate the variance provides additional flexibility and circumvents the need to backpropagate trough trajectories. In practice, a common choice remains $\mathbb{\hat P}=\mathrm{StopGrad}(\mathbb{P}_\theta)$\footnote{The dependence in the trainable parameters of the argument of StopGrad is ignored during learning.} as in the original KL formulation, albeit potentially re-using a buffer of trajectories generated at previous values of $\theta$. The log-variance loss is shown to exhibit lower variance and, to some extent, to be less prone to mode collapse than the traditional KL objective. 

Among other attempts to reduce computational cost, \cite{zhangDiffusionGenerativeFlow2023a} uses partial trajectories and \cite{havensAdjointSamplingHighly2025a} leverages the adjoint method to compute gradients of the KL divergence with respect to $\theta$, showing that under some conditions the adjoint-base objective can be heuristically simplified to avoid systematic resampling of trajectories at each training iteration.

Beyond these computational considerations, two additional aspects of the variational approach warrant careful examination: the parametrization of the control and the choice of the reference distribution $\rho_\text{ref}$. Concerning the control, \cite{heNoTrickNo2025} show that a so-called \emph{Langevin preconditioning}, first introduced by \cite{zhangPathIntegralSampler2021}, is necessary for the resulting generative process to properly cover the target distribution. This preconditioning consists in informing the score of the variational path measure $\mathbb{P}_\theta$ with the score of the target distribution $\pi$ following:
\begin{align}
\label{eq:04-langevin-precond}
s_\theta(x,t) = s_{\theta_1, \theta_2}(x,t) = \text{NN}_{\theta_1}(x,t) + \text{NN}_{\theta_2}(t) \nabla \log \pi(x),
\end{align}
with $\text{NN}_{\theta_1} : \mathbb{R}^d \times [0,1] \to \mathbb{R}^d$ and $\text{NN}_{\theta_2} : [0,1] \to \mathbb{R}$ neural networks initialized such that $\text{NN}_{\theta_1} \approx 0_d$ and $\text{NN}_{\theta_2} \approx 1$ at initialization. This parametrization ensures that, in the early stages of training, the denoising process closely resembles a Langevin dynamics targeting the distribution $\pi$, and therefore will progressively locate high-probability regions of the target. 

Concerning the path measure, in principle, any $\mathbb{P}_{\text{ref}}$ can be used within the variational framework, provided that the scores of the noised marginals $\rho_{\text{ref},\tau}$ remain tractable. In practice, however, the choice of $\rho_\text{ref}$ has a significant impact on training outcomes, particularly when targeting multimodal distributions, as evidenced by \cite{nobleLearnedReferencebasedDiffusion2024a}. In that work, the prescriptions of \cite{zhangPathIntegralSampler2021} and \cite{vargasDenoisingDiffusionSamplers2023a} to use a Gaussian reference distribution are shown to recover the correct mode weights of a bimodal target distribution only when the reference corresponds to the best Gaussian approximation of the target, even when using the more favorable log-variance training objective. This observation motivates an alternative strategy.  

Acknowledging the need to adapt the reference distribution to the target in order to address the challenges posed by multimodality, \cite{nobleLearnedReferencebasedDiffusion2024a} propose a pretraining strategy that learns the reference distribution $\rho_\text{ref}$ itself. As in adaptive methods for NFs and in the iterated score regression approach discussed earlier, this learning procedure leverages a small number of low-quality samples from the target distribution $\pi$, which in particular provide information about the location of the modes.
These initial samples can be used either to fit a Gaussian mixture model as $\rho_{\text{ref}}$  or to train a time-dependent energy-based model $\rho_{\text{ref},\tau}$, for both of which the scores of the noised marginals are tractable. The resulting learned reference distribution is then employed to define the variational path measure $\mathbb{P}_\theta$ and to train the control $u_\theta$, as described above. This approach is shown to significantly improve the performance of the resulting sampler on multimodal targets.\pbreak

We have now reviewed the main approaches proposed to train autoregressive models, normalizing flows—both in their discrete and continuous formulations—as well as diffusion models, in the sampling setting, where no training data from the target distribution are available a priori. 

For exact likelihood generative models, autoregressive models and discrete-time normalizing flows, the conceptual picture is essentially complete: as discussed in \cref{sec:03-exact_samplers}, once the model has been trained to approximate the target distribution, it can be embedded within sampling schemes offering rigorous convergence guarantees. 

For continuous-time models—namely continuous NFs and diffusion models—more remains to be said. The exact sampling strategies presented in \cref{sec:03-exact_samplers} rely critically on the ability to evaluate the model likelihood, a property that is readily available for discrete-time NFs but generally challenging in continuous time. This fundamental obstacle prevents a direct transposition of those constructions.
We now turn to the first proposals in the literature that aim to overcome this difficulty and to construct exact samplers leveraging the generative capabilities of continuous-time models.

\newcommand{\ditto}{\raisebox{0.4ex}{\textquotedbl}}

% \begingroup

\newcolumntype{L}[1]{>{\raggedright\arraybackslash}p{#1\textwidth}}

\begin{table}[htbp]
\centering
\footnotesize
\setlength{\tabcolsep}{4pt}
\renewcommand{\arraystretch}{1.3}

\begin{tabular}{@{}L{0.25} L{0.17} L{0.17} L{0.35}@{}}
\toprule
\textbf{Reference (acronym)} &
\textbf{Generative process} &
\textbf{Loss} &
\textbf{Additional comments}\\
\midrule

% =====================================================================
\multicolumn{4}{@{}l}{\rule{0pt}{3ex}\textbf{Annealed continuous flows --
\cref{subsec:04-annealed-cnfs} --} 
Target path: $\pi_t = e^{-U_t}/Z_t$, $U_t$ known (unless stated otherwise)}\\[0.3ex]
\midrule

\cite{mateLearningInterpolationsBoltzmann2023} &
ODE \cref{eq:04-annealed-ode} &
PINN \cref{eq:04-43-pinn-loss} & 
$U_t$ known or learned to minimize PINN loss with form \cref{eq:04-learned-path}
\\

\cite{tianLiouvilleFlowImportance2024} (LFIS) &
\ditto &
\ditto \\

\cite{fanPathGuidedParticlebasedSampling2024} (PGPS) &
\ditto &
\ditto \\

\cite{vargasTransportMeetsVariational2024} (CMCD) &
escorted SDE \cref{eq:04-43-controlled-langevin} &
\ditto \\

\cite{albergoNETSNonequilibriumTransport2025} (NETS) &
\ditto &
\ditto & 
$U_t$ possibly learned to match an interpolation process \cref{eq:02-interpolation-bridge}  \\

\midrule
% =====================================================================
\multicolumn{4}{@{}l}{\rule{0pt}{3ex}\textbf{Diffusion models --
\cref{subsec:04-diffusion}} -- Target path: noising process such as O.U. $\dd X_t = - X_t \dd t + \sqrt{2}\dd W_t$}\\[0.3ex]
\midrule

\cite{huangReverseDiffusionMonte2024} (RDMC), \cref{alg:04-RDMC} &
denoising SDE \cref{eq:04-denoising-sde-model} &
Monte Carlo \cref{eq:04-score-conditional-expectation} \\

\cite{greniouxStochasticLocalizationIterative2024} (SLIPS),  \cref{alg:04-SLIPS} &
\ditto (started at time $t_0 > 0$) &
\ditto &
robust to mode collapse when $t_0$ (allowing duality of log-concavity) is known \\

\cite{akhound-sadeghIteratedDenoisingEnergy2024a} (iDEM), \cref{alg:04-iterated-score-regression} &
\ditto &
MSE on Monte Carlo \cref{eq:04-mc-regression-loss}&
robust to mode collapse when location of mode is known to initialize buffer \\

\cite{tzenTheoreticalGuaranteesSampling2019} &
\ditto &
stochastic optimal control \cref{eq:04-kl-soc} \\

\cite{zhangPathIntegralSampler2021} (PIS) &
\ditto &
\ditto &
introduce also a Langevin preconditioning \cref{eq:04-langevin-precond} \\

\cite{bernerOptimalControlPerspective2023} &
\ditto &
\ditto \\

\cite{vargasDenoisingDiffusionSamplers2023a} (DDS) &
\ditto &
\ditto \\

\cite{zhangDiffusionGenerativeFlow2023a} (DGFS) &
\ditto &
\ditto &
use partial trajectories to limit computational cost  \\

\cite{havensAdjointSamplingHighly2025a} (adjoint sampling) &
\ditto &
\ditto &
simplify the objective to avoid sampling trajectories during optimization \\

\cite{richterImprovedSamplingLearned2023a} (DIS) &
\ditto &
log-variance \cref{eq:04-log-variance}&
log-variance loss avoids backpropagation through the sampling process \\

\cite{nobleLearnedReferencebasedDiffusion2024a} (LRDS) &
% \ditto  &
\ditto &
\cref{eq:04-kl-soc} or \cref{eq:04-log-variance} &
learned reference path \cref{eq:04-reference-path} from known mode locations to prevent mode collapse \\

\bottomrule
\end{tabular}
\caption{Overview of training methods for continuous-time generative models for sampling discussed in \cref{subsec:04-annealed-cnfs,subsec:04-diffusion}: for each
reference, the
generative process producing the samples, the training objective, and comments on the particular approach.
The mark \ditto{} indicates an entry identical to the one in the row above. 
While a systematic benchmark of all these methods is not currently available, \cite{heNoTrickNo2025} provides a comparison of several of the methods based on stochastic optimal control.}
\label{tab:04-continuous-overview}
\end{table}

% \endgroup

\pagebreak
\section{Verifiable sampling with continuous-time generative models}
\label{sec:05-cnf_fm_sampling}
As discussed previously, a key advantage of continuous-time generative models, beyond their great expressivity compared to exact likelihood models, is that they can be constructed to preserve the invariances of the target distribution (see the discussion in \cref{subsec:02-cnfs} around \cref{eq:02-equivariance-condition}). This property is particularly important in physics applications, where the distribution of interest is often invariant under groups of transformations such as rotations, translations, or permutations. In addition, for continuous normalizing flows (CNFs) and stochastic interpolants (SIs), the base distribution can be chosen freely, which—as already noted for discrete-time NFs (see paragraph Adapting the base distribution in \cref{subsec:03-tempering})—can significantly facilitate sampling. Finding ways to leverage continuous-time generative models for sampling is therefore a key question in the field which we address below. 

\emph{In this section, we return to the assumption that $\pi(x)$ is easy to compute up to a normalization constant, for any $x \in \X$, but hard to sample. We further assume that a CNF, a diffusion model or a SI has been trained to approximately sample from $\pi$.}

\subsection{Reweighing for continuous normalizing flows and stochastic interpolants}

\paragraph{Deterministic models}
A first, conceptually straightforward approach to using CNFs and SIs for verifiable sampling consists in accepting the cost of (approximately) evaluating the likelihood by integrating the instantaneous change-of-variables formula \cref{eq:02-instantaneous-change-of-variables-b}. Denoting $v_\theta$ the velocity field of the CNF or SI, the logarithm of the importance weight of a sample $X_T$, stemming from the computation of its likelihood along the generating trajectory from a base sample $X_0 \sim \rho_0$, is
\begin{align}
    \label{eq:05-importance-weight-cnfs}
   \log w(X_T) = \log \frac{\pi(X_T)}{\rzero(X_0)} \,  + { \, \int_0^T \nabla \cdot v_\theta(X_t,t) \, \dd t } \, .
\end{align}

This strategy was first explored by \cite{kohlerEquivariantFlowsExact2020}, who used a CNF to sample interacting particle systems with pairwise interactions. Their velocity field architecture relies on a decomposition into radial basis functions of pairwise distances and is specifically designed to be equivariant and so that the divergence of the vector field can be computed efficiently via automatic differentiation (requiring a single backward pass rather than as many passes as the dimension of the field in a naive implementation).
Thereby, \cite{kohlerEquivariantFlowsExact2020} were able to successfully reweigh CNF-generated configurations to the target Boltzmann distribution for simple interacting particle systems. In contrast, they showed that replacing the exact divergence with a Hutchinson estimator—commonly used in the CNF literature to reduce computational cost—yields inaccurate importance weights and significantly degrades sampling performance.

Building on the same architecture, \cite{jungNormalizingFlowsEnhanced2024a} considered more challenging systems, in particular amorphous particle systems at decreasing temperatures approaching the glass transition. To address the increased complexity, they selected as base distribution a higher-temperature Boltzmann distribution rather than the centered Gaussian used in \cite{kohlerEquivariantFlowsExact2020}. Their promising results showed performance comparable to PT and SMC in the difficult regime near the glass transition. 

Turning to SIs, \cite{greniouxRiemannianStochasticInterpolants2025} also investigated amorphous particle systems, examining in detail how to make SIs invariant to the symmetries of
multispecies systems in periodic boxes. Compared to CNFs, SIs rely on a different training objective that is more computationally efficient as it does not require likelihood evaluation. This removes the architectural constraints imposed in \cite{kohlerEquivariantFlowsExact2020} to speed up divergence computation during training and allows for more expressive equivariant velocity fields. However, the absence of the architectural constraints renders the weight computation during reweighing more costly.
Accepting this additional cost, \cite{greniouxRiemannianStochasticInterpolants2025} succesfully sampled amorphous particle systems with SIs. Moreover, the increased expressivity of the velocity field allowed them to use a uniform distribution on a periodic box as a base measure. This avoids the need to generate base samples via MCMC when using a high-temperature Boltzmann distribution as in \cite{jungNormalizingFlowsEnhanced2024a}.

\paragraph{Reweighing and resampling for (stochastic) annealed flows}
The importance reweighing strategy described above can be revisited in the context of annealed continuous flows, presented in \cref{subsec:04-annealed-cnfs}, trained to follow a path $\left(\pi_t\right)_{t \in [0,1]}$ between the base distribution $\rho_0$ and the target distribution $\pi$. This is the proposition of \cite{tianLiouvilleFlowImportance2024,vargasTransportMeetsVariational2024,albergoNETSNonequilibriumTransport2025}. 

Adopting again the Boltzmann representation $\pi_t = e^{-U_t}/\mathcal{Z}_t$, recall that the generative dynamics of such a (possibly stochastic) CNF with control-velocity field $v_\theta$ 
are given by (see \cref{eq:04-43-controlled-langevin})
\begin{align}
\dd X_t = v_\theta(X_t,t)\,\dd t
- \sigma_t \nabla U_t(X_t)\,\dd t
+ \sqrt{2\sigma_t}\,\dd W_t , \quad X_0 \sim \rho_0 \, .
\end{align}
The time-dependent diffusion coefficient $\sigma_t$ vanishes for deterministic CNFs and is strictly positive for stochastic ones.
The importance weight associated with $X_t$ from a trajectory $(X_s)_{s \in [0,t]}$ for sampling from $\pi_t$ can then be expressed as
\begin{align}
   \log w_t\left(\left(X_s\right)_{s \in [0,t]}\right) = 
     - \int_0^t \Big(\partial_s U_s(X_s) + v_\theta(X_s, s) \cdot \nabla U_s(X_s) + \nabla \cdot v_\theta(X_s,s)\Big) \, \dd s \,.
\end{align}
This expression can be derived by manipulating the Fokker-Planck equation associated with the controlled Langevin dynamics (see, e.g., the proof of Proposition 2.4 in Appendix A of \cite{albergoNETSNonequilibriumTransport2025}).
Importantly, the weight now depends explicitly on the entire trajectory, which is itself a random realization of the stochastic dynamics, rather than solely on the terminal configuration. This contrasts with direct reweighing for deterministic CNFs or SIs, where the trajectory is fully determined by its endpoint once the velocity field is fixed. In the deterministic case $\sigma_t = 0$, the expression above reduces to the previously obtained formula \cref{eq:05-importance-weight-cnfs}, as the first two terms in the integrand combine into the material derivative of the potential energy function along the trajectory. 

Conceptually, this annealed approach is closely related to the SMC-inspired (CR)AFT algorithm previously discussed for discrete-time flows \cite{arbelAnnealedFlowTransport2021,matthewsContinualRepeatedAnnealed2022a}. Whereas (CR)AFT relies on a finite composition of transport maps interleaved with MCMC steps, continuous-time annealed flows incorporate both ingredients directly into their dynamics: a transport term, $v_\theta(X_t,t)\,\dd t$, and a Langevin diffusion term, $-\sigma_t \nabla U_t(X_t)\,\dd t + \sqrt{2\sigma_t}\,\dd W_t$.
In direct analogy with (CR)AFT and more generally with SMC, the importance weights can be used to perform resampling at intermediate times along the trajectory. Resampling may be triggered when the ESS falls below a prescribed threshold, thereby mitigating weight degeneracy and improving overall sampling performance.

Nevertheless, the need to evaluate the divergence of the velocity field to compute the weights remains in this annealed formulation. This is the key computational bottleneck for CNFs and SIs. Restrictive architectures enable efficient divergence computation but may lack sufficient flexibility for complex targets, while more expressive models incur a significant computational cost during reweighing.
Moreover, an uncontrolled source of bias is the eventual discretization of the time integrations. 
A couple of recent works have proposed to circumvent both of these issues by reinterpreting the generative process of diffusion models as a stochastic normalizing flow, which allows for a principled construction of exact samplers without any likelihood evaluation. We present these approaches next.

\subsection{Diffusion samplers as stochastic flows}
\label{subsec:05-diffusion-samplers-as-stochastic-flows}
To avoid computing likelihoods or importance weights involving the divergence of the velocity field, one possible route to debias diffusion samples is to reinterpret the generative process as a discrete-time stochastic normalizing flow (see paragraph Combining NF with AIS: stochastic normalizing flows in \cref{subsec:03-tempering}). This perspective underlies the approaches of \cite{phillipsParticleDenoisingDiffusion2024a,zhangEfficientUnbiasedSampling2025} describing a reweighing à la annealed importance sampling, and of \cite{chenMarkovChainMonte2026} which proposes a Markov chain Monte Carlo (MCMC) sampler based on the super-detailed balance condition already encountered in the context of coarse-grainings (\cref{subsec:03-coarse-graining}).

We present here the main idea, focusing on the backward SDE of the OU process. Using the approximate score function $s_\theta$ learned by the diffusion model, the backward dynamics reads
\begin{align}
\dd X_t = \left( X_t + 2 s_\theta(X_t, T-t) \right) \, \dd t + \sqrt{2} \, \dd W_t \, , \quad X_0 \sim \rho_0 \, .
\end{align}
Discretizing the time interval $[0,T]$ with a grid $0 = t_0 < t_1 < \dots < t_K = T$, the joint distribution of the states $(x_k)_{k=0}^K$ visited by the Markovian backward process factorizes as
\begin{align}
   \label{eq:05-diffusion-as-snf}
\rho_\theta(x_0, \ldots, x_K) = \rho_0(x_0) \prod_{k=1}^K q_{\theta,k}(x_k \mid x_{k-1}).
\end{align}
Although the transition kernel $q_{\theta,k}$ is non-trivial due to the non-linear score function, it can be approximated by a Gaussian distribution obtained from a discretization of the backward SDE.
Going beyond the naive Euler-Maruyama scheme, \cite{phillipsParticleDenoisingDiffusion2024a} propose using an exponential integrator yielding the Gaussian transition kernel
\begin{align}
q_{\theta,k}(x_k \mid x_{k-1})
= \mathcal{N}\left(
x_k;\,
e^{-\Delta t_k} x_{k-1} + \mu_k,\,
(1-e^{-2\Delta t_k}) I_d
\right),
\end{align}
with $\Delta t_k = t_k - t_{k-1}$ and
$
\mu_k = 2(1-e^{-\Delta t_k})\bigl(x_{k-1} + s_\theta(x_{k-1}, t_{k-1})\bigr)$.\footnote{This expression follows from rewriting the backward SDE as $\dd X_t = - X_t ,\dd t + 2(X_t + s_\theta(X_t, T-t)),\dd t + \sqrt{2},\dd W_t$, applying an exponential integrator to the linear term $-X_t$, and treating $X_t + s_\theta(X_t,T-t)$ as frozen over the time step. This decomposition is consistent with the parametrization adopted in \cite{phillipsParticleDenoisingDiffusion2024a}, where the network directly models $x + s_\theta(x,t)$.}
This construction completes the definition of the stochastic flow defined in \cref{eq:05-diffusion-as-snf}.

To debias the samples generated by this stochastic flow, one considers the forward OU process started from the target distribution $\pi$, discretized on the same time grid. This forward process is exactly tractable and defines the joint distribution
\begin{align}
\pi(x_0, \ldots, x_K)
= \pi(x_K) \prod_{k=1}^K p_k(x_{k-1} \mid x_k),
\end{align}
where
\begin{align}
p_k(x_{k-1} \mid x_k)
= \mathcal{N}\left(
x_{k-1};\,
e^{-\Delta t_k} x_k\,,\,
(1-e^{-2\Delta t_k}) I_d
\right).
\end{align}
Let's examine two alternative strategies using this construction to debias the samples generated by the stochastic flow.

\paragraph{Annealed importance sampling along the diffusion path}
The first strategy, proposed by \cite{phillipsParticleDenoisingDiffusion2024a} and \cite{zhangEfficientUnbiasedSampling2025}, is to compute importance weights for the generated samples using the joint distributions of the forward and backward processes. The importance weights are given by the ratio of these two joint distributions:
\begin{align}
w(x_0, \ldots, x_K) = \frac{\pi(x_0, \ldots, x_K)}{\rho_\theta(x_0, \ldots, x_K)}.
\end{align}
Crucially, this ratio only involves Gaussian transition kernels and the target density $\pi(x_K)$. Therefore, it avoids any explicit likelihood computation or evaluation of divergence terms. In the overwhelmingly common case where $\pi$ is only known up to a normalizing constant, the self-normalized weights and estimators are used in practice (see e.g. \cref{alg:03-neural-IS}.)

Following this paradigm, \cite{phillipsParticleDenoisingDiffusion2024a} further develop a SMC version of the stochastic flow, incorporating intermediate resampling steps along the trajectory. For this, they not only learn the score of the noising process but also the time marginal densities $\pi_t$ along the trajectory, which allows them to compute the importance weights before the end of the trajectory and thus to perform resampling at intermediate times. As usual, resampling is triggered whenever the effective sample size (ESS) of the importance weights falls below a prescribed threshold.

In contrast, \cite{zhangAcceleratedParallelTempering2025} sticks to the vanilla AIS version of the reweighting in the joint space and considers a variance-exploding diffusion (i.e., a Wiener process) instead of the OU process. This modeling choice leads to a different approximation of the denoising transition kernels. Importantly, they propose to learn the variance of these kernels so as to directly minimize the variance of the resulting importance weights, yielding substantial improvements in practice.

\paragraph{Markov chain Monte Carlo with super-detailed balance}
Alternatively, an MCMC sampler can be constructed using a Metropolis-Hastings accept/reject construction where the proposal consists in a forward-backward step of the stochastic flow \cite{chenMarkovChainMonte2026}, similarly to the proposition of \cite{nealSamplingMultimodalDistributions1996} using a traditional annealing scheme. 

Using the same notation as above, an updated sample $x'$ is proposed from a current sample $x$ by first sampling a forward trajectory $(x_K=x, \ldots, x_0)$ from the forward noising kernel and then sampling the backward trajectory from the stochastic flow $(x_{1}', \ldots, x_K'=x')$ from the backward kernels. The resulting proposal distribution is
\begin{align}
\prod_{k=1}^K p_k(x_{k-1} \mid x_k) \delta(x_0'-x_0) \prod_{k=1}^K q_{\theta,k}(x_k' \mid x_{k-1}').
\end{align}
and the acceptance probability of the proposed sample $x'$ involves the ratio of the joint distribution of the entire trajectories reaching $x'$ from $x'$ and the joint distribution of the entire trajectories reaching $x$ from $x$, traversing the same intermediate states:
\begin{align}
\alpha(x,x') = \min\left(1,
 \frac{\pi(x')  \prod_{k=1}^K p_k(x_{k-1}' \mid x_k') \delta(x_0-x_0') \prod_{k=1}^K q_{\theta,k}(x_k \mid x_{k-1})
 }
 {
 \pi(x) \prod_{k=1}^K p_k(x_{k-1} \mid x_k) \delta(x_0'-x_0) \prod_{k=1}^K q_{\theta,k}(x_k' \mid x_{k-1}')}.
 \right),
\end{align}
This acceptance ratio relies on the super-detailed balance condition already encountered in the context of coarse-grainings (\cref{subsec:03-coarse-graining}) which guarantees the convergence to the target distribution $\pi$, without requiring any likelihood evaluation or importance weights. By updating the state through a full forward-backward trajectory, one expects to generate proposals that are more decorrelated from the current state than traditional local moves, potentially improving the mixing of the Markov chain. Note that one could imagine taking an additional random step in the latent space of the stochastic flow before proposing the backward trajectory, in the same spirit as neutraMCMCs for discrete normalizing flows (\cref{subsubsec:03-neutra-MCMC}).

Overall, interpreting diffusion models through the lens of stochastic flows provides a principled framework for exact sampling  without requiring costly divergence computations. However, in order to maintain importance weights with controlled variance, or high acceptance rates in MCMC, the joint laws over the discretized forward and backward processes must be sufficiently close. For this, the time discretization typically needs to become finer as the dimensionality of the problem increases. This refinement can significantly raise the computational cost, potentially limiting scalability in high-dimensional settings. 

\subsection{Free energy estimation with continuous-time generative models}

A related problem to sampling is the estimation of free energies, which can also be tackled with continuous-time generative models. Writing the target distribution in Boltzmann form,
$\pi(x) = e^{-U(x)}/\Z$ with $U$ the potential energy and $\Z$ the unknown partition function, the free energy is defined as $F = - \log \Z$.
Below we review two recent approaches to free energy estimation with continuous-time generative models, both assuming access to samples from the target distribution $\pi$ and a reference distribution $\rho_0$ with known free energy $F_0 = - \log \Z_0$. 

\paragraph{Neural thermodynamic integration}
Thermodynamic integration, first introduced in \cite{kirkwoodStatisticalMechanicsFluid1935}, computes the free energy difference between $\pi$ and a reference distribution $\rho_0$, provided a bridge of distributions $\pi_t = e^{-U_t}/\Z_t$ with $t \in [0,1]$ such that $\pi_1 = \pi$ and $\pi_0 = \rho_0$, and remarking that the free energy difference can be expressed as
\begin{align}
   \label{eq:05-thermodynamic-integration}
   F - F_0 = \log \frac{\Z}{\Z_0} = \int_0^1 \E_{\pi_t}[\partial_t U_t(X)] \, \dd t.
\end{align}
The computational bottleneck of this approach is the need to sample from the intermediate distributions $\pi_t$ along the path. \cite{mateNeuralThermodynamicIntegration2024a} propose to use a diffusion model converging to simple distribution $\rho_0 = e^{-U_0}/\Z_0$ with known free energy $F_0 = - \log \Z_0$ to build the bridge and subsequently sample from the intermediate distributions $\pi_t$.

The protocol is as follows. First, define a forward noising process $(\tilde X_\tau)_{\tau \in [0,1]}$ easily simulated and that approximately converges at time $\tau=1$ to the simple distribution $\rho_0$, $\pi_\tau$ being the time marginal of this process. Then, using denoising score matching (\cref{eq:02-denoising-score-matching-loss}), train a diffusion model to learn a score function $s_\theta(x,\tau)$ parametrized through a time and space dependent energy function $U_\theta$ such that 
\begin{align}
   s_\theta(x,\tau) = - \nabla_x U_\theta(x,1-\tau) \approx \nabla_x \log \pi_\tau(x).
\end{align}
The energy function $U_\theta$ is typically preconditioned with the initial and final time marginals of the noising process: 
\begin{align}
   U_\theta(x,t) = t(1-t) \hat U_\theta(x,t) + (1-t) U_0(x) + t U(x),
\end{align}
where $\hat U_\theta$ is the learnable component of the energy function, parametrized as a neural network. 
The free energy difference can then be computed following \cref{eq:05-thermodynamic-integration}, with $U_t = U_\theta(\cdot, t)$ and by sampling from the intermediate distributions $\pi_t$ using the forward noising process. 

Demonstration of the approach were made on Lennard-Jones particle systems
\cite{mateNeuralThermodynamicIntegration2024a} and on solvation free energies \cite{mateSolvationFreeEnergies2025}. However, two sources of error are difficult to control: the error in the learned score function that does not necessarily match the true score of the intermediate distributions, and the discretization error in the time integration of \cref{eq:05-thermodynamic-integration}. The method presented next circumvents these two issues.

\paragraph{Free energy estimators with adaptive transports}
To build a free energy estimator, \cite{duFEATFreeEnergy2025} combines several ideas from the previous sections. First, they train a stochastic interpolant to bridge the base distribution $\rho_0$ and the target distribution $\pi$. Then, they use the learned stochastic interpolant to define a forward and backward discrete stochastic flow, which allows them to compute the free energy difference between $\rho_0$ and $\pi$ with an annealed importance sampling strategy \cite{nealAnnealedImportanceSampling2001} using the joint distributions of the forward and backward flows. The method called Free energy estimators with adaptive transports is abbreviated as FEAT.

For learning, they rely on a stochastic interpolation process between samples from the reference $x_0\sim\rho_0$ and samples from the target $x_1\sim\pi$: $I_t(x_0, x_1) + \gamma(t) z$ as introduced in \cref{subsec:02-stochastic-interpolants} \cref{eq:02-interpolation-process}, with $z \sim \N(0, I_d)$ and $\gamma(t)$ a time-dependent diffusion coefficient. They fit both a velocity field $v_{\theta_1}$ using the SI loss \cref{eq:02-SI-loss} and the score function $s_{\theta_2}$ (see \cite{albergoBuildingNormalizingFlows2023} Corollary 2.9), parametrizing the latter through a time and space dependent energy function $U_{\theta_2}$ such that
$s_{\theta_1}(x,t) = - \nabla_x U_{\theta_1}(x,t)$. The additional fit of the score function allows them to define stochastic forward and backward process following approximately the path defined by the interpolation process:
\begin{align}
   \dd X_t &=  v_{\theta_1}(X_t, t) \dd t + \sigma_t s_{\theta_2}(X_t,t) \, \dd t + \sqrt{2\sigma_t} \, \dd W_t, \quad X_0 \sim \rho_0 \\
   \dd \tilde X_\tau &= - v_{\theta_1}(\tilde X_\tau, 1-\tau) \dd \tau + \sigma_{1-\tau} s_{\theta_2}(\tilde X_\tau, 1-\tau) \, \dd \tau + \sqrt{2\tilde \sigma_{1-\tau}} \, \dd \tilde W_\tau, \quad \tilde X_0 \sim \pi.
\end{align}
Discretizing these processes using a simple Euler-Maruyama scheme over a grid $t_0 = 0 < t_1 < \ldots < t_K = 1$ with step size $\Delta t$ yields the forward and backward joint distributions: 
\begin{align}
   q^{+}_{\theta_1, \theta_2}(x_0, \ldots, x_K) & = \rho_0(x_0) \prod_{k=1}^K \N(x_k; \mu_{\theta_1, \theta_2}^+(x_{k-1}, t_{k-1}), 2 \sigma_{t_{k-1}} \Delta t I_d) \\
   q^{-}_{\theta_1, \theta_2}(x_0, \ldots, x_K) & = \pi(x_K) \prod_{k=1}^K \N(x_{k-1}; \mu_{\theta_1, \theta_2}^-(x_k, t_k), 2  \sigma_{t_k} \Delta t I_d)
\end{align}
where
$\mu_{\theta_1, \theta_2}^+(x_{k-1}, t_{k-1}) = x_{k-1} + (v_{\theta_1}(x_{k-1}, t_{k-1}) + \sigma_{t_{k-1}} s_{\theta_2}(x_{k-1}, t_{k-1}) ) \Delta t$ and\\
${\mu_{\theta_1, \theta_2}^-(x_k, t_k) = x_k + (- v_{\theta_1}(x_k, t_k) +\sigma_{t_k} s_{\theta_2}(x_k, t_k) ) \Delta t}$.
The free energy difference can then be computed as the log of the expected ratio of the forward and backward joint distributions:
\begin{align}
   F - F_0 = \log \frac{\Z}{\Z_0} = \log \E_{q^{-}_{\theta_1, \theta_2}} \left[ 
      \frac{q^{+}_{\theta_1, \theta_2}(x_0, \ldots, x_K)}{q^{-}_{\theta_1, \theta_2}(x_0, \ldots, x_K)} \right] = - \log \E_{q^{+}_{\theta_1, \theta_2}} \left[ 
      \frac{q^{-}_{\theta_1, \theta_2}(x_0, \ldots, x_K)}{q^{+}_{\theta_1, \theta_2}(x_0, \ldots, x_K)}
   \right]\, .
\end{align}
This strategy aligns with the debiasing of diffusion models presented in \cref{subsec:05-diffusion-samplers-as-stochastic-flows}, where the forward and backward joint distributions are used to compute importance weights. In the present case, the ratio of the forward and backward joint distributions is used to compute the free energy difference. Compared to the neural thermodynamic integration approach, this approach is unbiased for any choice of discretization step size $\Delta t$ and quality of the learned velocity field and score function. However, the more accurate the learned velocity field and score function, the closer the forward and backward joint distributions are, and the lower the variance of the ratio.

This FEAT strategy was benchmarked on condensed matter systems and found competitive against free energy estimation methods using discrete-time flows (see targeted free energy perturbation \cref{eq:03-tfep-estimator} and Bennett Acceptance Ratio \cref{eq:03-bar-implicit equation}) \cite{schebekAssessingGenerativeModeling2025}.

\pbreak

This discussion on free energy estimation concludes our review of emerging methods for verifiable sampling with continuous-time generative models. We have seen that while direct reweighing of continuous normalizing flows samples is conceptually straightforward, it faces a fundamental computational bottleneck due to the need to evaluate the divergence of the velocity field. This bottleneck can be circumvented by reinterpreting continuous-time models as stochastic flows, as was proposed for diffusion models and stochastic interpolants. However, this approach introduces its own challenges, particularly in terms of controlling the variance of importance weights in high-dimensional settings. While a complete benchmark of the different approaches is still lacking, \cite{greniouxDiffusionbasedAnnealedBoltzmann2026} provide a preliminary comparison of the different strategies for debiasing diffusion samples, highlighting the trade-offs between computational cost and accuracy. Overall, these developments allowing to leverage the power of continuous-time generative models represent a significant step forward in the quest for efficient and verifiable sampling methods in high-dimensional spaces.

\pagebreak
\section{Perspectives}
\label{sec:07-perspectives}

\paragraph{Where do we stand?}
Proof of concepts of applications of generative-model–assisted samplers to scientific problems have been reported in the literature in various domains. To cite a few: in lattice quantum chromodynamics \cite{cranmerAdvancesMachinelearningbasedSampling2023}, in
disordered spin systems \cite{wuSolvingStatisticalMechanics2019,bonoDemonstratingRealAdvantage2025}, in diverse phases of condensed matter with crystals \cite{schebekScalableBoltzmannGenerators2025}, liquids \cite{corettiLearningMappingsEquilibrium2025}, and armorphous materials \cite{jungNormalizingFlowsEnhanced2024a,greniouxRiemannianStochasticInterpolants2025}, as well as in classical simulations of biomolecules \cite{noeBoltzmannGeneratorsSampling2019,invernizziSkippingReplicaExchange2022} and ab initio simulation of small molecules \cite{molina-tabordaActiveLearningBoltzmann2024a}, or again the Bayesian inference of gravitation wave events parameters \cite{wongFastGravitationalwaveParameter2023,woutersRobustParameterEstimation2024}.

However, the practical impact of these methods on scientific applications remains limited. In particular, generative-model–assisted samplers have not yet led to scientific discoveries that would have been unattainable with existing methodologies. 
Nevertheless, they introduce a fundamentally different mechanism from traditional enhanced sampling techniques and therefore offer a complementary perspective that deserves careful exploration. In particular, these methods often propose configurations that effectively jump across low-probability regions rather than traversing them through a sequence of intermediate states. This qualitative difference in how state space is explored may open new opportunities for tackling sampling problems characterized by pronounced metastability.

As discussed throughout this tutorial review, the field is evolving rapidly, with many recent proposals extending the earliest ideas. A particularly promising direction consists in combining generative models with established sampling strategies, such as tempering approaches or collective-variable–guided sampling methods. In addition, the noising–denoising paradigm introduced by diffusion models provides a novel way of constructing annealing paths that may circumvent the phase transitions that often hinder traditional tempering strategies. More broadly, the rapid progress in the design and training of generative models naturally fuels optimism about their future impact on scientific sampling problems.

A central question, however, is whether the computational cost associated with training a generative model is ultimately offset by the gains in sampling efficiency. The answer is necessarily problem-dependent and remains difficult to assess in general. Importantly, such an evaluation should also account for the design cost of a sampling strategy. While deep generative models involve numerous hyperparameters and architectural choices, designing an efficient traditional sampler is also often highly problem-specific and may require substantial expert knowledge. One may therefore hope that machine-learning–based samplers could eventually offer more automated and less system-dependent solutions. Whether this promise will materialize remains an open question and is closely related to the issue of transferability discussed below.

Finally, an important challenge lies in transferring the many methodological advances proposed in recent years—often developed and evaluated on toy benchmarks—to realistic scientific applications. Achieving this goal requires sustained efforts and close collaborations between machine-learning researchers and domain scientists.

\paragraph{What we did not cover} 
Several important directions have not been covered in detail in this tutorial review. Let us briefly discuss two of them to conclude.

The ability of generative models to characterize equilibrium states without explicitly resolving the transition regions between them is often presented as a strength. However, many scientific questions concern dynamical properties rather than equilibrium statistics alone. Several recent works have begun to address this challenge, particularly in the context of molecular simulations. Proposed approaches include learning transition-path ensembles with normalizing flows \cite{asgharEfficientRareEvent2024}, learning transition-state distributions to initialize transition path sampling \cite{falknerConditioningBoltzmannGenerators2023a}, and learning effective transition operators to accelerate molecular dynamics simulations \cite{kleinTimewarpTransferableAcceleration2023,schreinerImplicitTransferOperator2023a,diezTransferableGenerativeModels2025}.

Finally, returning to the question of amortizing training costs, an important objective is to design models that do not need to be retrained for every new target distribution. Instead, one would ideally train a generative model once and then reuse it across a family of related systems. This idea, often referred to as \emph{transferability}, has recently begun to receive attention, particularly in molecular modeling applications \cite{kleinTimewarpTransferableAcceleration2023,kleinTransferableBoltzmannGenerators2024a,diezTransferableGenerativeModels2025,lewisScalableEmulationProtein2025}. Developing transferable generative samplers capable of generalizing across related systems remains a key open challenge for the field.

\phantomsection                                    % hyperref: anchor for the link
\section*{Acknowledgments}
\addcontentsline{toc}{section}{Acknowledgments}
This tutorial review is based on my thesis for the "habilitation à diriger des recherches" defended at the École normale supérieure, PSL Research University, in July 2026. I was very lucky along the years to be able to collaborate with many talented researchers and students who have nourished my understanding of this field. I am greatly indebted to all of them. I would also like to thank in particular Tony Lelièvre, Guilhem Semerjian and Francesco Zamponi for their feedback on the draft, and to Luigi Fogliani and Baran Celik for carefully spotting a number of typos.

This work was supported by the French Agence Nationale de la Recherche under the project SWiNGs (ANR-25-CE23-2025). 

\printbibliography[heading=bibintoc]

@article{tabakDensityEstimationDual2010,
  title = {Density Estimation by Dual Ascent of the Log-Likelihood},
  author = {Tabak, Esteban G. and {Vanden-Eijnden}, Eric},
  year = 2010,
  journal = {Communications in Mathematical Sciences},
  volume = {8},
  number = {1},
  pages = {217--233},
  issn = {15396746, 19450796},
  doi = {10.4310/CMS.2010.v8.n1.a11}
}

@article{papamakariosNormalizingFlowsProbabilistic2021a,
  title = {Normalizing {{Flows}} for {{Probabilistic Modeling}} and {{Inference}}},
  author = {Papamakarios, George and Nalisnick, Eric and Rezende, Danilo Jimenez and Mohamed, Shakir and Lakshminarayanan, Balaji},
  year = 2021,
  journal = {Journal of Machine Learning Research},
  volume = {22},
  number = {57},
  pages = {1--64},
  issn = {1533-7928}
}

@article{kobyzevNormalizingFlowsIntroduction2021,
  title = {Normalizing {{Flows}}: {{An Introduction}} and {{Review}} of {{Current Methods}}},
  shorttitle = {Normalizing {{Flows}}},
  author = {Kobyzev, Ivan and Prince, Simon J.D. and Brubaker, Marcus A.},
  year = 2021,
  month = nov,
  journal = {IEEE Transactions on Pattern Analysis and Machine Intelligence},
  volume = {43},
  number = {11},
  pages = {3964--3979},
  issn = {0162-8828, 2160-9292, 1939-3539},
  doi = {10.1109/TPAMI.2020.2992934}
}

@inproceedings{dinhDensityEstimationUsing2017a,
  title = {Density Estimation Using {{Real NVP}}},
  booktitle = {International {{Conference}} on {{Learning Representations}}},
  author = {Dinh, Laurent and {Sohl-Dickstein}, Jascha and Bengio, Samy},
  year = 2017
}

@inproceedings{durkanNeuralSplineFlows2019a,
  title = {Neural {{Spline Flows}}},
  booktitle = {Advances in {{Neural Information Processing Systems}}},
  author = {Durkan, Conor and Bekasov, Artur and Murray, Iain and Papamakarios, George},
  year = 2019,
  volume = {32},
  publisher = {Curran Associates, Inc.}
}

@article{brenierPolarFactorizationMonotone1991,
  title = {Polar Factorization and Monotone Rearrangement of Vector-Valued Functions},
  author = {Brenier, Yann},
  year = 1991,
  journal = {Communications on Pure and Applied Mathematics},
  volume = {44},
  number = {4},
  pages = {375--417},
  issn = {1097-0312},
  doi = {10.1002/cpa.3160440402},
  copyright = {Copyright \copyright{} 1991 Wiley Periodicals, Inc., A Wiley Company}
}

@inproceedings{chenNeuralOrdinaryDifferential2018,
  title = {Neural {{Ordinary Differential Equations}}},
  booktitle = {Advances in {{Neural Information Processing Systems}}},
  author = {Chen, Ricky T. Q. and Rubanova, Yulia and Bettencourt, Jesse and Duvenaud, David K},
  year = 2018,
  volume = {31},
  publisher = {Curran Associates, Inc.}
}

@inproceedings{kohlerEquivariantFlowsExact2020,
  title = {Equivariant {{Flows}}: {{Exact Likelihood Generative Learning}} for {{Symmetric Densities}}},
  shorttitle = {Equivariant {{Flows}}},
  booktitle = {Proceedings of the 37th {{International Conference}} on {{Machine Learning}}},
  author = {K{\"o}hler, Jonas and Klein, Leon and Noe, Frank},
  year = 2020,
  month = nov,
  pages = {5361--5370},
  publisher = {PMLR},
  issn = {2640-3498}
}

@inproceedings{garciasatorrasEquivariantNormalizingFlows2021,
  title = {E(n) {{Equivariant Normalizing Flows}}},
  booktitle = {Advances in {{Neural Information Processing Systems}}},
  author = {Satorras, Victor and Hoogeboom, Emiel and Fuchs, Fabian and Posner, Ingmar and Welling, Max},
  year = 2021,
  volume = {34},
  pages = {4181--4192},
  publisher = {Curran Associates, Inc.}
}

@article{midgleySEEquivariantAugmented2023a,
  title = {{{SE}}(3) {{Equivariant Augmented Coupling Flows}}},
  author = {Midgley, Laurence and Stimper, Vincent and Antor{\'a}n, Javier and Mathieu, Emile and Sch{\"o}lkopf, Bernhard and {Hern{\'a}ndez-Lobato}, Jos{\'e} Miguel},
  year = 2023,
  month = dec,
  journal = {Advances in Neural Information Processing Systems},
  volume = {36},
  pages = {79200--79225}
}

@misc{geigerE3nnEuclideanNeural2022,
  title = {E3nn: {{Euclidean Neural Networks}}},
  shorttitle = {E3nn},
  author = {Geiger, Mario and Smidt, Tess},
  year = 2022,
  month = jul,
  number = {arXiv:2207.09453},
  eprint = {2207.09453},
  primaryclass = {cs},
  publisher = {arXiv},
  doi = {10.48550/arXiv.2207.09453},
  archiveprefix = {arXiv}
}

@inproceedings{satorrasEquivariantGraphNeural2021,
  title = {E(n) {{Equivariant Graph Neural Networks}}},
  booktitle = {Proceedings of the 38th {{International Conference}} on {{Machine Learning}}},
  author = {Satorras, V{\'\i}ctor Garcia and Hoogeboom, Emiel and Welling, Max},
  year = 2021,
  month = jul,
  pages = {9323--9332},
  publisher = {PMLR},
  issn = {2640-3498}
}

@inproceedings{sohl-dicksteinDeepUnsupervisedLearning2015a,
  title = {Deep {{Unsupervised Learning}} Using {{Nonequilibrium Thermodynamics}}},
  booktitle = {Proceedings of the 32nd {{International Conference}} on {{Machine Learning}}},
  author = {{Sohl-Dickstein}, Jascha and Weiss, Eric and Maheswaranathan, Niru and Ganguli, Surya},
  year = 2015,
  month = jun,
  pages = {2256--2265},
  publisher = {PMLR},
  issn = {1938-7228}
}

@inproceedings{songScoreBasedGenerativeModeling2020,
  title = {Score-{{Based Generative Modeling}} through {{Stochastic Differential Equations}}},
  booktitle = {International {{Conference}} on {{Learning Representations}}},
  author = {Song, Yang and {Sohl-Dickstein}, Jascha and Kingma, Diederik P. and Kumar, Abhishek and Ermon, Stefano and Poole, Ben},
  year = 2020,
  month = oct
}

@inproceedings{hoDenoisingDiffusionProbabilistic2020,
  title = {Denoising {{Diffusion Probabilistic Models}}},
  booktitle = {Neural {{Information Processing Systems}} 2020},
  author = {Ho, Jonathan and Jain, Ajay and Abbeel, Pieter},
  year = 2020,
  month = dec,
  eprint = {2006.11239},
  primaryclass = {cs, stat},
  publisher = {arXiv},
  archiveprefix = {arXiv}
}

@article{anderson82,
  title = {Reverse-Time Diffusion Equation Models},
  author = {Anderson, Brian D.O.},
  year = 1982,
  journal = {Stochastic Processes and their Applications},
  volume = {12},
  number = {3},
  pages = {313--326},
  issn = {0304-4149},
  doi = {10.1016/0304-4149(82)90051-5}
}

@inproceedings{albergoBuildingNormalizingFlows2023,
  title = {Building {{Normalizing Flows}} with {{Stochastic Interpolants}}},
  booktitle = {The {{Eleventh International Conference}} on {{Learning Representations}}},
  author = {Albergo, Michael Samuel and {Vanden-Eijnden}, Eric},
  year = 2023
}

@inproceedings{lipmanFlowMatchingGenerative2023,
  title = {Flow {{Matching}} for {{Generative Modeling}}},
  booktitle = {{{ICLR}}},
  author = {Lipman, Yaron and Chen, Ricky T. Q. and {Ben-Hamu}, Heli and Nickel, Maximilian and Le, Matthew},
  year = 2023,
  month = apr
}

@misc{albergoStochasticInterpolantsUnifying2023,
  title = {Stochastic {{Interpolants}}: {{A Unifying Framework}} for {{Flows}} and {{Diffusions}}},
  shorttitle = {Stochastic {{Interpolants}}},
  author = {Albergo, Michael S. and Boffi, Nicholas M. and {Vanden-Eijnden}, Eric},
  year = 2023,
  month = mar,
  number = {arXiv:2303.08797},
  eprint = {2303.08797},
  primaryclass = {cond-mat},
  publisher = {arXiv},
  archiveprefix = {arXiv}
}

@inproceedings{liuFlowStraightFast2023,
  title = {Flow {{Straight}} and {{Fast}}: {{Learning}} to {{Generate}} and {{Transfer Data}} with {{Rectified Flow}}},
  shorttitle = {Flow {{Straight}} and {{Fast}}},
  booktitle = {The {{Eleventh International Conference}} on {{Learning Representations}}},
  author = {Liu, Xingchao and Gong, Chengyue and Liu, Qiang},
  year = 2023
}

@article{mullerNeuralImportanceSampling2019a,
  title = {Neural {{Importance Sampling}}},
  author = {M{\"u}ller, Thomas and Mcwilliams, Brian and Rousselle, Fabrice and Gross, Markus and Nov{\'a}k, Jan},
  year = 2019,
  month = oct,
  journal = {ACM Transactions on Graphics},
  volume = {38},
  number = {5},
  pages = {1--19},
  issn = {0730-0301, 1557-7368},
  doi = {10.1145/3341156}
}

@article{noeBoltzmannGeneratorsSampling2019,
  title = {Boltzmann Generators: {{Sampling}} Equilibrium States of Many-Body Systems with Deep Learning},
  shorttitle = {Boltzmann Generators},
  author = {No{\'e}, Frank and Olsson, Simon and K{\"o}hler, Jonas and Wu, Hao},
  year = 2019,
  month = sep,
  journal = {Science},
  volume = {365},
  number = {6457},
  pages = {eaaw1147},
  issn = {0036-8075, 1095-9203},
  doi = {10.1126/science.aaw1147}
}

@article{albergoFlowbasedGenerativeModels2019,
  title = {Flow-Based Generative Models for {{Markov}} Chain {{Monte Carlo}} in Lattice Field Theory},
  author = {Albergo, M. S. and Kanwar, G. and Shanahan, P. E.},
  year = 2019,
  month = aug,
  journal = {Physical Review D},
  volume = {100},
  number = {3},
  pages = {034515},
  issn = {2470-0010, 2470-0029},
  doi = {10.1103/PhysRevD.100.034515}
}

@article{nicoliAsymptoticallyUnbiasedEstimation2020,
  title = {Asymptotically Unbiased Estimation of Physical Observables with Neural Samplers},
  author = {Nicoli, Kim A. and Nakajima, Shinichi and Strodthoff, Nils and Samek, Wojciech and {M {\"u}ller}, Klaus-Robert and Kessel, Pan},
  year = 2020,
  month = feb,
  journal = {Physical Review E},
  volume = {101},
  number = {2},
  eprint = {1910.13496},
  primaryclass = {cond-mat, stat},
  pages = {023304},
  issn = {2470-0045, 2470-0053},
  doi = {10.1103/PhysRevE.101.023304},
  archiveprefix = {arXiv}
}

@article{nicoliEstimationThermodynamicObservables2021,
  title = {Estimation of {{Thermodynamic Observables}} in {{Lattice Field Theories}} with {{Deep Generative Models}}},
  author = {Nicoli, Kim A. and Anders, Christopher J. and Funcke, Lena and Hartung, Tobias and Jansen, Karl and Kessel, Pan and Nakajima, Shinichi and Stornati, Paolo},
  year = 2021,
  month = jan,
  journal = {Physical Review Letters},
  volume = {126},
  number = {3},
  pages = {032001},
  issn = {0031-9007, 1079-7114},
  doi = {10.1103/PhysRevLett.126.032001}
}

@article{jarzynskiTargetedFreeEnergy2002,
  title = {Targeted Free Energy Perturbation},
  author = {Jarzynski, C.},
  year = 2002,
  month = apr,
  journal = {Physical Review E},
  volume = {65},
  number = {4},
  pages = {046122},
  publisher = {American Physical Society},
  doi = {10.1103/PhysRevE.65.046122}
}

@article{BENNETT1976245,
  title = {Efficient Estimation of Free Energy Differences from {{Monte Carlo}} Data},
  author = {Bennett, Charles H},
  year = 1976,
  journal = {Journal of Computational Physics},
  volume = {22},
  number = {2},
  pages = {245--268},
  issn = {0021-9991},
  doi = {10.1016/0021-9991(76)90078-4}
}

@article{HahnUsingBijectiveMaps2009,
  title = {Using bijective maps to improve free-energy estimates},
  author = {Hahn, A. M. and Then, H.},
  journal = {Phys. Rev. E},
  volume = {79},
  issue = {1},
  pages = {011113},
  numpages = {16},
  year = {2009},
  month = {Jan},
  publisher = {American Physical Society},
  doi = {10.1103/PhysRevE.79.011113},
  url = {https://link.aps.org/doi/10.1103/PhysRevE.79.011113}
}

@inproceedings{jiaNormalizingConstantEstimation2020,
  title = {Normalizing {{Constant Estimation}} with {{Gaussianized Bridge Sampling}}},
  booktitle = {Proceedings of {{The}} 2nd {{Symposium}} on  {{Advances}} in {{Approximate Bayesian Inference}}},
  author = {Jia, He and Seljak, Uros},
  year = 2020,
  month = feb,
  pages = {1--14},
  publisher = {PMLR},
  issn = {2640-3498}
}

@article{wirnsbergerTargetedFreeEnergy2020,
  title = {Targeted Free Energy Estimation via Learned Mappings},
  author = {Wirnsberger, Peter and Ballard, Andrew J. and Papamakarios, George and Abercrombie, Stuart and Racani{\`e}re, S{\'e}bastien and Pritzel, Alexander and Jimenez Rezende, Danilo and Blundell, Charles},
  year = 2020,
  month = oct,
  journal = {The Journal of Chemical Physics},
  volume = {153},
  number = {14},
  pages = {144112},
  issn = {0021-9606, 1089-7690},
  doi = {10.1063/5.0018903}
}

@article{wirnsbergerNormalizingFlowsAtomic2022,
  title = {Normalizing Flows for Atomic Solids},
  author = {Wirnsberger, Peter and Papamakarios, George and Ibarz, Borja and Racani{\`e}re, S{\'e}bastien and Ballard, Andrew J. and Pritzel, Alexander and Blundell, Charles},
  year = 2022,
  month = jun,
  journal = {Machine Learning: Science and Technology},
  volume = {3},
  number = {2},
  eprint = {2111.08696},
  primaryclass = {cond-mat, physics:physics, stat},
  pages = {025009},
  issn = {2632-2153},
  doi = {10.1088/2632-2153/ac6b16},
  archiveprefix = {arXiv}
}

@article{wirnsbergerEstimatingGibbsFree2023,
  title = {Estimating {{Gibbs}} Free Energies via Isobaric-Isothermal Flows},
  author = {Wirnsberger, Peter and Ibarz, Borja and Papamakarios, George},
  year = 2023,
  month = sep,
  journal = {Machine Learning: Science and Technology},
  volume = {4},
  number = {3},
  pages = {035039},
  publisher = {IOP Publishing},
  issn = {2632-2153},
  doi = {10.1088/2632-2153/acefa8}
}

@article{dingDeepBARFastExact2021,
  title = {{{DeepBAR}}: {{A Fast}} and {{Exact Method}} for {{Binding Free Energy Computation}}},
  shorttitle = {{{DeepBAR}}},
  author = {Ding, Xinqiang and Zhang, Bin},
  year = 2021,
  month = mar,
  journal = {The Journal of Physical Chemistry Letters},
  volume = {12},
  number = {10},
  pages = {2509--2515},
  issn = {1948-7185, 1948-7185},
  issn = {1948-7185, 1948-7185},
  doi = {10.1021/acs.jpclett.1c00189}
}

@article{schebekScalableBoltzmannGenerators2025,
  title = {Scalable {{Boltzmann}} Generators for Equilibrium Sampling of Large-Scale Materials},
  author = {Schebek, Maximilian and No{\'e}, Frank and Rogal, Jutta},
  year = 2026,
  month = jun,
  journal = {Nature Communications},
  volume = {17},
  number = {1},
  pages = {5010},
  publisher = {Nature Publishing Group},
  issn = {2041-1723},
  doi = {10.1038/s41467-026-73900-9},
  copyright = {2026 The Author(s)}
}

@misc{schebekAssessingGenerativeModeling2025,
  title = {Assessing Generative Modeling Approaches for Free Energy Estimates in Condensed Matter},
  author = {Schebek, Maximilian and He, Jiajun and Hoffmann, Emil and Du, Yuanqi and No{\'e}, Frank and Rogal, Jutta},
  year = 2025,
  month = dec,
  number = {arXiv:2512.23930},
  eprint = {2512.23930},
  primaryclass = {cond-mat},
  publisher = {arXiv},
  doi = {10.48550/arXiv.2512.23930},
  archiveprefix = {arXiv}
}

@article{olehnovicsAssessingAccuracyEfficiency2024,
  title = {Assessing the {{Accuracy}} and {{Efficiency}} of {{Free Energy Differences Obtained}} from {{Reweighted Flow-Based Probabilistic Generative Models}}},
  author = {Olehnovics, Edgar and Liu, Yifei Michelle and Mehio, Nada and Sheikh, Ahmad Y. and Shirts, Michael R. and Salvalaglio, Matteo},
  year = 2024,
  month = jul,
  journal = {Journal of Chemical Theory and Computation},
  volume = {20},
  number = {14},
  pages = {5913--5922},
  publisher = {American Chemical Society},
  issn = {1549-9618},
  doi = {10.1021/acs.jctc.4c00520}
}

@article{andrieuParticleMarkovChain2010a,
  title = {Particle {{Markov Chain Monte Carlo Methods}}},
  author = {Andrieu, Christophe and Doucet, Arnaud and Holenstein, Roman},
  year = 2010,
  month = jun,
  journal = {Journal of the Royal Statistical Society Series B: Statistical Methodology},
  volume = {72},
  number = {3},
  pages = {269--342},
  issn = {1369-7412},
  doi = {10.1111/j.1467-9868.2009.00736.x}
}

@article{delbonoPerformanceMachinelearningassistedMonte2025,
  title = {Performance of Machine-Learning-Assisted {{Monte Carlo}} in Sampling from Simple Statistical Physics Models},
  author = {Del Bono, Luca Maria and {Ricci-Tersenghi}, Federico and Zamponi, Francesco},
  year = 2025,
  month = oct,
  journal = {Physical Review E},
  volume = {112},
  number = {4},
  pages = {045307},
  publisher = {American Physical Society},
  doi = {10.1103/s1rm-29zx}
}

@misc{hoffmanNeuTralizingBadGeometry2019,
  title = {{{NeuTra-lizing Bad Geometry}} in {{Hamiltonian Monte Carlo Using Neural Transport}}},
  author = {Hoffman, Matthew and Sountsov, Pavel and Dillon, Joshua V. and Langmore, Ian and Tran, Dustin and Vasudevan, Srinivas},
  year = 2019,
  month = mar,
  number = {arXiv:1903.03704},
  eprint = {1903.03704},
  primaryclass = {stat},
  publisher = {arXiv},
  archiveprefix = {arXiv}
}

@inproceedings{cabezasTransportEllipticalSlice2023,
  title = {Transport {{Elliptical Slice Sampling}}},
  booktitle = {Proceedings of {{The}} 26th {{International Conference}} on {{Artificial Intelligence}} and {{Statistics}}},
  author = {Cabezas, Alberto and Nemeth, Christopher},
  year = 2023,
  month = apr,
  pages = {3664--3676},
  publisher = {PMLR},
  issn = {2640-3498}
}

@article{parnoTransportMapAccelerated2018,
  title = {Transport {{Map Accelerated Markov Chain Monte Carlo}}},
  author = {Parno, Matthew D. and Marzouk, Youssef M.},
  year = 2018,
  month = jan,
  journal = {SIAM/ASA Journal on Uncertainty Quantification},
  volume = {6},
  number = {2},
  pages = {645--682},
  issn = {2166-2525},
  doi = {10.1137/17M1134640}
}

@article{agapiouImportanceSamplingIntrinsic2017,
  title = {Importance {{Sampling}}: {{Intrinsic Dimension}} and {{Computational Cost}}},
  shorttitle = {Importance {{Sampling}}},
  author = {Agapiou, S. and Papaspiliopoulos, O. and {Sanz-Alonso}, D. and Stuart, A. M.},
  year = 2017,
  journal = {Statistical Science},
  volume = {32},
  number = {3},
  eprint = {26408299},
  eprinttype = {jstor},
  pages = {405--431},
  publisher = {Institute of Mathematical Statistics},
  issn = {0883-4237}
}

@article{deldebbioEfficientModellingTrivializing2021,
  title = {Efficient {{Modelling}} of {{Trivializing Maps}} for {{Lattice}} \$\textbackslash phi\textasciicircum 4\$ {{Theory Using Normalizing Flows}}: {{A First Look}} at {{Scalability}}},
  shorttitle = {Efficient {{Modelling}} of {{Trivializing Maps}} for {{Lattice}} \$\textbackslash phi\textasciicircum 4\$ {{Theory Using Normalizing Flows}}},
  author = {Del Debbio, Luigi and Rossney, Joe Marsh and Wilson, Michael},
  year = 2021,
  month = nov,
  journal = {Physical Review D},
  volume = {104},
  number = {9},
  eprint = {2105.12481},
  primaryclass = {hep-lat},
  pages = {094507},
  issn = {2470-0010, 2470-0029},
  doi = {10.1103/PhysRevD.104.094507},
  archiveprefix = {arXiv}
}

@article{nealAnnealedImportanceSampling2001,
  title = {Annealed Importance Sampling},
  author = {Neal, Radford M.},
  year = 2001,
  month = apr,
  journal = {Statistics and Computing},
  volume = {11},
  number = {2},
  pages = {125--139},
  issn = {1573-1375},
  doi = {10.1023/A:1008923215028}
}

@article{swendsenReplicaMonteCarlo1986,
  title = {Replica {{Monte Carlo Simulation}} of {{Spin-Glasses}}},
  author = {Swendsen, Robert H. and Wang, Jian-Sheng},
  year = 1986,
  month = nov,
  journal = {Physical Review Letters},
  volume = {57},
  number = {21},
  pages = {2607--2609},
  publisher = {American Physical Society},
  doi = {10.1103/PhysRevLett.57.2607}
}

@inproceedings{geyerMarkovChainMonte1991,
  title = {Markov {{Chain Monte Carlo Maximum Likelihood}}},
  booktitle = {Computing {{Science}} and {{Statistics}}: {{Proceedings}} of the 23rd {{Symposium}} on the {{Interface}}},
  author = {Geyer, Charles J},
  year = 1991,
  pages = {8},
  publisher = {Interface Foundation},
  address = {Fairfax}
}

@article{hukushimaExchangeMonteCarlo1996,
  title = {Exchange {{Monte Carlo Method}} and  {{Application}} to {{Spin Glass Simulations}}},
  author = {Hukushima, Koji and Nemoto, Koji},
  year = 1996,
  month = jun,
  journal = {Journal of the Physical Society of Japan},
  volume = {65},
  number = {6},
  pages = {1604--1608},
  publisher = {The Physical Society of Japan},
  issn = {0031-9015},
  doi = {10.1143/JPSJ.65.1604}
}

@article{delmoralSequentialMonteCarlo2006a,
  title = {Sequential {{Monte Carlo Samplers}}},
  author = {Del Moral, Pierre and Doucet, Arnaud and Jasra, Ajay},
  year = 2006,
  month = jun,
  journal = {Journal of the Royal Statistical Society Series B: Statistical Methodology},
  volume = {68},
  number = {3},
  pages = {411--436},
  issn = {1369-7412, 1467-9868},
  doi = {10.1111/j.1467-9868.2006.00553.x},
}

@article{hukushimaPopulationAnnealingIts2003,
  title = {Population {{Annealing}} and {{Its Application}} to a {{Spin Glass}}},
  author = {Hukushima, K. and Iba, Y.},
  year = 2003,
  month = nov,
  journal = {AIP Conference Proceedings},
  volume = {690},
  number = {1},
  pages = {200--206},
  issn = {0094-243X},
  doi = {10.1063/1.1632130}
}

@article{schebekEfficientMappingPhase2024a,
  title = {Efficient Mapping of Phase Diagrams with Conditional {{Boltzmann Generators}}},
  author = {Schebek, Maximilian and Invernizzi, Michele and No{\'e}, Frank and Rogal, Jutta},
  year = 2024,
  month = nov,
  journal = {Machine Learning: Science and Technology},
  volume = {5},
  number = {4},
  pages = {045045},
  publisher = {IOP Publishing},
  issn = {2632-2153},
  doi = {10.1088/2632-2153/ad849d}
}

@article{jungNormalizingFlowsEnhanced2024a,
  title = {Normalizing Flows as an Enhanced Sampling Method for Atomistic Supercooled Liquids},
  author = {Jung, Gerhard and Biroli, Giulio and Berthier, Ludovic},
  year = 2024,
  month = aug,
  journal = {Machine Learning: Science and Technology},
  volume = {5},
  number = {3},
  pages = {035053},
  publisher = {IOP Publishing},
  issn = {2632-2153},
  doi = {10.1088/2632-2153/ad6ca0}
}

@inproceedings{midgleyFlowAnnealedImportance2022b,
  title = {Flow {{Annealed Importance Sampling Bootstrap}}},
  booktitle = {The {{Eleventh International Conference}} on {{Learning Representations}}},
  author = {Midgley, Laurence Illing and Stimper, Vincent and Simm, Gregor N. C. and Sch{\"o}lkopf, Bernhard and {Hern{\'a}ndez-Lobato}, Jos{\'e} Miguel},
  year = 2022,
  month = sep
}

@inproceedings{tanScalableEquilibriumSampling2025,
  title = {Scalable {{Equilibrium Sampling}} with {{Sequential Boltzmann Generators}}},
  booktitle = {Proceedings of the 42nd {{International Conference}} on {{Machine Learning}}},
  author = {Tan, Charlie B. and Bose, Joey and Lin, Chen and Klein, Leon and Bronstein, Michael M. and Tong, Alexander},
  year = 2025,
  month = oct,
  pages = {58467--58498},
  publisher = {PMLR},
  issn = {2640-3498}
}

@inproceedings{wuStochasticNormalizingFlows2020a,
  title = {Stochastic {{Normalizing Flows}}},
  booktitle = {Advances in {{Neural Information Processing Systems}}},
  author = {Wu, Hao and K{\"o}hler, Jonas and Noe, Frank},
  year = 2020,
  volume = {33},
  pages = {5933--5944},
  publisher = {Curran Associates, Inc.}
}

@inproceedings{arbelAnnealedFlowTransport2021,
  title = {Annealed {{Flow Transport Monte Carlo}}},
  booktitle = {Proceedings of the 38th {{International Conference}} on {{Machine Learning}}},
  author = {Arbel, Michael and Matthews, Alex and Doucet, Arnaud},
  year = 2021,
  month = jul,
  pages = {318--330},
  publisher = {PMLR},
  issn = {2640-3498}
}

@inproceedings{matthewsContinualRepeatedAnnealed2022a,
  title = {Continual {{Repeated Annealed Flow Transport Monte Carlo}}},
  booktitle = {Proceedings of the 39th {{International Conference}} on {{Machine Learning}}},
  author = {Matthews, Alex and Arbel, Michael and Rezende, Danilo Jimenez and Doucet, Arnaud},
  year = 2022,
  month = jun,
  pages = {15196--15219},
  publisher = {PMLR},
  issn = {2640-3498}
}

@article{caselleStochasticNormalizingFlows2022,
  title = {Stochastic Normalizing Flows as Non-Equilibrium Transformations},
  author = {Caselle, Michele and Cellini, Elia and Nada, Alessandro and Panero, Marco},
  year = 2022,
  month = jul,
  journal = {Journal of High Energy Physics},
  volume = {2022},
  number = {7},
  pages = {15},
  issn = {1029-8479},
  doi = {10.1007/JHEP07(2022)015}
}

@article{jarzynskiNonequilibriumEqualityFree1997,
  title = {Nonequilibrium {{Equality}} for {{Free Energy Differences}}},
  author = {Jarzynski, C.},
  year = 1997,
  month = apr,
  journal = {Physical Review Letters},
  volume = {78},
  number = {14},
  pages = {2690--2693},
  issn = {0031-9007, 1079-7114},
  doi = {10.1103/PhysRevLett.78.2690}
}

@article{crooksNonequilibriumMeasurementsFree1998,
  title = {Nonequilibrium {{Measurements}} of {{Free Energy Differences}} for {{Microscopically Reversible Markovian Systems}}},
  author = {Crooks, Gavin E.},
  year = 1998,
  month = mar,
  journal = {Journal of Statistical Physics},
  volume = {90},
  number = {5},
  pages = {1481--1487},
  issn = {1572-9613},
  doi = {10.1023/A:1023208217925}
}

@article{nilmeierNonequilibriumCandidateMonte2011,
  title = {Nonequilibrium Candidate {{Monte Carlo}} Is an Efficient Tool for Equilibrium Simulation},
  author = {Nilmeier, Jerome P. and Crooks, Gavin E. and Minh, David D. L. and Chodera, John D.},
  year = 2011,
  month = nov,
  journal = {Proceedings of the National Academy of Sciences},
  volume = {108},
  number = {45},
  pages = {E1009-E1018},
  publisher = {Proceedings of the National Academy of Sciences},
  doi = {10.1073/pnas.1106094108}
}

@article{caselleSamplingLatticeNambuGoto2024,
  title = {Sampling the Lattice {{Nambu-Goto}} String Using {{Continuous Normalizing Flows}}},
  author = {Caselle, Michele and Cellini, Elia and Nada, Alessandro},
  year = 2024,
  month = feb,
  journal = {Journal of High Energy Physics},
  volume = {2024},
  number = {2},
  pages = {48},
  issn = {1029-8479},
  doi = {10.1007/JHEP02(2024)048}
}

@article{caselleNumericalDeterminationWidth2025,
  title = {Numerical Determination of the Width and Shape of the Effective String Using {{Stochastic Normalizing Flows}}},
  author = {Caselle, Michele and Cellini, Elia and Nada, Alessandro},
  year = 2025,
  month = feb,
  journal = {Journal of High Energy Physics},
  volume = {2025},
  number = {2},
  pages = {90},
  issn = {1029-8479},
  doi = {10.1007/JHEP02(2025)090}
}

@article{karamanisAcceleratingAstronomicalCosmological2022,
  title = {Accelerating Astronomical and Cosmological Inference with {{Preconditioned Monte Carlo}}},
  author = {Karamanis, Minas and Beutler, Florian and Peacock, John A. and Nabergoj, David and Seljak, Uros},
  year = 2022,
  month = sep,
  journal = {Monthly Notices of the Royal Astronomical Society},
  volume = {516},
  number = {2},
  eprint = {2207.05652},
  primaryclass = {astro-ph, physics:physics},
  pages = {1644--1653},
  issn = {0035-8711, 1365-2966},
  doi = {10.1093/mnras/stac2272},
  archiveprefix = {arXiv}
}

@misc{zhangAcceleratedParallelTempering2025,
  title = {Accelerated {{Parallel Tempering}} via {{Neural Transports}}},
  author = {Zhang, Leo and Potaptchik, Peter and He, Jiajun and Du, Yuanqi and Doucet, Arnaud and Vargas, Francisco and Dau, Hai-Dang and Syed, Saifuddin},
  year = 2025,
  month = sep,
  number = {arXiv:2502.10328},
  eprint = {2502.10328},
  primaryclass = {stat},
  publisher = {arXiv},
  doi = {10.48550/arXiv.2502.10328},
  archiveprefix = {arXiv}
}

@article{invernizziSkippingReplicaExchange2022,
  title = {Skipping the {{Replica Exchange Ladder}} with {{Normalizing Flows}}},
  author = {Invernizzi, Michele and Kr{\"a}mer, Andreas and Clementi, Cecilia and No{\'e}, Frank},
  year = 2022,
  month = dec,
  journal = {The Journal of Physical Chemistry Letters},
  volume = {13},
  number = {50},
  pages = {11643--11649},
  publisher = {American Chemical Society},
  doi = {10.1021/acs.jpclett.2c03327}
}

@article{kishSamplingOrganizationsGroups1965,
  title = {Sampling {{Organizations}} and {{Groups}} of {{Unequal Sizes}} On},
  author = {Kish, Leslie},
  year = 1965,
  journal = {American Sociological Review},
  volume = {30},
  number = {4},
  eprint = {2091346},
  eprinttype = {jstor},
  pages = {564--572},
  doi = {10.2307/2091346}
}

@article{pengFlowPerturbationAccelerate2025,
  title = {Flow Perturbation to Accelerate {{Boltzmann}} Sampling},
  author = {Peng, Xin and Gao, Ang},
  year = 2025,
  month = jul,
  journal = {Nature Communications},
  volume = {16},
  number = {1},
  pages = {6604},
  publisher = {Nature Publishing Group},
  issn = {2041-1723},
  doi = {10.1038/s41467-025-62039-8},
  copyright = {2025 The Author(s)}
}

@article{wuUnbiasedMonteCarlo2021,
  title = {Unbiased {{Monte Carlo Cluster Updates}} with {{Autoregressive Neural Networks}}},
  author = {Wu, Dian and Rossi, Riccardo and Carleo, Giuseppe},
  year = 2021,
  month = nov,
  journal = {Physical Review Research},
  volume = {3},
  number = {4},
  eprint = {2105.05650},
  primaryclass = {cond-mat, stat},
  pages = {L042024},
  issn = {2643-1564},
  doi = {10.1103/PhysRevResearch.3.L042024},
  archiveprefix = {arXiv}
}

@article{heninEnhancedSamplingMethods2022,
  title = {Enhanced {{Sampling Methods}} for {{Molecular Dynamics Simulations}} [{{Article}} v1.0]},
  author = {H{\'e}nin, J{\'e}r{\^o}me and Leli{\`e}vre, Tony and Shirts, Michael R. and Valsson, Omar and Delemotte, Lucie},
  year = 2022,
  month = dec,
  journal = {Living Journal of Computational Molecular Science},
  volume = {4},
  number = {1},
  pages = {1583--1583},
  issn = {2575-6524},
  doi = {10.33011/livecoms.4.1.1583},
  copyright = {Copyright (c) 2022 J\'er\^ome  H\'enin, Tony Leli\`evre, Michael Shirts, Omar Valsson, Lucie Delemotte}
}

@article{frenkelSpeedupMonteCarlo2004,
  title = {Speed-up of {{Monte Carlo}} Simulations by Sampling of Rejected States},
  author = {Frenkel, Daan},
  year = 2004,
  month = dec,
  journal = {Proceedings of the National Academy of Sciences},
  volume = {101},
  number = {51},
  pages = {17571--17575},
  publisher = {Proceedings of the National Academy of Sciences},
  doi = {10.1073/pnas.0407950101}
}

@article{chenGeneralizedMetropolisAcceptance2015,
  title = {Generalized {{Metropolis}} Acceptance Criterion for Hybrid Non-Equilibrium Molecular Dynamics---{{Monte Carlo}} Simulations},
  author = {Chen, Yunjie and Roux, Beno{\^i}t},
  year = 2015,
  month = jan,
  journal = {The Journal of Chemical Physics},
  volume = {142},
  number = {2},
  pages = {024101},
  issn = {0021-9606, 1089-7690},
  doi = {10.1063/1.4904889}
}

@article{nealSamplingMultimodalDistributions1996,
  title = {Sampling from Multimodal Distributions Using Tempered Transitions},
  author = {Neal, Radford M.},
  year = 1996,
  month = dec,
  journal = {Statistics and Computing},
  volume = {6},
  number = {4},
  pages = {353--366},
  issn = {1573-1375},
  doi = {10.1007/BF00143556}
}

@article{athenesComputationChemicalPotential2002,
  title = {Computation of a Chemical Potential Using a Residence Weight Algorithm},
  author = {Ath{\`e}nes, M.},
  year = 2002,
  month = oct,
  journal = {Physical Review E},
  volume = {66},
  number = {4},
  pages = {046705},
  publisher = {American Physical Society},
  doi = {10.1103/PhysRevE.66.046705}
}

@article{bleiVariationalInferenceReview2017,
  title = {Variational {{Inference}}: {{A Review}} for {{Statisticians}}},
  shorttitle = {Variational {{Inference}}},
  author = {Blei, David M. and Kucukelbir, Alp and McAuliffe, Jon D.},
  year = 2017,
  month = apr,
  journal = {Journal of the American Statistical Association},
  volume = {112},
  number = {518},
  eprint = {1601.00670},
  primaryclass = {cs, stat},
  pages = {859--877},
  issn = {0162-1459, 1537-274X},
  doi = {10.1080/01621459.2017.1285773},
  archiveprefix = {arXiv}
}

@article{jordanIntroductionVariationalMethods1999,
  title = {An {{Introduction}} to {{Variational Methods}} for {{Graphical Models}}},
  author = {Jordan, Michael I. and Ghahramani, Zoubin and Jaakkola, Tommi S. and Saul, Lawrence K.},
  year = 1999,
  month = nov,
  journal = {Machine Learning},
  volume = {37},
  number = {2},
  pages = {183--233},
  issn = {1573-0565},
  doi = {10.1023/A:1007665907178}
}

@inproceedings{rezendeVariationalInferenceNormalizing2015,
  title = {Variational {{Inference}} with {{Normalizing Flows}}},
  booktitle = {Proceedings of the 32nd {{International Conference}} on {{Machine Learning}}},
  author = {Rezende, Danilo and Mohamed, Shakir},
  year = 2015,
  month = jun,
  pages = {1530--1538},
  publisher = {PMLR},
  issn = {1938-7228}
}

@article{wuSolvingStatisticalMechanics2019,
  title = {Solving {{Statistical Mechanics Using Variational Autoregressive Networks}}},
  author = {Wu, Dian and Wang, Lei and Zhang, Pan},
  year = 2019,
  month = feb,
  journal = {Physical Review Letters},
  volume = {122},
  number = {8},
  eprint = {1809.10606},
  primaryclass = {cond-mat, stat},
  pages = {080602},
  issn = {0031-9007, 1079-7114},
  doi = {10.1103/PhysRevLett.122.080602},
  archiveprefix = {arXiv}
}

@techreport{minkaDivergenceMeasuresMessage2005,
  title = {Divergence Measures and Message Passing},
  author = {Minka, Tom},
  year = 2005,
  number = {MSR-TR-2005-173},
  institution = {Technical report, Microsoft Research}
}

@inproceedings{blessingELBOsLargeScaleEvaluation2024a,
  title = {Beyond {{ELBOs}}: {{A Large-Scale Evaluation}} of {{Variational Methods}} for {{Sampling}}},
  shorttitle = {Beyond {{ELBOs}}},
  booktitle = {Proceedings of the 41st {{International Conference}} on {{Machine Learning}}},
  author = {Blessing, Denis and Jia, Xiaogang and Esslinger, Johannes and Vargas, Francisco and Neumann, Gerhard},
  year = 2024,
  month = jul,
  pages = {4205--4229},
  publisher = {PMLR},
  issn = {2640-3498}
}

@article{nicoliDetectingMitigatingModecollapse2023,
  title = {Detecting and Mitigating Mode-Collapse for Flow-Based Sampling of Lattice Field Theories},
  author = {Nicoli, Kim A. and Anders, Christopher J. and Hartung, Tobias and Jansen, Karl and Kessel, Pan and Nakajima, Shinichi},
  year = 2023,
  month = dec,
  journal = {Physical Review D},
  volume = {108},
  number = {11},
  pages = {114501},
  publisher = {American Physical Society},
  doi = {10.1103/PhysRevD.108.114501}
}

@inproceedings{schopmansTemperatureAnnealedBoltzmannGenerators2025,
  title = {Temperature-{{Annealed Boltzmann Generators}}},
  booktitle = {Forty-Second {{International Conference}} on {{Machine Learning}}},
  author = {Schopmans, Henrik and Friederich, Pascal},
  year = 2025,
  month = jun
}

@inproceedings{huangImprovingExplorabilityVariational2018,
  title = {Improving {{Explorability}} in {{Variational Inference}} with {{Annealed Variational Objectives}}},
  booktitle = {Advances in {{Neural Information Processing Systems}}},
  author = {Huang, Chin-Wei and Tan, Shawn and Lacoste, Alexandre and Courville, Aaron C},
  year = 2018,
  volume = {31},
  publisher = {Curran Associates, Inc.}
}

@misc{wangMitigatingModeCollapse2025,
  title = {Mitigating Mode Collapse in Normalizing Flows by Annealing with an Adaptive Schedule: {{Application}} to Parameter Estimation},
  shorttitle = {Mitigating Mode Collapse in Normalizing Flows by Annealing with an Adaptive Schedule},
  author = {Wang, Yihang and Chi, Chris and Dinner, Aaron R.},
  year = 2025,
  month = may,
  number = {arXiv:2505.03652},
  eprint = {2505.03652},
  primaryclass = {cs},
  publisher = {arXiv},
  doi = {10.48550/arXiv.2505.03652},
  archiveprefix = {arXiv}
}

@misc{leminhNaturalVariationalAnnealing2025,
  title = {Natural {{Variational Annealing}} for {{Multimodal Optimization}}},
  author = {LeMinh, T{\^a}m and Arbel, Julyan and M{\"o}llenhoff, Thomas and Khan, Mohammad Emtiyaz and Forbes, Florence},
  year = 2025,
  month = dec,
  number = {arXiv:2501.04667},
  eprint = {2501.04667},
  primaryclass = {stat},
  publisher = {arXiv},
  doi = {10.48550/arXiv.2501.04667},
  archiveprefix = {arXiv}
}

@article{hibat-allahVariationalNeuralAnnealing2021a,
  title = {Variational Neural Annealing},
  author = {{Hibat-Allah}, Mohamed and Inack, Estelle M. and Wiersema, Roeland and Melko, Roger G. and Carrasquilla, Juan},
  year = 2021,
  month = nov,
  journal = {Nature Machine Intelligence},
  volume = {3},
  number = {11},
  pages = {952--961},
  publisher = {Nature Publishing Group},
  issn = {2522-5839},
  doi = {10.1038/s42256-021-00401-3},
  copyright = {2021 The Author(s), under exclusive licence to Springer Nature Limited}
}

@article{khandokerLatticeProteinFolding2025,
  title = {Lattice Protein Folding with Variational Annealing},
  author = {Khandoker, Shoummo A and Inack, Estelle M and {Hibat-Allah}, Mohamed},
  year = 2025,
  month = aug,
  journal = {Machine Learning: Science and Technology},
  volume = {6},
  number = {3},
  pages = {035023},
  publisher = {IOP Publishing},
  issn = {2632-2153},
  doi = {10.1088/2632-2153/adf376}
}

@inproceedings{naessethMarkovianScoreClimbing2020,
  title = {Markovian {{Score Climbing}}: {{Variational Inference}} with {{KL}}(P\textbackslash vert \textbackslash vert q)},
  shorttitle = {Markovian {{Score Climbing}}},
  booktitle = {Advances in {{Neural Information Processing Systems}}},
  author = {Naesseth, Christian and Lindsten, Fredrik and Blei, David},
  year = 2020,
  volume = {33},
  pages = {15499--15510},
  publisher = {Curran Associates, Inc.}
}

@inproceedings{brofosAdaptationIndependentMetropolisHastings2022,
  title = {Adaptation of the {{Independent Metropolis-Hastings Sampler}} with {{Normalizing Flow Proposals}}},
  booktitle = {Proceedings of {{The}} 25th {{International Conference}} on {{Artificial Intelligence}} and {{Statistics}}},
  author = {Brofos, James and Gabrie, Marylou and Brubaker, Marcus A. and Lederman, Roy R.},
  year = 2022,
  month = may,
  pages = {5949--5986},
  publisher = {PMLR},
  issn = {2640-3498}
}

@misc{hackettFlowbasedSamplingMultimodal2021,
  title = {Flow-Based Sampling for Multimodal Distributions in Lattice Field Theory},
  author = {Hackett, Daniel C. and Hsieh, Chung-Chun and Albergo, Michael S. and Boyda, Denis and Chen, Jiunn-Wei and Chen, Kai-Feng and Cranmer, Kyle and Kanwar, Gurtej and Shanahan, Phiala E.},
  year = 2021,
  month = jul,
  number = {arXiv:2107.00734},
  eprint = {2107.00734},
  primaryclass = {cond-mat, physics:hep-lat},
  publisher = {arXiv},
  archiveprefix = {arXiv}
}

@article{mcnaughtonBoostingMonteCarlo2020,
  title = {Boosting {{Monte Carlo}} Simulations of Spin Glasses Using Autoregressive Neural Networks},
  author = {McNaughton, B. and Milo{\v s}evi{\'c}, M. V. and Perali, A. and Pilati, S.},
  year = 2020,
  month = may,
  journal = {Physical Review E},
  volume = {101},
  number = {5},
  eprint = {2002.04292},
  primaryclass = {cond-mat, physics:physics},
  pages = {053312},
  issn = {2470-0045, 2470-0053},
  doi = {10.1103/PhysRevE.101.053312},
  archiveprefix = {arXiv}
}

@article{ciarellaMachinelearningassistedMonteCarlo2023,
  title = {Machine-Learning-Assisted {{Monte Carlo}} Fails at Sampling Computationally Hard Problems},
  author = {Ciarella, Simone and Trinquier, Jeanne and Weigt, Martin and Zamponi, Francesco},
  year = 2023,
  month = mar,
  journal = {Machine Learning: Science and Technology},
  volume = {4},
  number = {1},
  pages = {010501},
  issn = {2632-2153},
  doi = {10.1088/2632-2153/acbe91}
}

@misc{bonoDemonstratingRealAdvantage2025,
  title = {Demonstrating {{Real Advantage}} of {{Machine-Learning-Enhanced Monte Carlo}} for {{Combinatorial Optimization}}},
  author = {Bono, Luca Maria Del and {Ricci-Tersenghi}, Federico and Zamponi, Francesco},
  year = 2025,
  month = oct,
  number = {arXiv:2510.19544},
  eprint = {2510.19544},
  primaryclass = {cond-mat},
  publisher = {arXiv},
  doi = {10.48550/arXiv.2510.19544},
  archiveprefix = {arXiv}
}

@inproceedings{skretaSuperpositionDiffusionModels2024,
  title = {The {{Superposition}} of {{Diffusion Models Using}} the {{It\^o Density Estimator}}},
  booktitle = {The {{Thirteenth International Conference}} on {{Learning Representations}}},
  author = {Skreta, Marta and Atanackovic, Lazar and Bose, Joey and Tong, Alexander and Neklyudov, Kirill},
  year = 2024,
  month = oct
}

@inproceedings{choiDensityRatioEstimation2022,
  title = {Density {{Ratio Estimation}} via {{Infinitesimal Classification}}},
  booktitle = {Proceedings of {{The}} 25th {{International Conference}} on {{Artificial Intelligence}} and {{Statistics}}},
  author = {Choi, Kristy and Meng, Chenlin and Song, Yang and Ermon, Stefano},
  year = 2022,
  month = may,
  pages = {2552--2573},
  publisher = {PMLR},
  issn = {2640-3498}
}

@inproceedings{huangReverseDiffusionMonte2024,
  title = {Reverse {{Diffusion Monte Carlo}}},
  booktitle = {The {{Twelfth International Conference}} on {{Learning Representations}}},
  author = {Huang, Xunpeng and Dong, Hanze and Hao, Yifan and Ma, Yian and Zhang, Tong},
  year = 2024
}

@inproceedings{akhound-sadeghIteratedDenoisingEnergy2024a,
  title = {Iterated {{Denoising Energy Matching}} for {{Sampling}} from {{Boltzmann Densities}}},
  booktitle = {Forty-First {{International Conference}} on {{Machine Learning}}},
  author = {{Akhound-Sadegh}, Tara and {Rector-Brooks}, Jarrid and Bose, Joey and Mittal, Sarthak and Lemos, Pablo and Liu, Cheng-Hao and Sendera, Marcin and Ravanbakhsh, Siamak and Gidel, Gauthier and Bengio, Yoshua and Malkin, Nikolay and Tong, Alexander},
  year = 2024,
  month = jun
}

@inproceedings{salmonaCanPushforwardGenerative2022,
  title = {Can {{Push-forward Generative Models Fit Multimodal Distributions}}?},
  booktitle = {Advances in {{Neural Information Processing Systems}}},
  author = {Salmona, Antoine and De Bortoli, Valentin and Delon, Julie and Desolneux, Agnes},
  editor = {Koyejo, S. and Mohamed, S. and Agarwal, A. and Belgrave, D. and Cho, K. and Oh, A.},
  year = 2022,
  volume = {35},
  pages = {10766--10779},
  publisher = {Curran Associates, Inc.}
}

@article{cornishRelaxingBijectivityConstraints2020,
  title = {Relaxing {{Bijectivity Constraints}} with {{Continuously Indexed Normalising Flows}}},
  author = {Cornish, Rob and Caterini, Anthony and Deligiannidis, George and Doucet, Arnaud},
  year = 2020,
  journal = {Proceedings of the 37 th International Conference on Machine Learning},
  volume = {PMLR 119}
}

@inproceedings{tzenTheoreticalGuaranteesSampling2019,
  title = {Theoretical Guarantees for Sampling and Inference in Generative Models with Latent Diffusions},
  booktitle = {Proceedings of the {{Thirty-Second Conference}} on {{Learning Theory}}},
  author = {Tzen, Belinda and Raginsky, Maxim},
  year = 2019,
  month = jun,
  pages = {3084--3114},
  publisher = {PMLR},
  issn = {2640-3498}
}

@inproceedings{richterImprovedSamplingLearned2023a,
  title = {Improved Sampling via Learned Diffusions},
  booktitle = {The {{Twelfth International Conference}} on {{Learning Representations}}},
  author = {Richter, Lorenz and Berner, Julius},
  year = 2023,
  month = oct
}

@inproceedings{zhangPathIntegralSampler2021,
  title = {Path {{Integral Sampler}}: {{A Stochastic Control Approach For Sampling}}},
  shorttitle = {Path {{Integral Sampler}}},
  booktitle = {International {{Conference}} on {{Learning Representations}}},
  author = {Zhang, Qinsheng and Chen, Yongxin},
  year = 2021,
  month = oct
}

@inproceedings{vargasDenoisingDiffusionSamplers2023a,
  title = {Denoising {{Diffusion Samplers}}},
  booktitle = {The {{Eleventh International Conference}} on {{Learning Representations}}},
  author = {Vargas, Francisco and Grathwohl, Will Sussman and Doucet, Arnaud},
  year = 2023,
  month = sep
}

@inproceedings{havensAdjointSamplingHighly2025a,
  title = {Adjoint {{Sampling}}: {{Highly Scalable Diffusion Samplers}} via {{Adjoint Matching}}},
  shorttitle = {Adjoint {{Sampling}}},
  booktitle = {Forty-Second {{International Conference}} on {{Machine Learning}}},
  author = {Havens, Aaron J. and Miller, Benjamin Kurt and Yan, Bing and {Domingo-Enrich}, Carles and Sriram, Anuroop and Levine, Daniel S. and Wood, Brandon M. and Hu, Bin and Amos, Brandon and Karrer, Brian and Fu, Xiang and Liu, Guan-Horng and Chen, Ricky T. Q.},
  year = 2025,
  month = jun
}

@article{bernerOptimalControlPerspective2023,
  title = {An Optimal Control Perspective on Diffusion-Based Generative Modeling},
  author = {Berner, Julius and Richter, Lorenz and Ullrich, Karen},
  year = 2023,
  month = oct,
  journal = {Transactions on Machine Learning Research},
  issn = {2835-8856}
}

@article{zhangEfficientUnbiasedSampling2025,
  title = {Efficient and {{Unbiased Sampling}} from {{Boltzmann Distributions}} via {{Variance-Tuned Diffusion Models}}},
  author = {Zhang, Fengzhe and Midgley, Laurence Illing and {Hern{\'a}ndez-Lobato}, Jos{\'e} Miguel},
  year = 2025,
  month = aug,
  journal = {Transactions on Machine Learning Research},
  issn = {2835-8856}
}

@inproceedings{phillipsParticleDenoisingDiffusion2024a,
  title = {Particle {{Denoising Diffusion Sampler}}},
  booktitle = {Proceedings of the 41st {{International Conference}} on {{Machine Learning}}},
  author = {Phillips, Angus and Dau, Hai-Dang and Hutchinson, Michael John and Bortoli, Valentin De and Deligiannidis, George and Doucet, Arnaud},
  year = 2024,
  month = jul,
  pages = {40688--40724},
  publisher = {PMLR},
  issn = {2640-3498}
}

@inproceedings{zhangDiffusionGenerativeFlow2023a,
  title = {Diffusion {{Generative Flow Samplers}}: {{Improving}} Learning Signals through Partial Trajectory Optimization},
  shorttitle = {Diffusion {{Generative Flow Samplers}}},
  booktitle = {The {{Twelfth International Conference}} on {{Learning Representations}}},
  author = {Zhang, Dinghuai and Chen, Ricky T. Q. and Liu, Cheng-Hao and Courville, Aaron and Bengio, Yoshua},
  year = 2023,
  month = oct
}

@inproceedings{heNoTrickNo2025,
  title = {No {{Trick}}, {{No Treat}}: {{Pursuits}} and {{Challenges Towards Simulation-free Training}} of {{Neural Samplers}}},
  shorttitle = {No {{Trick}}, {{No Treat}}},
  booktitle = {Frontiers in {{Probabilistic Inference}}: {{Learning}} Meets {{Sampling}} - {{ICLR}} Workshop},
  author = {He, Jiajun and Du, Yuanqi and Vargas, Francisco and Zhang, Dinghuai and Padhy, Shreyas and OuYang, RuiKang and Gomes, Carla P. and {Hern{\'a}ndez-Lobato}, Jos{\'e} Miguel},
  year = 2025,
  month = apr
}

@article{mateLearningInterpolationsBoltzmann2023,
  title = {Learning {{Interpolations}} between {{Boltzmann Densities}}},
  author = {M{\'a}t{\'e}, B{\'a}lint and Fleuret, Fran{\c c}ois},
  year = 2023,
  month = jan,
  journal = {Transactions on Machine Learning Research},
  issn = {2835-8856}
}

@inproceedings{tianLiouvilleFlowImportance2024,
  title = {Liouville {{Flow Importance Sampler}}},
  booktitle = {Proceedings of the 41st {{International Conference}} on {{Machine Learning}}},
  author = {Tian, Yifeng and Panda, Nishant and Lin, Yen Ting},
  year = 2024,
  month = jul,
  pages = {48186--48210},
  publisher = {PMLR},
  issn = {2640-3498}
}

@inproceedings{fanPathGuidedParticlebasedSampling2024,
  title = {Path-{{Guided Particle-based Sampling}}},
  booktitle = {Proceedings of the 41st {{International Conference}} on {{Machine Learning}}},
  author = {Fan, Mingzhou and Zhou, Ruida and Tian, Chao and Qian, Xiaoning},
  year = 2024,
  month = jul,
  pages = {12916--12934},
  publisher = {PMLR},
  issn = {2640-3498}
}

@inproceedings{vargasTransportMeetsVariational2024,
  title = {Transport Meets {{Variational Inference}}: {{Controlled Monte Carlo Diffusions}}},
  shorttitle = {Transport Meets {{Variational Inference}}},
  booktitle = {The {{Twelfth International Conference}} on {{Learning Representations}}},
  author = {Vargas, Francisco and Padhy, Shreyas and Blessing, Denis and N{\"u}sken, Nikolas},
  year = 2024
}

@inproceedings{albergoNETSNonequilibriumTransport2025,
  title = {{{NETS}}: {{A Non-equilibrium Transport Sampler}}},
  shorttitle = {{{NETS}}},
  booktitle = {Forty-Second {{International Conference}} on {{Machine Learning}}},
  author = {Albergo, Michael Samuel and {Vanden-Eijnden}, Eric},
  year = 2025,
  month = jun
}

@article{vaikuntanathanEscortedFreeEnergy2008,
  title = {Escorted Free Energy Simulations: Improving Convergence by Reducing Dissipation},
  shorttitle = {Escorted Free Energy Simulations},
  author = {Vaikuntanathan, Suriyanarayanan and Jarzynski, Christopher},
  year = 2008,
  month = may,
  journal = {Physical Review Letters},
  volume = {100},
  number = {19},
  pages = {190601},
  issn = {0031-9007},
  doi = {10.1103/PhysRevLett.100.190601},
  pmid = {18518431}
}

@article{metropolisMonteCarloMethod1949,
  title = {The {{Monte Carlo Method}}},
  author = {Metropolis, Nicholas and Ulam, S.},
  year = 1949,
  month = sep,
  journal = {Journal of the American Statistical Association},
  volume = {44},
  number = {247},
  pages = {335--341},
  publisher = {Taylor \& Francis},
  issn = {0162-1459},
  doi = {10.1080/01621459.1949.10483310},
  pmid = {18139350}
}

@article{vonNeumannVariousTechniques1951,
  title = {Various {{Techniques Used}} in {{Connection With Random Digits}}},
  author = {{von Neumann}, John},
  year = 1951,
  journal = {Journal of Research of the National Bureau of Standards},
  volume = {3},
  pages = {36--38}
}

@article{metropolisEquationStateCalculations1953,
  title = {Equation of {{State Calculations}} by {{Fast Computing Machines}}},
  author = {Metropolis, Nicholas and Rosenbluth, Arianna W. and Rosenbluth, Marshall N. and Teller, Augusta H. and Teller, Edward},
  year = 1953,
  month = jun,
  journal = {The Journal of Chemical Physics},
  volume = {21},
  number = {6},
  pages = {1087--1092},
  issn = {0021-9606},
  doi = {10.1063/1.1699114}
}

@book{liuMonteCarloStrategies2004,
  title = {Monte {{Carlo Strategies}} in {{Scientific Computing}}},
  author = {Liu, Jun S.},
  year = 2004,
  series = {Springer {{Series}} in {{Statistics}}},
  publisher = {Springer},
  address = {New York, NY},
  doi = {10.1007/978-0-387-76371-2},
  isbn = {978-0-387-76369-9 978-0-387-76371-2}
}

@article{tjelmelandModeJumpingProposals2001b,
  title = {Mode {{Jumping Proposals}} in {{MCMC}}},
  author = {Tjelmeland, Hakon and Hegstad, Bjorn Kare},
  year = 2001,
  journal = {Scandinavian Journal of Statistics},
  volume = {28},
  number = {1},
  pages = {205--223},
  issn = {1467-9469},
  doi = {10.1111/1467-9469.00232},
  copyright = {Board of the Foundation of the Scandinavian Journal of Statistics 2001}
}

@article{pompeFrameworkAdaptiveMCMC2020,
  title = {A Framework for Adaptive {{MCMC}} Targeting Multimodal Distributions},
  author = {Pompe, Emilia and Holmes, Chris and {\L}atuszy{\'n}ski, Krzysztof},
  year = 2020,
  month = oct,
  journal = {The Annals of Statistics},
  volume = {48},
  number = {5},
  pages = {2930--2952},
  publisher = {Institute of Mathematical Statistics},
  issn = {0090-5364, 2168-8966},
  doi = {10.1214/19-AOS1916}
}

@book{chopinIntroductionSequentialMonte2020,
  title = {An Introduction to Sequential {{Monte Carlo}}},
  author = {Chopin, Nicolas and Papaspiliopoulos, Omiros},
  year = 2020,
  series = {Springer {{Series}} in {{Statistics}}},
  edition = {1st ed. 2020},
  publisher = {Springer International Publishing},
  address = {Cham},
  doi = {10.1007/978-3-030-47845-2},
  isbn = {978-3-030-47847-6 978-3-030-47845-2}
}

@article{jumperHighlyAccurateProtein2021,
  title = {Highly Accurate Protein Structure Prediction with {{AlphaFold}}},
  author = {Jumper, John and Evans, Richard and Pritzel, Alexander and Green, Tim and Figurnov, Michael and Ronneberger, Olaf and Tunyasuvunakool, Kathryn and Bates, Russ and {\v Z}{\'i}dek, Augustin and Potapenko, Anna and Bridgland, Alex and Meyer, Clemens and Kohl, Simon A. A. and Ballard, Andrew J. and Cowie, Andrew and {Romera-Paredes}, Bernardino and Nikolov, Stanislav and Jain, Rishub and Adler, Jonas and Back, Trevor and Petersen, Stig and Reiman, David and Clancy, Ellen and Zielinski, Michal and Steinegger, Martin and Pacholska, Michalina and Berghammer, Tamas and Bodenstein, Sebastian and Silver, David and Vinyals, Oriol and Senior, Andrew W. and Kavukcuoglu, Koray and Kohli, Pushmeet and Hassabis, Demis},
  year = 2021,
  month = aug,
  journal = {Nature},
  volume = {596},
  number = {7873},
  pages = {583--589},
  issn = {0028-0836, 1476-4687},
  doi = {10.1038/s41586-021-03819-2}
}

@inproceedings{corsoDiffDockDiffusionSteps2023,
  title = {{{DiffDock}}: {{Diffusion Steps}}, {{Twists}}, and {{Turns}} for {{Molecular Docking}}},
  shorttitle = {{{DiffDock}}},
  booktitle = {The {{Eleventh International Conference}} on {{Learning Representations}}},
  author = {Corso, Gabriele and St{\"a}rk, Hannes and Jing, Bowen and Barzilay, Regina and Jaakkola, Tommi S.},
  year = 2023
}

@article{liNeuralNetworkRenormalization2018,
  title = {Neural {{Network Renormalization Group}}},
  author = {Li, Shuo-Hui and Wang, Lei},
  year = 2018,
  month = dec,
  journal = {Physical Review Letters},
  volume = {121},
  number = {26},
  pages = {260601},
  issn = {0031-9007, 1079-7114},
  doi = {10.1103/PhysRevLett.121.260601}
}

@article{abbottApplicationsFlowModels2024,
  title = {Applications of Flow Models to the Generation of Correlated Lattice {{QCD}} Ensembles},
  author = {Abbott, Ryan and Botev, Aleksandar and Boyda, Denis and Hackett, Daniel C. and Kanwar, Gurtej and Racani{\`e}re, S{\'e}bastien and Rezende, Danilo J. and {Romero-L{\'o}pez}, Fernando and Shanahan, Phiala E. and Urban, Julian M.},
  year = 2024,
  month = may,
  journal = {Physical Review D},
  volume = {109},
  number = {9},
  pages = {094514},
  publisher = {American Physical Society},
  doi = {10.1103/PhysRevD.109.094514}
}

@article{cranmerAdvancesMachinelearningbasedSampling2023,
  title = {Advances in Machine-Learning-Based Sampling Motivated by Lattice Quantum Chromodynamics},
  author = {Cranmer, Kyle and Kanwar, Gurtej and Racani{\`e}re, S{\'e}bastien and Rezende, Danilo J. and Shanahan, Phiala E.},
  year = 2023,
  month = sep,
  journal = {Nature Reviews Physics},
  volume = {5},
  number = {9},
  pages = {526--535},
  publisher = {Nature Publishing Group},
  issn = {2522-5820},
  doi = {10.1038/s42254-023-00616-w},
  copyright = {2023 Springer Nature Limited}
}

@article{corettiBoltzmannGeneratorsNew2024,
  title = {Boltzmann {{Generators}} and the {{New Frontier}} of {{Computational Sampling}} in {{Many-Body Systems}}},
  author = {Coretti, Alessandro and Falkner, Sebastian and Weinreich, Jan and Dellago, Christoph and von Lilienfeld, O. Anatole},
  year = 2024,
  month = apr,
  journal = {KIM Review},
  volume = {2},
  number = {3},
  pages = {1--8},
  publisher = {KIM Review},
  issn = {2998-2421},
  doi = {10.25950/bfa99422}
}

@article{albergoLearningSampleBetter2024,
  title = {Learning to Sample Better},
  author = {Albergo, Michael Samuel and {Vanden-Eijnden}, Eric},
  year = 2024,
  month = oct,
  journal = {Journal of Statistical Mechanics: Theory and Experiment},
  volume = {2024},
  number = {10},
  pages = {104014},
  publisher = {IOP Publishing},
  issn = {1742-5468},
  doi = {10.1088/1742-5468/ad363c}
}

@article{corettiLearningMappingsEquilibrium2025,
  title = {Learning Mappings between Equilibrium States of Liquid Systems Using Normalizing Flows},
  author = {Coretti, Alessandro and Falkner, Sebastian and Geissler, Phillip L. and Dellago, Christoph},
  year = 2025,
  month = may,
  journal = {The Journal of Chemical Physics},
  volume = {162},
  number = {18},
  pages = {184102},
  issn = {0021-9606},
  doi = {10.1063/5.0253034}
}

@article{hutchinsonStochasticEstimatorTrace1989,
  title = {A {{Stochastic Estimator}} of the {{Trace}} of the {{Influence Matrix}} for {{Laplacian Smoothing Splines}}},
  author = {Hutchinson, M.F.},
  year = 1989,
  month = jan,
  journal = {Communications in Statistics - Simulation and Computation},
  volume = {18},
  number = {3},
  pages = {1059--1076},
  publisher = {Taylor \& Francis},
  issn = {0361-0918},
  doi = {10.1080/03610918908812806}
}

@inproceedings{grathwohlFFJORDFreeFormContinuous2018,
  title = {{{FFJORD}}: {{Free-Form Continuous Dynamics}} for {{Scalable Reversible Generative Models}}},
  shorttitle = {{{FFJORD}}},
  booktitle = {International {{Conference}} on {{Learning Representations}}},
  author = {Grathwohl, Will and Chen, Ricky T. Q. and Bettencourt, Jesse and Sutskever, Ilya and Duvenaud, David},
  year = 2018,
  month = sep
}

@article{karrasElucidatingDesignSpace2022,
  title = {Elucidating the {{Design Space}} of {{Diffusion-Based Generative Models}}},
  author = {Karras, Tero and Aittala, Miika and Aila, Timo and Laine, Samuli},
  year = 2022,
  month = dec,
  journal = {Advances in Neural Information Processing Systems},
  volume = {35},
  pages = {26565--26577}
}

@misc{montanariSamplingDiffusionsStochastic2023d,
  title = {Sampling, {{Diffusions}}, and {{Stochastic Localization}}},
  author = {Montanari, Andrea},
  year = 2023,
  month = may,
  number = {arXiv:2305.10690},
  eprint = {2305.10690},
  primaryclass = {cs},
  publisher = {arXiv},
  archiveprefix = {arXiv}
}

@inproceedings{debortoliDiffusionSchrodingerBridge2021,
  title = {Diffusion {{Schr\"odinger Bridge}} with {{Applications}} to {{Score-Based Generative Modeling}}},
  booktitle = {Advances in {{Neural Information Processing Systems}}},
  author = {De Bortoli, Valentin and Thornton, James and Heng, Jeremy and Doucet, Arnaud},
  year = 2021,
  volume = {34},
  pages = {17695--17709},
  publisher = {Curran Associates, Inc.}
}

@article{uriaNeuralAutoregressiveDistribution2016c,
  title = {Neural {{Autoregressive Distribution Estimation}}},
  author = {Uria, Benigno and C{\^o}t{\'e}, Marc-Alexandre and Gregor, Karol and Murray, Iain and Larochelle, Hugo},
  year = 2016,
  journal = {Journal of Machine Learning Research},
  volume = {17},
  number = {205},
  pages = {1--37},
  issn = {1533-7928}
}

@inproceedings{germainMADEMaskedAutoencoder2015,
  title = {{{MADE}}: {{Masked Autoencoder}} for {{Distribution Estimation}}},
  shorttitle = {{{MADE}}},
  booktitle = {Proceedings of the 32nd {{International Conference}} on {{Machine Learning}}},
  author = {Germain, Mathieu and Gregor, Karol and Murray, Iain and Larochelle, Hugo},
  year = 2015,
  month = jun,
  pages = {881--889},
  publisher = {PMLR},
  issn = {1938-7228}
}

@inproceedings{alaouiSamplingSherringtonKirkpatrickGibbs2022a,
  title = {Sampling from the {{Sherrington-Kirkpatrick Gibbs}} Measure via Algorithmic Stochastic Localization},
  booktitle = {2022 {{IEEE}} 63rd {{Annual Symposium}} on {{Foundations}} of {{Computer Science}} ({{FOCS}})},
  author = {Alaoui, Ahmed El and Montanari, Andrea and Sellke, Mark},
  year = 2022,
  month = oct,
  pages = {323--334},
  issn = {2575-8454},
  doi = {10.1109/FOCS54457.2022.00038}
}

@misc{montanariPosteriorSamplingHigh2024,
  title = {Posterior {{Sampling}} in {{High Dimension}} via {{Diffusion Processes}}},
  author = {Montanari, Andrea and Wu, Yuchen},
  year = 2024,
  month = aug,
  number = {arXiv:2304.11449},
  eprint = {2304.11449},
  primaryclass = {math},
  publisher = {arXiv},
  doi = {10.48550/arXiv.2304.11449},
  archiveprefix = {arXiv}
}

@article{ghioSamplingFlowsDiffusion2024,
  title = {Sampling with Flows, Diffusion, and Autoregressive Neural Networks from a Spin-Glass Perspective},
  author = {Ghio, Davide and Dandi, Yatin and Krzakala, Florent and Zdeborov{\'a}, Lenka},
  year = 2024,
  month = jul,
  journal = {Proceedings of the National Academy of Sciences},
  volume = {121},
  number = {27},
  pages = {e2311810121},
  publisher = {Proceedings of the National Academy of Sciences},
  doi = {10.1073/pnas.2311810121}
}

@article{olssonGenerativeMolecularDynamics2026,
  title = {Generative Molecular Dynamics},
  author = {Olsson, Simon},
  year = 2026,
  month = feb,
  journal = {Current Opinion in Structural Biology},
  volume = {96},
  pages = {103213},
  issn = {0959-440X},
  doi = {10.1016/j.sbi.2025.103213}
}

@article{asgharEfficientRareEvent2024,
  title = {Efficient Rare Event Sampling with Unsupervised Normalizing Flows},
  author = {Asghar, Solomon and Pei, Qing-Xiang and Volpe, Giorgio and Ni, Ran},
  year = 2024,
  month = nov,
  journal = {Nature Machine Intelligence},
  volume = {6},
  number = {11},
  pages = {1370--1381},
  publisher = {Nature Publishing Group},
  issn = {2522-5839},
  doi = {10.1038/s42256-024-00918-3},
  copyright = {2024 The Author(s)}
}

@article{falknerConditioningBoltzmannGenerators2023a,
  title = {Conditioning {{Boltzmann}} Generators for Rare Event Sampling},
  author = {Falkner, Sebastian and Coretti, Alessandro and Romano, Salvatore and Geissler, Phillip L. and Dellago, Christoph},
  year = 2023,
  month = sep,
  journal = {Machine Learning: Science and Technology},
  volume = {4},
  number = {3},
  pages = {035050},
  publisher = {IOP Publishing},
  issn = {2632-2153},
  doi = {10.1088/2632-2153/acf55c}
}

@inproceedings{kleinTimewarpTransferableAcceleration2023,
  title = {Timewarp: {{Transferable}} Acceleration of Molecular Dynamics by Learning Time-Coarsened Dynamics},
  booktitle = {Advances in Neural Information Processing Systems},
  author = {Klein, Leon and Foong, Andrew and Fjelde, Tor and Mlodozeniec, Bruno and Brockschmidt, Marc and Nowozin, Sebastian and Noe, Frank and Tomioka, Ryota},
  editor = {Oh, A. and Naumann, T. and Globerson, A. and Saenko, K. and Hardt, M. and Levine, S.},
  year = 2023,
  volume = {36},
  pages = {52863--52883},
  publisher = {Curran Associates, Inc.}
}

@article{schreinerImplicitTransferOperator2023a,
  title = {Implicit {{Transfer Operator Learning}}: {{Multiple Time-Resolution Models}} for {{Molecular Dynamics}}},
  shorttitle = {Implicit {{Transfer Operator Learning}}},
  author = {Schreiner, Mathias and Winther, Ole and Olsson, Simon},
  year = 2023,
  month = dec,
  journal = {Advances in Neural Information Processing Systems},
  volume = {36},
  pages = {36449--36462}
}

@misc{diezTransferableGenerativeModels2025,
  title = {Transferable {{Generative Models Bridge Femtosecond}} to {{Nanosecond Time-Step Molecular Dynamics}}},
  author = {Diez, Juan Viguera and Schreiner, Mathias and Olsson, Simon},
  year = 2025,
  month = oct,
  number = {arXiv:2510.07589},
  eprint = {2510.07589},
  primaryclass = {physics},
  publisher = {arXiv},
  doi = {10.48550/arXiv.2510.07589},
  archiveprefix = {arXiv}
}

@article{lewisScalableEmulationProtein2025,
  title = {Scalable Emulation of Protein Equilibrium Ensembles with Generative Deep Learning},
  author = {Lewis, Sarah and Hempel, Tim and {Jim{\'e}nez-Luna}, Jos{\'e} and Gastegger, Michael and Xie, Yu and Foong, Andrew Y. K. and Satorras, Victor Garc{\'i}a and Abdin, Osama and Veeling, Bastiaan S. and Zaporozhets, Iryna and Chen, Yaoyi and Yang, Soojung and Foster, Adam E. and Schneuing, Arne and Nigam, Jigyasa and Barbero, Federico and Stimper, Vincent and Campbell, Andrew and Yim, Jason and Lienen, Marten and Shi, Yu and Zheng, Shuxin and Schulz, Hannes and Munir, Usman and Sordillo, Roberto and Tomioka, Ryota and Clementi, Cecilia and No{\'e}, Frank},
  year = 2025,
  month = jul,
  journal = {Science},
  volume = {389},
  number = {6761},
  pages = {eadv9817},
  publisher = {American Association for the Advancement of Science},
  doi = {10.1126/science.adv9817}
}

@inproceedings{kleinTransferableBoltzmannGenerators2024a,
  title = {Transferable {{Boltzmann Generators}}},
  booktitle = {The {{Thirty-eighth Annual Conference}} on {{Neural Information Processing Systems}}},
  author = {Klein, Leon and Noe, Frank},
  year = 2024,
  month = nov
}

@article{wolffCollectiveMonteCarlo1989,
  title = {Collective {{Monte Carlo Updating}} for {{Spin Systems}}},
  author = {Wolff, Ulli},
  year = 1989,
  month = jan,
  journal = {Physical Review Letters},
  volume = {62},
  number = {4},
  pages = {361--364},
  publisher = {American Physical Society},
  doi = {10.1103/PhysRevLett.62.361}
}

@article{swendsenNonuniversalCriticalDynamics1987,
  title = {Nonuniversal Critical Dynamics in {{Monte Carlo}} Simulations},
  author = {Swendsen, Robert H. and Wang, Jian-Sheng},
  year = 1987,
  month = jan,
  journal = {Physical Review Letters},
  volume = {58},
  number = {2},
  pages = {86--88},
  publisher = {American Physical Society},
  doi = {10.1103/PhysRevLett.58.86}
}

@article{berthierEfficientSwapAlgorithms2019,
  title = {Efficient Swap Algorithms for Molecular Dynamics Simulations of Equilibrium Supercooled Liquids},
  author = {Berthier, Ludovic and Flenner, Elijah and Fullerton, Christopher J and Scalliet, Camille and Singh, Murari},
  year = 2019,
  month = jun,
  journal = {Journal of Statistical Mechanics: Theory and Experiment},
  volume = {2019},
  number = {6},
  pages = {064004},
  publisher = {{IOP Publishing and SISSA}},
  issn = {1742-5468},
  doi = {10.1088/1742-5468/ab1910}
}

@article{grigeraFastMonteCarlo2001,
  title = {Fast {{Monte Carlo}} Algorithm for Supercooled Soft Spheres},
  author = {Grigera, Tom{\'a}s S. and Parisi, Giorgio},
  year = 2001,
  month = mar,
  journal = {Physical Review E},
  volume = {63},
  number = {4},
  pages = {045102},
  publisher = {American Physical Society},
  doi = {10.1103/PhysRevE.63.045102}
}

@article{bernardEventchainMonteCarlo2009,
  title = {Event-Chain {{Monte Carlo}} Algorithms for Hard-Sphere Systems},
  author = {Bernard, Etienne P. and Krauth, Werner and Wilson, David B.},
  year = 2009,
  month = nov,
  journal = {Physical Review E},
  volume = {80},
  number = {5},
  pages = {056704},
  publisher = {American Physical Society},
  doi = {10.1103/PhysRevE.80.056704}
}

@article{shirtsEquilibriumFreeEnergies2003,
  title = {Equilibrium {{Free Energies}} from {{Nonequilibrium Measurements Using Maximum-Likelihood Methods}}},
  author = {Shirts, Michael R. and Bair, Eric and Hooker, Giles and Pande, Vijay S.},
  year = 2003,
  month = oct,
  journal = {Physical Review Letters},
  volume = {91},
  number = {14},
  pages = {140601},
  publisher = {American Physical Society},
  doi = {10.1103/PhysRevLett.91.140601}
}

@article{eldan2013thin,
	title        = {Thin shell implies spectral gap up to polylog via a stochastic localization scheme},
	author       = {Eldan, Ronen},
	year         = 2013,
	journal      = {Geometric and Functional Analysis},
	publisher    = {Springer},
	volume       = 23,
	number       = 2,
	pages        = {532--569}
}

@article{eldan2020taming,
	title        = {Taming correlations through entropy-efficient measure decompositions with applications to mean-field approximation},
	author       = {Eldan, Ronen},
	year         = 2020,
	journal      = {Probability Theory and Related Fields},
	publisher    = {Springer},
	volume       = 176,
	number       = {3-4},
	pages        = {737--755}
}

@article{eldan2022analysis,
	title        = {Analysis of high-dimensional distributions using pathwise methods},
	author       = {Eldan, Ronen},
	year         = 2022,
	journal      = {Proceedings of ICM, to appear}
}

@inproceedings{chen2022localization,
	title        = {Localization schemes: A framework for proving mixing bounds for {Markov} chains},
	author       = {Chen, Yuansi and Eldan, Ronen},
	year         = 2022,
	booktitle    = {2022 IEEE 63rd Annual Symposium on Foundations of Computer Science (FOCS)},
	pages        = {110--122},
	organization = {IEEE}
}

@article{rotskoffSamplingThermodynamicEnsembles2024,
  title = {Sampling Thermodynamic Ensembles of Molecular Systems with Generative Neural Networks: {{Will}} Integrating Physics-Based Models Close the Generalization Gap?},
  shorttitle = {Sampling Thermodynamic Ensembles of Molecular Systems with Generative Neural Networks},
  author = {Rotskoff, Grant M.},
  year = 2024,
  month = jun,
  journal = {Current Opinion in Solid State and Materials Science},
  volume = {30},
  pages = {101158},
  issn = {1359-0286},
  doi = {10.1016/j.cossms.2024.101158}
}

@article{raissiPhysicsinformedNeuralNetworks2019,
  title = {Physics-Informed Neural Networks: {{A}} Deep Learning Framework for Solving Forward and Inverse Problems Involving Nonlinear Partial Differential Equations},
  shorttitle = {Physics-Informed Neural Networks},
  author = {Raissi, M. and Perdikaris, P. and Karniadakis, G.E.},
  year = 2019,
  month = feb,
  journal = {Journal of Computational Physics},
  volume = {378},
  pages = {686--707},
  issn = {00219991},
  doi = {10.1016/j.jcp.2018.10.045}
}

@inproceedings{duFEATFreeEnergy2025,
  title = {{{FEAT}}: {{Free}} Energy {{Estimators}} with {{Adaptive Transport}}},
  shorttitle = {{{FEAT}}},
  booktitle = {The {{Thirty-ninth Annual Conference}} on {{Neural Information Processing Systems}}},
  author = {Du, Yuanqi and He, Jiajun and Vargas, Francisco and Wang, Yuanqing and Gomes, Carla P. and {Hern{\'a}ndez-Lobato}, Jos{\'e} Miguel and {Vanden-Eijnden}, Eric},
  year = 2025,
  month = oct
}

@inproceedings{cabezasMarkovianFlowMatching2024,
  title = {Markovian {{Flow Matching}}: {{Accelerating MCMC}} with {{Continuous Normalizing Flows}}},
  shorttitle = {Markovian {{Flow Matching}}},
  booktitle = {The {{Thirty-eighth Annual Conference}} on {{Neural Information Processing Systems}}},
  author = {Cabezas, Alberto and Sharrock, Louis and Nemeth, Christopher},
  year = 2024,
  month = nov
}

@misc{chenMarkovChainMonte2026,
  title = {Markov {{Chain Monte Carlo}} with {{Diffusion Paths}}},
  author = {Chen, Han and Liu, Sifan and Yang, Jun},
  year = 2026,
  month = jul,
  number = {arXiv:2607.11631},
  eprint = {2607.11631},
  primaryclass = {stat.CO},
  publisher = {arXiv},
  doi = {10.48550/arXiv.2607.11631},
  archiveprefix = {arXiv}
}

@article{mateNeuralThermodynamicIntegration2024a,
  title = {Neural {{Thermodynamic Integration}}: {{Free Energies}} from {{Energy-Based Diffusion Models}}},
  shorttitle = {Neural {{Thermodynamic Integration}}},
  author = {M{\'a}t{\'e}, B{\'a}lint and Fleuret, Fran{\c c}ois and Bereau, Tristan},
  year = 2024,
  month = nov,
  journal = {The Journal of Physical Chemistry Letters},
  volume = {15},
  number = {45},
  pages = {11395--11404},
  publisher = {American Chemical Society},
  doi = {10.1021/acs.jpclett.4c01958}
}

@article{mateSolvationFreeEnergies2025,
  title = {Solvation Free Energies from Neural Thermodynamic Integration},
  author = {M{\'a}t{\'e}, B{\'a}lint and Fleuret, Fran{\c c}ois and Bereau, Tristan},
  year = 2025,
  month = mar,
  journal = {The Journal of Chemical Physics},
  volume = {162},
  number = {12},
  pages = {124107},
  issn = {0021-9606},
  doi = {10.1063/5.0251736}
}

@article{kirkwoodStatisticalMechanicsFluid1935,
  title = {Statistical {{Mechanics}} of {{Fluid Mixtures}}},
  author = {Kirkwood, John G.},
  year = 1935,
  month = may,
  journal = {The Journal of Chemical Physics},
  volume = {3},
  number = {5},
  pages = {300--313},
  issn = {0021-9606},
  doi = {10.1063/1.1749657}
}

@article{greniouxDiffusionbasedAnnealedBoltzmann2026,
  title = {Diffusion-Based {{Annealed Boltzmann Generators}} : Benefits, Pitfalls and Hopes},
  shorttitle = {Diffusion-Based {{Annealed Boltzmann Generators}}},
  author = {Grenioux, Louis and Noble, Maxence},
  year = 2026,
  month = feb,
  journal = {Transactions on Machine Learning Research},
  issn = {2835-8856}
}

@article{wongFastGravitationalwaveParameter2023,
  title = {Fast {{Gravitational-wave Parameter Estimation}} without {{Compromises}}},
  author = {Wong, Kaze W. K. and Isi, Maximiliano and Edwards, Thomas D. P.},
  year = 2023,
  month = nov,
  journal = {The Astrophysical Journal},
  volume = {958},
  number = {2},
  pages = {129},
  publisher = {The American Astronomical Society},
  issn = {0004-637X},
  doi = {10.3847/1538-4357/acf5cd}
}

@article{woutersRobustParameterEstimation2024,
  title = {Robust Parameter Estimation within Minutes on Gravitational Wave Signals from Binary Neutron Star Inspirals},
  author = {Wouters, Thibeau and Pang, Peter T. H. and Dietrich, Tim and Van Den Broeck, Chris},
  year = 2024,
  month = oct,
  journal = {Physical Review D},
  volume = {110},
  number = {8},
  pages = {083033},
  publisher = {American Physical Society},
  doi = {10.1103/PhysRevD.110.083033}
}

@article{robertsExponentialConvergenceLangevin1996,
  title = {Exponential Convergence of {{Langevin}} Distributions and Their Discrete Approximations},
  author = {Roberts, Gareth O. and Tweedie, Richard L.},
  year = {1996},
  journal = {Bernoulli},
  volume = {2},
  number = {4},
  pages = {341--363},
  doi = {10.2307/3318418}
}

@article{rosskyBrownianDynamicsSmart1978,
  title = {Brownian Dynamics as Smart {{Monte Carlo}} Simulation},
  author = {Rossky, P. J. and Doll, J. D. and Friedman, H. L.},
  year = {1978},
  journal = {The Journal of Chemical Physics},
  volume = {69},
  number = {10},
  pages = {4628--4633},
  doi = {10.1063/1.436415}
}

@article{duaneHybridMonteCarlo1987,
  title = {Hybrid {{Monte Carlo}}},
  author = {Duane, Simon and Kennedy, A. D. and Pendleton, Brian J. and Roweth, Duncan},
  year = {1987},
  journal = {Physics Letters B},
  volume = {195},
  number = {2},
  pages = {216--222},
  doi = {10.1016/0370-2693(87)91197-X}
}

@incollection{nealMCMCUsingHamiltonian2011,
  title = {{{MCMC}} Using {{Hamiltonian}} Dynamics},
  author = {Neal, Radford M.},
  year = {2011},
  booktitle = {Handbook of {{Markov Chain Monte Carlo}}},
  editor = {Brooks, Steve and Gelman, Andrew and Jones, Galin L. and Meng, Xiao-Li},
  publisher = {Chapman and Hall/CRC},
  pages = {113--162},
  doi = {10.1201/b10905}
}

@article{hoffmanNoUTurnSampler2014,
  title = {The {{No-U-Turn}} Sampler: Adaptively Setting Path Lengths in {{Hamiltonian Monte Carlo}}},
  author = {Hoffman, Matthew D. and Gelman, Andrew},
  year = {2014},
  journal = {Journal of Machine Learning Research},
  volume = {15},
  number = {47},
  pages = {1593--1623}
}

@misc{foglianiAnnealingVariationalInference2026,
  title = {Annealing in Variational Inference Mitigates Mode Collapse: {{A}} Theoretical Study on {{Gaussian}} Mixtures},
  shorttitle = {Annealing in Variational Inference Mitigates Mode Collapse},
  author = {Fogliani, Luigi and Loureiro, Bruno and Gabrié, Marylou},
  year = 2026,
  month = feb,
  number = {arXiv:2602.12923},
  eprint = {2602.12923},
  primaryclass = {stat},
  publisher = {arXiv},
  doi = {10.48550/arXiv.2602.12923},
  archiveprefix = {arXiv}
}

@misc{greniouxRiemannianStochasticInterpolants2025,
  title = {Riemannian {{Stochastic Interpolants}} for {{Amorphous Particle Systems}}},
  author = {Grenioux, Louis and Galliano, Leonardo and Berthier, Ludovic and Biroli, Giulio and Gabrié, Marylou},
  year = 2025,
  month = dec,
  number = {arXiv:2512.16607},
  eprint = {2512.16607},
  primaryclass = {stat},
  publisher = {arXiv},
  doi = {10.48550/arXiv.2512.16607},
  archiveprefix = {arXiv}
}

@misc{schonleEfficientMonteCarloSampling2025,
  title = {Efficient {{Monte-Carlo}} Sampling of Metastable Systems Using Non-Local Collective Variable Updates},
  author = {Sch{\"o}nle, Christoph and Carbone, Davide and Gabrié, Marylou and Leli{\`e}vre, Tony and Stoltz, Gabriel},
  year = 2025,
  month = dec,
  number = {arXiv:2512.16812},
  eprint = {2512.16812},
  primaryclass = {cond-mat},
  publisher = {arXiv},
  doi = {10.48550/arXiv.2512.16812},
  archiveprefix = {arXiv},
  
}

@article{soletskyiTheoreticalPerspectiveMode2025,
	title = {A theoretical perspective on mode collapse in variational inference},
	volume = {6},
	issn = {2632-2153},
	url = {https://dx.doi.org/10.1088/2632-2153/adde2a},
	doi = {10.1088/2632-2153/adde2a},
	language = {en},
	number = {2},
	urldate = {2025-06-11},
	journal = {Machine Learning: Science and Technology},
	author = {Soletskyi, Roman and Gabrié, Marylou and Loureiro, Bruno},
	month = jun,
	year = {2025},
	note = {Publisher: IOP Publishing},
	pages = {025056}
}

@inproceedings{greniouxImprovingEvaluationSamplers2025b,
	title = {Improving the evaluation of samplers on multi-modal targets},
	url = {https://openreview.net/forum?id=d91E9RhVFU},
	language = {en},
	urldate = {2025-05-27},
	booktitle = {{ICLR} workshop on {Frontiers} in {Probabilistic} {Inference}: {Learning} meets {Sampling}},
	author = {Grenioux, Louis and Noble, Maxence and Gabrié, Marylou},
	month = apr,
	year = {2025}
}

@inproceedings{nobleLearnedReferencebasedDiffusion2024a,
	title = {Learned {Reference}-based {Diffusion} {Sampler} for multi-modal distributions},
	url = {https://openreview.net/forum?id=fmJUYgmMbL},
	language = {en},
    booktitle = {The Thirteenth International Conference on Learning Representations},
	author = {Noble, Maxence and Grenioux, Louis and Gabrié, Marylou and Durmus, Alain Oliviero},
	year = {2025}
}

@article{schonleSamplingMetastableSystems2025,
	title = {Sampling metastable systems using collective variables and {Jarzynski}–{Crooks} paths},
	volume = {527},
	issn = {0021-9991},
	url = {https://www.sciencedirect.com/science/article/pii/S0021999125000890},
	doi = {10.1016/j.jcp.2025.113806},
	urldate = {2025-02-21},
	journal = {Journal of Computational Physics},
	author = {Schönle, C. and Gabrié, Marylou and Lelièvre, T. and Stoltz, G.},
	month = apr,
	year = {2025},
	pages = {113806}
}

@article{tamagnoneCoarseGrainedMolecularDynamics2024,
	title = {Coarse-{Grained} {Molecular} {Dynamics} with {Normalizing} {Flows}},
	issn = {1549-9618},
	url = {https://pubs.acs.org/doi/abs/10.1021/acs.jctc.4c00700},
	doi = {10.1021/acs.jctc.4c00700},
	journal = {Journal of Chemical Theory and Computation},
	author = {Tamagnone, Samuel and Laio, Alessandro and Gabrié, Marylou},
	month = sep,
	year = {2024},
	note = {Publisher: American Chemical Society}
}

@inproceedings{greniouxStochasticLocalizationIterative2024,
	title = {Stochastic {Localization} via {Iterative} {Posterior} {Sampling}},
	url = {https://proceedings.mlr.press/v235/grenioux24a.html},
	language = {en},
	booktitle = {Proceedings of the 41st {International} {Conference} on {Machine} {Learning}},
	publisher = {PMLR},
	author = {Grenioux, Louis and Noble, Maxence and Gabrié, Marylou and Durmus, Alain Oliviero},
	month = jul,
	year = {2024},
	note = {ISSN: 2640-3498},
	pages = {16337--16376}
}

@article{molina-tabordaActiveLearningBoltzmann2024a,
	title = {Active {Learning} of {Boltzmann} {Samplers} and {Potential} {Energies} with {Quantum} {Mechanical} {Accuracy}},
	issn = {1549-9618},
	url = {https://doi.org/10.1021/acs.jctc.4c00506},
	doi = {10.1021/acs.jctc.4c00506},
	journal = {Journal of Chemical Theory and Computation},
	author = {Molina-Taborda, Ana and Cossio, Pilar and Lopez-Acevedo, Olga and Gabrié, Marylou},
	month = oct,
	year = {2024},
	note = {Publisher: American Chemical Society}
}

@inproceedings{greniouxSampling2023,
  title = {On {{Sampling}} with {{Approximate Transport Maps}}},
  booktitle = {Proceedings of the 40th {{International Conference}} on {{Machine Learning}}},
  author = {Grenioux, Louis and Durmus, Alain Oliviero and Moulines, Eric and Gabrié, Marylou},
  year = {2023},
  month = jul,
  pages = {11698--11733},
  publisher = {PMLR},
  issn = {2640-3498},
  urldate = {2024-03-29},
  langid = {english}
}

@inproceedings{schonleOptimizingMarkovChain2023,
	title = {Optimizing {Markov} {Chain} {Monte} {Carlo} {Convergence} with {Normalizing} {Flows} and {Gibbs} {Sampling}},
	url = {https://openreview.net/forum?id=iazHKyT4YK},
	language = {en},
	urldate = {2024-02-29},
	booktitle = {{NeurIPS} 2023 {AI} for {Science} {Workshop}},
	author = {Schönle, Christoph and Gabrié, Marylou},
	month = oct,
	year = {2023}
}

@inproceedings{samsonov2022localglobal,
title={Local-Global {MCMC} kernels: the best of both worlds},
author={Sergey Samsonov and Evgeny Lagutin and Gabrié, Marylou and Alain Durmus and Alexey Naumov and Eric Moulines},
booktitle={Advances in Neural Information Processing Systems},
editor={Alice H. Oh and Alekh Agarwal and Danielle Belgrave and Kyunghyun Cho},
year={2022},
url={https://openreview.net/forum?id=zb-xfApk4ZK}
}

@article{Gabrie2021,
archivePrefix = {arXiv},
arxivId = {2105.12603},
author = {Gabrié, Marylou and Rotskoff, Grant M. and Vanden-Eijnden, Eric},
doi = {10.1073/pnas.2109420119},
eprint = {2105.12603},
issn = {0027-8424},
journal = {Proceedings of the National Academy of Sciences},
month = {mar},
number = {10},
title = {{Adaptive Monte Carlo augmented with normalizing flows}},
url = {http://arxiv.org/abs/2105.12603 https://pnas.org/doi/full/10.1073/pnas.2109420119},
volume = {119},
year = {2022}
}

@book{dawidMachineLearningQuantum2025,
  title = {Machine {{Learning}} in {{Quantum Sciences}}},
  author = {Dawid, Anna and Arnold, Julian and Requena, Borja and Gresch, Alexander and P{\l}odzie{\'n}, Marcin and Donatella, Kaelan and Nicoli, Kim A. and Stornati, Paolo and Koch, Rouven and B{\"u}ttner, Miriam and Oku{\l}a, Robert and {Mu{\~n}oz-Gil}, Gorka and {Vargas-Hern{\'a}ndez}, Rodrigo A. and {Cervera-Lierta}, Alba and Carrasquilla, Juan and Dunjko, Vedran and Gabri{\'e}, Marylou and Huembeli, Patrick and {van Nieuwenburg}, Evert and Vicentini, Filippo and Wang, Lei and Wetzel, Sebastian J. and Carleo, Giuseppe and Greplov{\'a}, Eli{\v s}ka and Krems, Roman and Marquardt, Florian and Tomza, Micha{\l} and Lewenstein, Maciej and Dauphin, Alexandre},
  year = 2025,
  publisher = {Cambridge University Press},
  address = {Cambridge},
  doi = {10.1017/9781009504942},
  isbn = {978-1-009-50493-5}
}

@inproceedings{Gabrie2021a,
author = {Gabrié, Marylou and Rotskoff, Grant M. and Vanden-Eijnden, Eric},
booktitle = {Invertible Neural Networks, NormalizingFlows, and Explicit Likelihood Models (ICML Workshop).},
title = {{Efficient Bayesian Sampling Using Normalizing Flows to Assist Markov Chain Monte Carlo Methods}},
url = {https://arxiv.org/abs/2107.08001},
year = {2021}
}

\pagebreak

\appendix
\phantomsection                            % only if you use hyperref
\addcontentsline{toc}{section}{Appendix}  
\addtocontents{toc}{\protect\setcounter{tocdepth}{-1}}
\section{Reminders on local MCMC samplers}
\label{app:local-samplers}

We recall here, for reference, the definition three gradient-based widely used local samplers invoked notably in \cref{sec:03-exact_samplers}: the unadjusted Langevin algorithm (ULA), the Metropolis-adjusted Langevin algorithm (MALA) and Hamiltonian Monte Carlo (HMC).

Throughout this appendix the target distribution on $\X = \R^d$ is written in Boltzmann form,
\begin{align}
    \pi(x) = \frac{e^{-U(x)}}{\Z_\pi}, \qquad U(x) = - \log \pi(x) \; \text{(up to an additive constant)},
    \label{eq:app-mcmc-target}
\end{align}
with $U$ differentiable and $\nabla U$ available at a cost comparable to that of $U$ --- possibly by automatic differentiation. The local samplers recalled below require only $\nabla U$ and differences of $U$, never the intractable normalization constant $\Z_\pi$. They are said to be \emph{local} because each iteration displaces the state by an amount controlled by a small time-step $\delta t$: this is what makes them scale well with the dimension $d$, and also what makes them struggle against metastability as their local moves are unlikely to cross energy barriers.

\subsection{Overdamped Langevin dynamics: ULA and MALA}
\label{app:subsec:langevin}

The overdamped Langevin diffusion associated with the potential $U$ is the solution of the stochastic differential equation
\begin{align}
    \dd X_t = - \nabla U(X_t) \, \dd t + \sqrt{2} \, \dd W_t \, ,
    \label{eq:app-langevin-sde}
\end{align}
where $(W_t)_{t \geq 0}$ is a standard $d$-dimensional Brownian motion.

\paragraph{Unadjusted Langevin Algorithm (ULA)} Simulating \cref{eq:app-langevin-sde} requires a time discretization. The Euler-Maruyama scheme with time-step $\delta t$ defines the unadjusted Langevin algorithm,
\begin{align}
    x^{(k)} = x^{(k-1)} - \delta t \, \nabla U(x^{(k-1)}) + \sqrt{2 \delta t} \; \xi^{(k)}, \qquad \xi^{(k)} \sim \N(0, I_d) \; \text{i.i.d.},
    \label{eq:app-ula-update}
\end{align}
or equivalently the Markov kernel
\begin{align}
    M_{\delta t}(x'|x) = \N \left( x' ; \, x - \delta t \, \nabla U(x), \, 2 \delta t \, I_d \right).
    \label{eq:app-ula-kernel}
\end{align}
Each iteration costs a single gradient evaluation and no rejection ever occurs. The discretization however breaks the exact invariance of $\pi$: the chain \cref{eq:app-ula-update} admits an invariant distribution $\pi_{\delta t} \neq \pi$, at a distance from $\pi$ which vanishes with $\delta t$ (typically as $\mathcal{O}(\delta t)$ in total variation). ULA is therefore an \emph{asymptotically biased} sampler, and the choice of $\delta t$ trades this bias against the speed at which the chain decorrelates. It remains widely used when a controlled bias is acceptable, or as a building block in schemes where the bias is corrected elsewhere.

\paragraph{Metropolis-Adjusted Langevin Algorithm (MALA)} The bias is removed by treating \cref{eq:app-ula-kernel} as a proposal within a Metropolis-Hastings accept-reject step \cite{rosskyBrownianDynamicsSmart1978,robertsExponentialConvergenceLangevin1996}.

\begin{minipage}{\textwidth}
\captionof{algorithm}{\textbf{Metropolis-adjusted Langevin algorithm (MALA)}}
\label{alg:app-mala}
\vspace{-0.4em}
\begin{algorithmic}
    \State Initialize $x^{(0)}$, choose a time-step $\delta t$,
    \For {$k = 1 \ldots K$}
        \State Sample $\xi^{(k)} \sim \N(0, I_d)$ and propose $x' = x^{(k-1)} - \delta t \, \nabla U(x^{(k-1)}) + \sqrt{2\delta t} \, \xi^{(k)}$,
        \State Set $x^{(k)} = x'$ with probability $\alpha$, else $x^{(k)} = x^{(k-1)}$, with
        $\displaystyle \alpha = \min\left(1 \,, \; \frac{\pi(x')}{\pi(x^{(k-1)})} \times \frac{M_{\delta t}(x^{(k-1)}|x')}{M_{\delta t}(x'|x^{(k-1)})} \right) \, ,$
    \EndFor
    \Return samples $\{x^{(k)}\}_{k=1}^K$
\end{algorithmic}
\setlength{\parskip}{\savedparskip}
\vspace{0.5em}
\end{minipage}

MALA leaves $\pi$ exactly invariant for any value of $\delta t$, the price being a rejection rate which grows with $\delta t$. 

\paragraph{Preconditioning} When the target is ill-conditioned, i.e. when its high-probability region has widely different extensions along different directions, a single scalar step-size $\delta t$ is necessarily a compromise: it is limited by the most confined direction while the chain must diffuse across the most extended one. A standard remedy is to precondition the dynamics with a symmetric positive-definite matrix $A$, replacing \cref{eq:app-ula-update} by $x^{(k)} = x^{(k-1)} - \delta t \, A \nabla U(x^{(k-1)}) + \sqrt{2 \delta t} \, A^{1/2} \xi^{(k)}$, which amounts to running the sampler in the linearly reparametrized variable $A^{-1/2} x$. Seen from this angle, the neutra-MCMC strategy of \cref{subsubsec:03-neutra-MCMC} is a \emph{non-linear} generalization of preconditioning, in which the linear map $A^{1/2}$ is replaced by the learned transport map $T_\theta$.

\subsection{Hamiltonian Monte Carlo (HMC)}
\label{app:subsec:hmc}

HMC \cite{duaneHybridMonteCarlo1987,nealMCMCUsingHamiltonian2011} takes a different route to producing distant yet likely proposals: it augments the state space with a momentum variable $p \in \R^d$ and exploits the fact that Hamiltonian dynamics conserves energy, so that following it for a long time moves the state far away while keeping it on the same probability level set.

\paragraph{Augmented target and Hamiltonian dynamics} One introduces the Hamiltonian and the associated extended target
\begin{align}
    H(x,p) = U(x) + \tfrac{1}{2} p^\top \Sigma^{-1} p \, , \qquad
    \hat \pi(x,p) = \frac{e^{-H(x,p)}}{\Z_\pi \, \Z_p} = \pi(x) \, \N(p; 0, \Sigma) \, ,
    \label{eq:app-hmc-hamiltonian}
\end{align}
where $\Z_p = (2\pi)^{d/2} \sqrt{\det \Sigma}$ and where the covariance $\Sigma$ of the momenta --- the inverse of the mass matrix in the mechanical analogy --- plays the role of a preconditioner and is usually taken to be $I_d$ or a diagonal estimate of the target covariance. By construction the extended target $\hat \pi$ factorizes, so that its $x$-marginal is exactly $\pi$: sampling $\hat \pi$ and discarding the momenta samples $\pi$. The corresponding Hamiltonian dynamics reads
\begin{align}
    \dot x = \nabla_p H = \Sigma^{-1} p \, , \qquad \dot p = - \nabla_x H = - \nabla U(x) \, ,
    \label{eq:app-hmc-flow}
\end{align}
and its exact flow has two properties which make it an ideal proposal mechanism: it conserves $H$, hence the density $\hat \pi$, and it preserves the volume of phase space.

\paragraph{Leapfrog integration and Metropolis correction} As for Langevin, \cref{eq:app-hmc-flow} must be integrated numerically. The leapfrog (velocity-Verlet) scheme with time-step $\delta t$,
\begin{align}
    p^{t + 1/2} = p^{t} - \tfrac{\delta t}{2} \nabla U(x^{t}) \, , \quad
    x^{t+1} = x^{t} + \delta t \, \Sigma^{-1} p^{t+1/2} \, , \quad
    p^{t+1} = p^{t+1/2} - \tfrac{\delta t}{2} \nabla U(x^{t+1}) \, ,
    \label{eq:app-hmc-leapfrog}
\end{align}
is chosen precisely because it retains exactly the two structural properties above except energy conservation: it is volume-preserving --- so that no Jacobian factor enters the acceptance probability --- and reversible up to a momentum flip, i.e. integrating $L$ steps from $(x,p)$ to $(x_L, p_L)$ and then $L$ steps from $(x_L, -p_L)$ returns to $(x,-p)$. The energy is only conserved up to a $\mathcal{O}(\delta t^2)$ error, and it is this residual error alone which is corrected by a Metropolis accept-reject step:

\begin{minipage}{\textwidth}
\captionof{algorithm}{\textbf{Hamiltonian Monte Carlo (HMC)}}
\label{alg:app-hmc}
\vspace{-0.4em}
\begin{algorithmic}
    \State Initialize $x^{(0)}$, choose a time-step $\delta t$ and a number of leapfrog steps $L$,
    \For {$k = 1 \ldots K$}
        \State Draw a fresh momentum $p^0 \sim \N(0, \Sigma)$ and set $x^0 = x^{(k-1)}$,
        \State Integrate $L$ leapfrog steps \cref{eq:app-hmc-leapfrog} to obtain $(x^L, p^L)$,
        \State Compute the acceptance probability
        $\displaystyle \alpha = \min\left(1 \,, \; e^{-\left(H(x^L, p^L) - H(x^0, p^0)\right)} \right) \, ,$
        \State Set $x^{(k)} = x^L$ with probability $\alpha$, else $x^{(k)} = x^{(k-1)}$,
    \EndFor
    \Return samples $\{x^{(k)}\}_{k=1}^K$
\end{algorithmic}
\setlength{\parskip}{\savedparskip}
\vspace{0.5em}
\end{minipage}
The momentum is resampled from its Gaussian conditional at every iteration, which is a Gibbs update leaving $\hat \pi$ invariant; the momentum flip needed for reversibility is immaterial once the momenta are discarded and is therefore omitted from the algorithm. The case $L=1$ recovers a variant of MALA, and the interest of HMC lies in taking $L > 1$: the state then travels a distance of order $L \, \delta t$ in a persistent direction rather than diffusively. Too large a value of $L$ is however wasteful, the trajectory eventually turning back towards its starting point; adaptive schemes such as the no-U-turn sampler \cite{hoffmanNoUTurnSampler2014} select $L$ automatically and are the default in most probabilistic programming libraries. We emphasize that, despite these longer moves, HMC remains a local sampler in the sense discussed above: it typically explores efficiently metastable basins but struggles to jump reliably across energy barriers.

\subsection{Running these samplers in the latent space of a normalizing flow}
\label{app:subsec:latent-samplers}

In the neutra-MCMC scheme of \cref{subsubsec:03-neutra-MCMC}, the kernels above are not applied to $\pi$ but to the pullback distribution $\pi_{T_\theta}$ defined in \cref{eq:03-pullback}. All the formulas of this appendix then apply verbatim upon substituting for $U$ the pullback potential
\begin{align}
    U_{T_\theta}(z) = - \log \pi_{T_\theta}(z) = U(T_\theta(z)) - \log \left| \det \nabla_z T_\theta(z) \right| \, ,
    \label{eq:app-pullback-potential}
\end{align}
and for $x$ the latent variable $z$, the chain being finally mapped back to the state space through $x^{(k)} = T_\theta(z^{(k)})$. The gradient $\nabla_z U_{T_\theta}$ required by ULA, MALA and HMC involves differentiating through the map $T_\theta$ and through its log-Jacobian, which is the source of the increased cost per iteration mentioned in \cref{subsubsec:03-neutra-MCMC}. In exchange, if the flow is well trained, $\pi_{T_\theta}$ is close to the well-conditioned base distribution $\rzero = \N(0, I_d)$, and a much larger time-step $\delta t$ can be afforded than when sampling $\pi$ directly.

\section{Derivation of the BAR free energy estimator}
\label{app:bar-derivation}

We detail here the derivation of the implicit equation \cref{eq:03-bar-implicit equation} defining the BAR estimator $\hat F_\pi^{\text{BAR}}$ introduced in \cref{subsubsec:03-is}. We follow the maximum likelihood argument of \cite{shirtsEquilibriumFreeEnergies2003}, which recasts the variance-optimal estimator originally derived by Bennett \cite{BENNETT1976245} as the solution of a binary classification problem --- a formulation which makes both its optimality properties and the uniqueness of its solution transparent.

\paragraph{Step 0: setup and notations}
To emphasize the symmetry of the construction, we write both distributions in Boltzmann form,
\begin{align}
    \rtheta(x) = \frac{e^{-U_\theta(x)}}{\Z_\theta}, \quad U_\theta(x) = - \log \rtheta(x), \quad \Z_\theta = 1, \qquad
    \pi(x) = \frac{e^{-U(x)}}{\Z_\pi},
    \label{eq:app-bar-boltzmann}
\end{align}
so that the free energy of the generative model vanishes, $F_\theta = - \log \Z_\theta = 0$, and the free energy difference between the two distributions reduces to the free energy of the target, $\Delta F = F_\pi - F_\theta = F_\pi$. We assume that we are given $N$ i.i.d. samples $\{x_i\}_{i=1}^N \sim \rtheta$ --- readily generated by the flow --- and $M$ samples $\{x_j\}_{j=1}^M \sim \pi$. Note that the derivation below is in no way specific to the case where one of the two distributions is a generative model: it holds verbatim for any pair of distributions known up to their normalization constants, which is the setting in which BAR is used in the molecular simulation literature.

Consider now the following classification problem: a sample $x$ is drawn either from $\rtheta$ or from $\pi$, and we wish to decide which. Denoting by $c$ the corresponding class label, the pooled dataset defines the class-conditional densities and the class priors
\begin{align}
    p(x|c=\theta) = \rtheta(x), \quad p(x|c=\pi) = \pi(x) \quad \text{and} \quad \frac{p(c=\theta)}{p(c=\pi)} = \frac{N}{M}.
    \label{eq:app-bar-classes}
\end{align}

\paragraph{Step 1: the posterior class probability is a logistic function of the energy difference}
Bayes' rule, $p(c|x) = p(x|c) p(c) / p(x)$, applied to the two classes and divided term by term, gives
\begin{align}
    \frac{p(x|c=\theta)}{p(x|c=\pi)} = \frac{p(c=\theta|x)}{p(c=\pi|x)} \, \frac{p(c=\pi)}{p(c=\theta)} = \frac{p(c=\theta|x)}{1 - p(c=\theta|x)} \, \frac{M}{N},
    \label{eq:app-bar-bayes}
\end{align}
where the unknown evidence $p(x)$ has cancelled. Solving \cref{eq:app-bar-bayes} for the posterior probability yields
\begin{align}
    p(c=\theta|x) = \frac{1}{1 + \frac{M}{N} \frac{p(x|c=\pi)}{p(x|c=\theta)}}.
    \label{eq:app-bar-posterior-generic}
\end{align}
Hence the likelihood ratio appearing in \cref{eq:app-bar-posterior-generic} is the exponential of an energy difference offset by the sought free energy,
\begin{align}
    \frac{p(x|c=\pi)}{p(x|c=\theta)} = \frac{\pi(x)}{\rtheta(x)} = \frac{e^{-U(x)}}{\Z_\pi \, e^{-U_\theta(x)}} = e^{-(U(x) - U_\theta(x)) + F_\pi},
    \label{eq:app-bar-ratio}
\end{align}
so that
\begin{align}
    p(c=\theta|x) = \frac{1}{1 + \frac{M}{N} e^{-(U(x) - U_\theta(x)) + F_\pi}} = \frac{1}{1 + \frac{M}{N} e^{-U(x) - \log \rtheta(x) + F_\pi}}.
    \label{eq:app-bar-posterior}
\end{align}
The exact classifier is thus a logistic function of the energy difference $U - U_\theta$, in which the free energy $F_\pi$ enters as the sole unknown parameter --- an offset shifting the decision boundary. This suggests estimating $F_\pi$ by fitting this classifier to the data.

\paragraph{Step 2: maximum likelihood over the classifier}
Since $F_\pi$ is unknown, we treat it as a parameter and consider the one-parameter statistical model obtained by substituting a candidate value $\Delta$ for $F_\pi$ in \cref{eq:app-bar-posterior},
\begin{align}
    q(c=\theta|x; \Delta) = \frac{1}{1 + \frac{M}{N} e^{-U(x) - \log \rtheta(x) + \Delta}} \in [0,1], \qquad q(c=\pi|x; \Delta) = 1 - q(c=\theta|x; \Delta).
    \label{eq:app-bar-model}
\end{align}
Each of the $N+M$ available samples comes with a known class label, so that the conditional log-likelihood of the labels given the samples reads
\begin{align}
    \mathcal{L}(\Delta) &= \sum_{i=1}^N \log q(c=\theta|x_i; \Delta) + \sum_{j=1}^M \log q(c=\pi|x_j; \Delta) \notag \\
    &= - \sum_{i=1}^N \log \left( 1 + \frac{M}{N} e^{-U(x_i) - \log \rtheta(x_i) + \Delta} \right) - \sum_{j=1}^M \log \left( 1 + \frac{N}{M} e^{U(x_j) + \log \rtheta(x_j) - \Delta} \right).
    \label{eq:app-bar-loglik}
\end{align}
One recognizes in \cref{eq:app-bar-loglik} the cross-entropy loss of a logistic regression whose sole trainable parameter is the bias $\Delta$, the ``features'' $U(x) + \log \rtheta(x)$ being fixed and known.

Denoting $u_i = \frac{M}{N} e^{-U(x_i) - \log \rtheta(x_i) + \Delta}$ and $v_j = \frac{N}{M} e^{U(x_j) + \log \rtheta(x_j) - \Delta}$, the stationarity condition $\dd \mathcal{L} / \dd \Delta = 0$ reads
\begin{align}
    \sum_{i=1}^N \frac{u_i}{1 + u_i} = \sum_{j=1}^M \frac{v_j}{1 + v_j}
    \quad \Longleftrightarrow \quad
    \sum_{i=1}^N \frac{1}{1 + u_i^{-1}} = \sum_{j=1}^M \frac{1}{1 + v_j^{-1}},
    \label{eq:app-bar-stationarity}
\end{align}
which, upon substituting $u_i^{-1}$ and $v_j^{-1}$, is precisely the implicit equation \cref{eq:03-bar-implicit equation} defining $\hat F_\pi^{\text{BAR}}$. Its interpretation is transparent: at the optimum, the total probability the classifier assigns to the $N$ flow samples of having been produced by $\pi$ exactly matches the total probability it assigns to the $M$ target samples of having been produced by $\rtheta$; in other words, the two types of misclassification are balanced.

\paragraph{Step 3: existence and uniqueness of the solution}
Introducing the logistic function $\sigma(z) = (1 + e^{-z})^{-1}$, which satisfies $\sigma'(z) = \sigma(z)(1 - \sigma(z)) > 0$, the difference between the two sides of \cref{eq:app-bar-stationarity} can be written as
\begin{align}
    g(\Delta) = \sum_{i=1}^N \sigma\left( - U(x_i) - \log \rtheta(x_i) + \Delta - \log \tfrac{N}{M} \right) - \sum_{j=1}^M \sigma\left( U(x_j) + \log \rtheta(x_j) - \Delta - \log \tfrac{M}{N} \right),
    \label{eq:app-bar-g}
\end{align}
so that $\hat F_\pi^{\text{BAR}}$ is characterized by $g(\hat F_\pi^{\text{BAR}}) = 0$. The parameter $\Delta$ enters the arguments of the first sum with a $+$ sign and those of the second with a $-$ sign, so that
\begin{align}
    g'(\Delta) = \sum_{i=1}^N \sigma'\left( - U(x_i) - \log \rtheta(x_i) + \Delta - \log \tfrac{N}{M} \right) + \sum_{j=1}^M \sigma'\left( U(x_j) + \log \rtheta(x_j) - \Delta - \log \tfrac{M}{N} \right),
    \label{eq:app-bar-gprime}
\end{align}
hence $g'(\Delta) > 0$ for all $\Delta$, so that
and $g$ is strictly increasing: \cref{eq:03-bar-implicit equation} admits at most one solution. Moreover, since $\sigma(-\infty) = 0$ and $\sigma(+\infty) = 1$,
\begin{align}
    \lim_{\Delta \to -\infty} g(\Delta) = - M < 0 \quad \text{and} \quad \lim_{\Delta \to + \infty} g(\Delta) = N > 0 ,
    \label{eq:app-bar-limits}
\end{align}
so that, $g$ being smooth with limits of opposite signs, the intermediate value theorem guarantees the existence of a solution. Hence \cref{eq:03-bar-implicit equation} has a unique solution, which is the global maximum of the log-likelihood \cref{eq:app-bar-loglik}. In practice it is located by a one-dimensional root-finding routine.

\paragraph{Remarks}
Two comments are in order. First, because $\hat F_\pi^{\text{BAR}}$ has been obtained as a maximum likelihood estimator of a well-specified parametric model, it inherits the standard asymptotic guarantees: it is consistent, asymptotically unbiased and efficient as $N, M \to \infty$, i.e. it saturates the Cramér-Rao bound and therefore has minimal asymptotic variance among all estimators built from the same two sets of samples. This is the modern counterpart of Bennett's original derivation \cite{BENNETT1976245}, in which the estimator was obtained by minimizing the variance over a family of acceptance-ratio estimators. Second, the finite-sample variance is controlled by how well the classification problem can be solved: if $\rtheta$ and $\pi$ overlap poorly, the two classes are nearly separable, the log-likelihood \cref{eq:app-bar-loglik} becomes flat in $\Delta$ around its maximum and the estimator becomes ill-conditioned. This is the precise sense in which the quality of the generative model controls the quality of the free energy estimate. A derivation adapted to the case where the two distributions are related by a bijective map, as is the case for NFs, can be found in \cite{HahnUsingBijectiveMaps2009}.

\section{Derivation of the stochastic optimal control objective}
\label{app:soc-derivation}
 
We detail here the derivation of the training objective \cref{eq:04-kl-soc} presented in \cref{subsubsec:04-soc}. We work with a generic diffusion coefficient $\sigma > 0$, both to make explicit the $\sigma$-dependence of the prefactors and to facilitate the comparison with the conventions of \cite{zhangPathIntegralSampler2021,bernerOptimalControlPerspective2023,richterImprovedSamplingLearned2023a}. The main text corresponds to the choice $\sigma = \sqrt{2}$. Throughout, all path measures are defined on the space of continuous trajectories $\Omega = \mathcal{C}([0,T], \mathbb{R}^d)$, and we assume enough regularity on the drifts and on the densities involved for the successive Radon-Nikodym derivatives to be well defined --- in particular $\rho_\text{ref} \ll \pi$ and $D_\mathrm{KL}(\mathbb{P}_\theta \| \mathbb{P}_\pi) < \infty$.
 
\paragraph{Step 0: setup and notations}
The OU noising process of the target distribution $\pi$, and its exact time-reversal, now read
\begin{align}
    \dd \tilde X_\tau &= - \tilde X_\tau \, \dd \tau + \sigma \, \dd W_\tau, \quad \tilde X_0 \sim \pi \quad (\text{with time marginals} \; \pi_\tau) \label{eq:app-soc-forward}\\
    \dd X_t &= (X_t + \sigma^2 \nabla \log \pi_{T-t}(X_t)) \, \dd t + \sigma \, \dd W_t, \quad X_0 \sim \pi_T. \label{eq:app-soc-backward}
\end{align}
The invariant distribution of the noising process \cref{eq:app-soc-forward} is the Gaussian $\gamma_\sigma = \N(0, \tfrac{\sigma^2}{2} I)$, which reduces to $\N(0,I)$ for $\sigma = \sqrt{2}$ and which we use as the initial distribution of the generative processes below.
 
Since the derivation involves both time directions, we distinguish them notationally. We denote by $\tilde{\mathbb{P}}_\pi$ the path measure of the noising process \cref{eq:app-soc-forward} and by ${\mathbb{P}}_\pi$ the path measure of the denoising process \cref{eq:app-soc-backward}. The time-reversal involution on path space is denoted
\begin{align}
    R : \Omega \to \Omega, \qquad (R x)_t = x_{T-t}.
\end{align}
The statement that \cref{eq:app-soc-backward} is the exact time-reversal of \cref{eq:app-soc-forward} --- a classical result of \cite{anderson82} --- precisely means that its path measure is the pushforward
\begin{align}
    \mathbb{P}_\pi = R_{\#} \tilde{\mathbb{P}}_\pi.
    \label{eq:app-soc-pushforward}
\end{align}
Because $R$ is a measurable bijection, pushing forward leaves Radon-Nikodym derivatives and divergences unchanged, in the sense that for any two path measures $\mu, \nu$ on $\Omega$,
\begin{align}
    \frac{\dd R_\# \mu}{\dd R_\# \nu}(x) = \frac{\dd \mu}{\dd \nu}(R x), \qquad D_\mathrm{KL}(R_\# \mu \| R_\# \nu) = D_\mathrm{KL}(\mu \| \nu).
    \label{eq:app-soc-invariance}
\end{align}
The discrepancy between two path measures may therefore be evaluated in whichever time direction is the more convenient --- an observation which is the crux of Step 3.
 
We introduce the same objects for the reference distribution $\rho_\text{ref}$: its noised marginals are $\rho_{\text{ref},\tau}$, its noising path measure is $\tilde{\mathbb{P}}_{\text{ref}}$, and $\mathbb{P}_{\text{ref}} = R_\# \tilde{\mathbb{P}}_{\text{ref}}$. The variational path measure $\mathbb{P}_\theta$, in contrast, is defined directly in generative time and has no noising counterpart; it is generated by
\begin{align}
    \dd X_t = (X_t + \sigma^2 s_\theta(X_t, T-t)) \, \dd t + \sigma \, \dd W_t, \quad X_0 \sim \gamma_\sigma.
\end{align}
It is convenient to gather the drifts of the three generative processes,
\begin{align}
    b_\theta(x,t) &= x + \sigma^2 s_\theta(x,T-t), \\
    b_\text{ref}(x,t) &= x + \sigma^2 \nabla \log \rho_{\text{ref},T-t}(x), \\
    b_\pi(x,t) &= x + \sigma^2 \nabla \log \pi_{T-t}(x),
\end{align}
and to note that the reparametrization $s_\theta(x,t) = \nabla \log \rho_{\text{ref},T-t}(x) + u_\theta(x,t)$ of the score in terms of the residual control $u_\theta$ amounts to
\begin{align}
    b_\theta(x,t) - b_\text{ref}(x,t) = \sigma^2 \, u_\theta(x,T-t).
    \label{eq:app-soc-drift-gap}
\end{align}
The factor $\sigma^2$ in \cref{eq:app-soc-drift-gap} originates from the score coefficient of the reverse SDE, and is responsible for the prefactor obtained in Step 2 below.
 
\paragraph{Step 1: decomposition of the KL divergence}
All three generative path measures $\mathbb{P}_\theta$, $\mathbb{P}_{\text{ref}}$ and $\mathbb{P}_\pi$ being mutually absolutely continuous, the chain rule for Radon-Nikodym derivatives gives, trajectory-wise,
\begin{align}
    \log \frac{\dd \mathbb{P}_\theta}{\dd \mathbb{P}_\pi} = \log \frac{\dd \mathbb{P}_\theta}{\dd \mathbb{P}_{\text{ref}}} + \log \frac{\dd \mathbb{P}_{\text{ref}}}{\dd \mathbb{P}_\pi},
\end{align}
so that, taking the expectation under $\mathbb{P}_\theta$,
\begin{align}
    D_\mathrm{KL}(\mathbb{P}_\theta \| \mathbb{P}_\pi) = D_\mathrm{KL}(\mathbb{P}_\theta \| \mathbb{P}_{\text{ref}}) + \E_{\mathbb{P}_\theta} \left[ \log \frac{\dd \mathbb{P}_{\text{ref}}}{\dd \mathbb{P}_\pi} \right].
    \label{eq:app-soc-chain-rule}
\end{align}
The reference path measure plays here the role of an auxiliary object through which the intractable target measure $\mathbb{P}_\pi$ is reached; the first term of \cref{eq:app-soc-chain-rule} is made tractable by Girsanov's theorem (Step 2), while the second is made tractable by the fact that $\mathbb{P}_{\text{ref}}$ and $\mathbb{P}_\pi$ differ only through their initial condition \emph{in the noising direction} (Step 3).
 
\paragraph{Step 2: Girsanov's theorem and the running cost}
Let $\mathbb{P}$ and $\mathbb{P}'$ be the path measures of two It\^o diffusions sharing the same diffusion coefficient $\sigma$, with respective drifts $b$ and $b'$ and respective initial distributions $\mu$ and $\mu'$. Girsanov's theorem states that
\begin{align}
    \log \frac{\dd \mathbb{P}}{\dd \mathbb{P}'}(X) =  \log \frac{\dd \mu}{\dd \mu'}(X_0) & + \frac{1}{\sigma^2}\int_0^T (b(X_t,t) - b'(X_t,t)) \cdot \left( \dd X_t - b'(X_t,t) \, \dd t \right) \notag\\
    & - \frac{1}{2\sigma^2}\int_0^T \|b(X_t,t) - b'(X_t,t)\|^2 \, \dd t,
    \label{eq:app-soc-girsanov}
\end{align}
where $\cdot$ denotes the Euclidean inner product and the stochastic integrals are understood in the It\^o sense.
Under $\mathbb{P}$, one has $\dd X_t - b' \dd t = (b-b') \dd t + \sigma \dd W_t$, so that the stochastic integral in \cref{eq:app-soc-girsanov} is zero in expectation, and the two remaining integrals combine into
\begin{align}
    D_\mathrm{KL}(\mathbb{P} \| \mathbb{P}') = D_\mathrm{KL}(\mu \| \mu') + \frac{1}{2\sigma^2}\, \E_{\mathbb{P}} \left[ \int_0^T \|b(X_t,t) - b'(X_t,t)\|^2 \, \dd t \right].
    \label{eq:app-soc-kl-girsanov}
\end{align}
Applying \cref{eq:app-soc-kl-girsanov} to $\mathbb{P}_\theta$ and $\mathbb{P}_{\text{ref}}$, and using the drift discrepancy \cref{eq:app-soc-drift-gap}, we obtain
\begin{align}
    D_\mathrm{KL}(\mathbb{P}_\theta \| \mathbb{P}_{\text{ref}}) &= D_\mathrm{KL}(\gamma_\sigma \| \rho_{\text{ref},T}) + \frac{1}{2\sigma^2} \, \E_{\mathbb{P}_\theta} \left[ \int_0^T \|\sigma^2 u_\theta(X_t, T-t)\|^2 \, \dd t \right] \notag \\
    &= D_\mathrm{KL}(\gamma_\sigma \| \rho_{\text{ref},T}) + \frac{\sigma^2}{2} \, \E_{\mathbb{P}_\theta} \left[ \int_0^T \|u_\theta(X_t, T-t)\|^2 \, \dd t \right].
    \label{eq:app-soc-running-cost}
\end{align}
The term $D_\mathrm{KL}(\gamma_\sigma \| \rho_{\text{ref},T})$ accounts for the mismatch between the initial distribution $\gamma_\sigma$ of the generative process and the exact marginal $\rho_{\text{ref},T}$ of the noised reference; it does not depend on $\theta$ and can therefore be discarded from the optimization. It is in any case exponentially small in $T$, since the noising process converges exponentially fast towards its invariant distribution $\gamma_\sigma$.
 
\paragraph{Step 3: the terminal cost}
Evaluating the second term of \cref{eq:app-soc-chain-rule} directly in generative time would require Girsanov's theorem once more, and would produce a running cost involving the discrepancy $\nabla \log \rho_{\text{ref},T-t} - \nabla \log \pi_{T-t}$ between the reference and target scores --- which is precisely the intractable object we are trying to avoid. The invariance \cref{eq:app-soc-invariance} allows us to evaluate the same quantity in the noising direction instead, where it becomes elementary. Indeed, the two noising processes associated with $\tilde{\mathbb{P}}_{\text{ref}}$ and $\tilde{\mathbb{P}}_\pi$ obey the same SDE \cref{eq:app-soc-forward} and differ only through their initial distribution, $\rho_\text{ref}$ instead of $\pi$. Disintegrating both measures over their common transition kernel, all the conditional factors cancel and
\begin{align}
    \frac{\dd \tilde{\mathbb{P}}_{\text{ref}}}{\dd \tilde{\mathbb{P}}_\pi}(\tilde X) = \frac{\rho_\text{ref}(\tilde X_0)}{\pi(\tilde X_0)}.
    \label{eq:app-soc-forward-rn}
\end{align}
Pushing forward by $R$, and using that $(R\tilde X)_T = \tilde X_0$, the first identity of \cref{eq:app-soc-invariance} transports \cref{eq:app-soc-forward-rn} into
\begin{align}
    \frac{\dd \mathbb{P}_{\text{ref}}}{\dd \mathbb{P}_\pi}(X) = \frac{\rho_\text{ref}(X_T)}{\pi(X_T)}.
    \label{eq:app-soc-terminal}
\end{align}
What is a score discrepancy accumulated along the entire trajectory in generative time is thus, read in the other time direction, a mere terminal ratio --- and this is the reason for introducing $\mathbb{P}_{\text{ref}}$ in the first place. Using the unnormalized parametrization $\pi = e^{-U}/Z$ of the target distribution, we finally get
\begin{align}
    \E_{\mathbb{P}_\theta} \left[ \log \frac{\dd \mathbb{P}_{\text{ref}}}{\dd \mathbb{P}_\pi} \right] = \E_{\mathbb{P}_\theta} \left[ \log \frac{\rho_\text{ref}(X_T)}{e^{-U(X_T)}} \right] + \log Z,
    \label{eq:app-soc-terminal-cost}
\end{align}
where the normalization constant $Z$, although unknown, does not depend on $\theta$.
 
\paragraph{Step 4: the training objective}
Combining \cref{eq:app-soc-chain-rule,eq:app-soc-running-cost,eq:app-soc-terminal-cost}, we obtain
\begin{align}
    D_\mathrm{KL}(\mathbb{P}_\theta \| \mathbb{P}_\pi) = \E_{\mathbb{P}_\theta} \left[ \frac{\sigma^2}{2} \int_0^T \|u_\theta(X_t, T-t)\|^2 \, \dd t + \log \frac{\rho_\text{ref}(X_T)}{e^{-U(X_T)}} \right] + \text{const.},
    \label{eq:app-soc-final}
\end{align}
with $\text{const.} = \log Z + D_\mathrm{KL}(\gamma_\sigma \| \rho_{\text{ref},T})$. Both terms of the cost functional are now tractable: the running cost only involves the learnable control, evaluated along trajectories that can be simulated since the scores of the noised reference marginals are known in closed form, while the terminal cost only requires evaluating the unnormalized density $e^{-U}$ of the target.

\paragraph{Pathwise expression and log-variance loss}
The log-variance loss of \cite{richterImprovedSamplingLearned2023a} requires the log Radon-Nikodym derivative itself, and not only its expectation. Keeping the stochastic integral of \cref{eq:app-soc-girsanov} and combining it with \cref{eq:app-soc-terminal}, one gets, for a trajectory $X$,
\begin{align}
    \log \frac{\dd \mathbb{P}_\theta}{\dd \mathbb{P}_\pi}(X) = & \int_0^T u_\theta(X_t,T-t) \cdot \dd X_t - \int_0^T u_\theta(X_t,T-t) \cdot b_\text{ref}(X_t,t) \, \dd t \notag \\
    & - \frac{\sigma^2}{2} \int_0^T \|u_\theta(X_t,T-t)\|^2 \, \dd t + \log \frac{\rho_\text{ref}(X_T)}{e^{-U(X_T)}} + \log \frac{\dd \gamma_\sigma}{\dd \rho_{\text{ref},T}}(X_0) + \log Z.
    \label{eq:app-soc-pathwise}
\end{align}
Taking the variance under an arbitrary path measure $\mathbb{\tilde P}$ yields the log-variance loss $D_{\mathrm{LV}}$, for which the unknown constant $\log Z$ --- and more generally any $\theta$-independent additive term --- drops out identically, which is one of the practical advantages of this formulation.

\end{document}